\documentclass[journal]{IEEEtran}
\usepackage{amsmath,amsfonts}
\usepackage{algorithmic}
\usepackage{array}
\usepackage[caption=false,font=normalsize,labelfont=sf,textfont=sf]{subfig}
\usepackage{textcomp}
\usepackage{stfloats}
\usepackage{url}
\usepackage{verbatim}
\usepackage{graphicx}
\def\BibTeX{{\rm B\kern-.05em{\sc i\kern-.025em b}\kern-.08em
    T\kern-.1667em\lower.7ex\hbox{E}\kern-.125emX}}
\usepackage{balance}
\usepackage{epsfig}
\usepackage{multirow}
\usepackage{overpic}
\usepackage{wrapfig,lipsum,booktabs}
\usepackage{cite}
\usepackage{color}
\usepackage{algorithm}
\usepackage{algorithmic}
\usepackage{hyperref}
\ifdefined \GramaCheck
  \newcommand{\CheckRmv}[1]{}
  \newcommand{\figref}[1]{Figure 1}%
  \newcommand{\tabref}[1]{Table 1}%
  \newcommand{\secref}[1]{Section 1}
  \renewcommand{\eqref}[1]{Equation 1}
\else
  \newcommand{\CheckRmv}[1]{#1}
  \newcommand{\figref}[1]{Fig.~\ref{#1}}%
  \newcommand{\tabref}[1]{Table~\ref{#1}}%
  \newcommand{\secref}[1]{Section~\ref{#1}}
  \renewcommand{\eqref}[1]{Equation~(\ref{#1})}
\fi

\newcommand{\tabFormat}{\centering \renewcommand{\arraystretch}{1.05}}
\newcommand{\myPara}[1]{\subsubsection{#1}}
\newcommand{\ourM}{{FsaNet~}}
\newcommand{\sArt}{{state-of-the-art~}}
\newcommand{\FCN}{{FCN~\cite{long2015fully}}}
\newcommand{\EncNet}{{EncNet~\cite{zhang2018context}}}
\newcommand{\APCNet}{{APCNet~\cite{he2019adaptive}}}
\newcommand{\ANN}{{ANN~\cite{zhu2019asymmetric}}}
\newcommand{\GCNet}{{GCNet~\cite{cao2019gcnet}}}
\newcommand{\DMNet}{{DMNet~\cite{he2019dynamic}}}
\newcommand{\DNLNet}{{DNLNet~\cite{yin2020disentangled}}}
\newcommand{\CCNet}{{CCNet~\cite{huang2020ccnet}}}
\newcommand{\DANet}{{DANet~\cite{fu2019dual}}}
\newcommand{\ISANet}{{ISANet~\cite{huang2019interlaced}}}
\newcommand{\UPerNet}{{UPerNet~\cite{xiao2018unified}}}
\newcommand{\NonlocalNetwork}{{Non-local~Network~\cite{wang2018non}}}
\newcommand{\EMANet}{{EMANet~\cite{li2019expectation}}}
\newcommand{\OCRNet}{{OCRNet~\cite{yuan2020object}}}
\newcommand{\PSANet}{{PSANet~\cite{zhao2018psanet}}}
\newcommand{\PSPNet}{{PSPNet~\cite{zhao2017pyramid}}}
\newcommand{\DeepLabVthree}{{DeepLabV3~\cite{chen2017rethinking}}}
\newcommand{\DecoupledSegNets}{{DecoupledSegNets~\cite{li2020improving}}}
\newcommand{\DGCNet}{{DGCNet~\cite{zhang2019dual}}}
\newcommand{\HANet}{{HANet~\cite{choi2020cars}}}
\newcommand{\GloRe}{{GloRe~\cite{chen2019graph}}}
\newcommand{\SRNet}{{SRNet~\cite{Li2020SRNet}}}
\newcommand{\CGBNet}{{CGBNet~\cite{ding2020semantic}}}
\newcommand{\bq}{\mathbf{q}}
\newcommand{\bk}{\mathbf{k}}
\newcommand{\bv}{\mathbf{v}}
\newcommand{\Softmax}{\mathrm{Softmax}}
\newcommand{\bbR}{\mathbb{R}}

\begin{document}
\title{FsaNet: Frequency Self-attention for Semantic Segmentation}
\author{
Fengyu Zhang,~
Ashkan Panahi,~
and Guangjun Gao
\thanks{F. Zhang and G. Gao are with the School of Traffic \& Transportation Engineering, Central South University, Changsha, 410075, China (e-mail: fengyuzhang@csu.edu.cn; gjgao@csu.edu.cn).  A. Panahi and also F. Zhang are with the Department of Computer Science and Engineering, Chalmers University of Technology, SE-412 96 Göteborg, Sweden.
}}

\markboth{IEEE Transactions on Image Processing,~Vol.~xx, No.~xx, June~2023}%
{Zhang \MakeLowercase{\textit{et al.}}: 
FsaNet: Frequency Self-attention for Semantic Segmentation}

\maketitle
\begin{abstract}
Considering the spectral properties of images, we propose a new self-attention mechanism with highly reduced computational complexity, up to a linear rate. To better preserve edges while promoting similarity within objects, we propose individualized processes over different frequency bands. In particular, we study a case where the process is merely over low-frequency components. By ablation study, we show that low frequency self-attention can achieve very close or better performance relative to full frequency even without retraining the network. Accordingly, we design and embed novel plug-and-play modules to the head of a CNN network that we refer to as FsaNet. The frequency self-attention 1) requires only a few low frequency coefficients as input, 2) can be mathematically equivalent to spatial domain self-attention with linear structures, 3) simplifies token mapping ($1\times1$ convolution) stage and token mixing stage simultaneously. We show that frequency self-attention requires $87.29\% \sim 90.04\%$ less memory, $96.13\% \sim 98.07\%$ less FLOPs, and $97.56\% \sim 98.18\%$ in run time than the regular self-attention. Compared to other ResNet101-based self-attention networks, \ourM achieves a new \sArt result ($83.0\%$ mIoU) on Cityscape test dataset and competitive results on ADE20k and VOCaug. \ourM can also enhance MASK R-CNN for instance segmentation on COCO. In addition, utilizing the proposed module, Segformer can be boosted on a series of models with different scales, and Segformer-B5 can be improved even without retraining. Code is accessible at 
\url{https://github.com/zfy-csu/FsaNet}.
\end{abstract}


\maketitle
\IEEEdisplaynontitleabstractindextext
\IEEEpeerreviewmaketitle

\section{Introduction}\label{sec:introduction}
\IEEEPARstart{S}{emantic} segmentation is the task of classifying different compartments of an image at a fine level of pixels. It is widely used in the fields of autonomous driving, industrial defect detection, and medical tissue segmentation. To improve the performance of semantic segmentation, two main principles are improving the inner consistency of the objects in an image and refining their detail along their boundaries. From the perspective of traditional image processing, the inner body and the edge correspond to the low and high frequency components, respectively. Self-attention is a newly developed technique that can effectively improve the object's inner consistency by modeling the non-local context. However, it adopts identical learnable parameters to process all frequency components, which makes it difficult to achieve these two goals at the same time. On the other hand, the computational cost of self-attention is extremely high and the network integrated with self-attention has high requirements on the hardware platform. This is a huge obstacle for real-world applications, especially in mobile scenarios such as autonomous driving.

This paper addresses the above problems by explicitly utilizing Discrete Cosine Transform (DCT) and processing the data in the frequency domain. We first conduct suitable ablation studies confirming that global self-attention mainly focuses on low frequency components. Moreover, by visualization, we find that full-frequency processing promotes internal consistency, but destroys edge details. These two facts suggest that low-frequency components should be processed individually. However, introducing the operation of extracting low-frequency components from the input can exacerbate the computational cost of self-attention. This motivates us to design a more ingenious attention mechanism that requires only a few low-frequency coefficients as input. In this way, the proposed method can effectively improve the segmentation accuracy while reducing the complexity by exploiting the sparsity in the frequency domain.
     
We start our study on CNN-based self-attention networks. In this scenario, the input is a sequence of vectors obtained by row-wise unrolling the feature maps extracted from a convolutional network. A vector in the input is often referred to as a token. Denote the size of the feature maps by $H\times W\times C$. Then, the number of tokens $N$ is equal to $HW$. Each output token is a weighted sum of all input tokens, where the weights are a measure of similarity between individual tokens. The similarity between all tokens needs to be calculated and stored in a so-called attention map of size $N\times N$. Both memory and computation are quadratic with respect to $N$. The situation is even worse as dense prediction tasks tend to have higher resolution feature maps.

In response, we design a self-attention mechanism that heavily reduces the computational complexity, while improving the quality of the attention mechanism. We achieve this goal by first transforming the tokens into the frequency domain and alternatively taking frequency coefficients as input.  As a result, we can easily decouple different frequencies and simplify the computation by taking advantage of the sparsity in the frequency domain.  While the easiest decoupling method is to transform the input to the frequency domain, retain the specified frequency coefficients, and then reconstruct a new input, in this way, the sparsity in the frequency domain cannot be effectively utilized. Hence, we find an equivalent method to ensure that the result of reconstructing the new frequency coefficients is as consistent as possible with direct calculation in the spatial domain. Some aspects of our method are briefly explained in the sequel.

{\bf Replacing the input of self-attention} The input of a conventional self-attention is a one-dimensional vector extracted row-wise from a feature map. However, if the frequency domain transformation is performed directly on the vector, the similarity in both vertical and horizontal directions cannot be effectively utilized, the original spatial relations are lost, and the resulting frequency sparsity will be insignificant. Therefore, we restore the vector as a feature map, perform a two-dimensional frequency domain transform, retain the specified frequency block, and expand it row by row as a new input. To simplify the calculation, we equivalently replace the above series of operations by multiplying a projection matrix. With fixed crop size, the matrix only needs to be calculated once during the entire process of training or inference.

\textbf{Modifying the structure of self-attention} The transformation of the input from the spatial domain to the frequency domain is linear. If the expression of self-attention also has a linear computational structure, it will be possible to transfer the self-attention operations from the spatial domain to the frequency domain by the associative law of matrix multiplication. In the standard approach, non-linearity is solely introduced by the exponential function in Softmax. We replace the exponential function with a linear approximation to obtain two self-attention mechanisms with linear structure. The replacement also facilitates the acquisition of global contextual information due to a more uniform distribution of attention weights.

\textbf{The benefits of frequency self-attention} Based on the above, self-attention applying to spatial position tokens is transformed into self-attention applying to frequency coefficients representing different scales. Thanks to the orthonormal nature of DCT transform, each frequency coefficient is independent and global. As such, removing frequency coefficient tokens will not do harm to receptive field. Benefiting from the sparsity of the frequency domain, the token number can be greatly reduced. The method of compressing the computation in the frequency domain is tailored to the general characteristics of images. Even if the spatial self-attention is replaced directly by low frequency self-attention, it will only cause a trivial difference in results. Therefore, the number of retained frequency components can be quickly determined through ablation study on input frequency components. In addition, we move the mapping convolution behind dimension reduction, which further reduces computational costs. 

\textbf{Contributions}
 We highlight the novelties below:\\
 •	We develop two new self-attention modules, solely processing low frequency components, which can preserve edge details while promoting subject similarity.\\
 •	We translate a series of operations of reshape, 2D-DCT, and 2D-IDCT to a linear map, yielding a simple transformation form between spatial position tokens and frequency tokens.\\
 •	We show the proposed frequency self-attention is mathematically equivalent to spatial self-attention with linear structure.\\
 •	We experimentally verify that \ourM outperforms or is comparable to the \sArt methods and has the lowest computational cost. 

The remainder of this article proceeds as follows: \secref{sec:relatedwork} gives a brief review of related works. Our method is discussed in \secref{sec:Methodology}. 
\secref{sec:experiments} analyzes ablation study on frequency components of self-attention and reports the performance in segmentation tasks on 4 datasets. The complexity analysis of our technique is presented in \secref{sec:Experiments_Complex}. Finally, concluding remarks and future works are described in \secref{sec:futurework}.

\section{Related Work}\label{sec:relatedwork}
In this section, we first briefly describe several popular semantic segmentation algorithms. Then, we review relevant simplified self-attention solutions in CNN and transformer.

\subsection{Semantic segmentation}\label{sec:Current_works}
At present, the most successful semantic segmentation methods employ general-purpose backbones to encode features, such as ResNet\cite{he2016deep}, and ViT\cite{dosovitskiy2020image}. Then the features will be decoded to the pixel space by a decode-head network. \FCN~is the first encoder-decoder method. Since then, many improvements have been proposed by backbone enhancement, context module design, decoupling body and edge, and better feature supervision. Our work is an improvement work from the perspective of context module design and decoupling.

\textbf{Backbone enhancement} CNN networks are widely used traditional visual backbones. For better performance, HRNet\cite{wang2020deep} designs a wide backbone to maintain high resolution; U-Net\cite{li2018h} utilizes features in multi-stage with long-range skip connections. For real-time semantic segmentation, some lightweight backbones are proposed, such as bisenetv2\cite{yu2021bisenet} and CGNet\cite{wu2020cgnet}. GFF\cite{li2020gated} selectively fuses features from multiple levels using gates. EFCN\cite{shuai2018toward} introduces a dense-skip architecture to retain low-level information and a convolutional network to aggregate contexts. K-Net\cite{zhang2021k} proposes a unified framework for image segmentation by a group of learnable kernels. Video K-Net\cite{li2022video} learns to segment and track things and stuff with kernel-based appearance modeling and cross-temporal kernel interaction. After ViT\cite{dosovitskiy2020image} introduced Transformer from natural language processing (NLP) to computer vision (CV), Transformer backbones are becoming popular. SETR\cite{zheng2021rethinking}, Segmenter\cite{strudel2021segmenter}, SegFormer\cite{xie2021segformer}, SWIN\cite{liu2021swin}, TWINs\cite{chu2021twins} and MaskFormer\cite{cheng2021per} outperform CNN-based methods. Subsequently, some CNN-based methods achieved new state-of-the-art results, such as DWNet\cite{han2021demystifying} and ConvNeXt\cite{liu2022convnet}. CNN-Transformer hybrid backbone is the latest trend, such as Cvt\cite{wu2021cvt}, Next-ViT\cite{li2022next}.

\textbf{Context module design} Multiple types of context modules have been proposed to increase the receptive field size. \PSPNet~presented a pyramid pooling module by adaptively pooling at different scales. Compared with \PSPNet, \HANet~ introduces a height prior to avoid confusion in predicting the elevation of the objects; \UPerNet~ fuses different scale information at different stages of upsampling. \CGBNet~enhances the paring results by context encoding and multi-path decoding. \DeepLabVthree~proposed a spatial pyramid pooling module by pooling and atrous convolution with different dilated rates. Based on \DeepLabVthree, DeepLabv3+\cite{chen2018encoder} fuses low-level features with high-level features to improve the accuracy of segmentation boundaries. \GCNet~learns a globally shared spatial attention to collect context. \DMNet~perform depth-wise convolution with filters of different sizes, which are dynamically generated through adaptive pooling at different scales. \EncNet~adopts channel attention to highlight class-dependent feature and introduces a loss to de-emphasize the probability of a vehicle appearing in an indoor scene. GALD\cite{li2021global} uses global aggregation and then local distribution to avoid oversmooth regions that contain small patterns.
After \NonlocalNetwork~introduces self-attention as a context module, a lot of follow-up studies have been proposed. \DANet~adopts an extra channel attention module which is placed in parallel. Similarly, \DGCNet~models contextual information in spatial and channel dimensions by two branches of graph convolutional network. \PSANet~has collection branch and distribution branch for adaptively learning attention maps, but requires more parameters and memory. \GloRe~designs a global reasoning unit that implements the coordinate-interaction space mapping by pooling and broadcasting, and the relation reasoning via graph convolution. \SRNet~squeezes the input feature into a channel-wise global vector and
perform reasoning within the single vector. There is also plenty of research devoted to simplifying self-attention, and related works are detailed in \secref{Self-attention simplification in CNN}.

\textbf{Decoupling body and edge} The two tasks of improving edge contrast and consistency within objects are contradictory in nature. Therefore, some methods decouple these two processes. Gated-SCNN\cite{takikawa2019gated} considers an extra shape stream branch, which only processes boundary information. \DecoupledSegNets~split the body feature and edge feature by flow filed learning. RPCNet\cite{zhen2020joint} enhance edge by supervising the boundary obtained by a spatial derivative. STDC-Seg\cite{fan2021rethinking} adopts the Laplacian convolution kernel to generate edges to assist in the supervision of the segmentation. \DNLNet~divides the attention map into two independent terms, which tend to focus on the within-category and boundary respectively. Although these studies put forward the concept that the body and edge correspond to the high and low frequency components, no real frequency decomposition is performed. Our work is the first to decouple the high and low frequency in the frequency domain by signal processing technique, which allows to reduce computational cost while improving performance.

\textbf{Better feature supervision} ContrastiveSeg\cite{wang2021exploring} optimizations the training process by designing a loss function based on contrastive learning and taking the global context between pixels across different images as supervisory information. In contrast to the context module design method, the global context is not fused into the original features. It is only used for supervision and will be removed during testing. ProtoSeg\cite{zhou2022rethinking} represents each class as a set of non-learning prototypes by online clustering and achieves the prediction by nearest prototype retrieving. Based on this, it proposes a novel decoding network and loss function to better monitor the pixel embedding. These two methods are orthogonal to the various methods mentioned above and can be combined together to improve performance.

\subsection{Self-attention simplification}
Self-attention simplification in various scenarios has been extensively studied due to its quadratic complexity. Our work can also be regarded as a self-attention simplification. 

\textbf{Self-attention simplification in CNN}\label{Self-attention simplification in CNN}
Some methods treat key and value as bases and reduce their number. In \CCNet, every pixel only collect information from the criss-cross path and obtain global information by recurrent processing. \ISANet~adopts a short-range attention implemented within a block and a long-range attention implemented between blocks. \ANN~introduces pyramid pooling into key and value branches. ACFNet\cite{zhang2019acfnet} and \OCRNet~characterize a pixel based on the object class features, which are calculated by pre-evaluated coarse segmentation results. \EMANet~removes the convolution for query and iteratively estimates a compact set of Gaussian bases by expectation-maximization. Some linear methods remove the restriction of Softmax to avoid explicitly modeling the attention map. Linear Attention\cite{li2020linear} considers a linear attention module, based on the first-order Taylor expansion. A2-Net\cite{chen20182} gathers a compact set of features through second-order attention pooling and modified the structure of self-attention into implement Softmax separately on query and key.

\textbf{Self-attention simplification in Transformer}
Some methods reduce the length of the key-value pairs. In CSWin\cite{dong2021cswin}, query tokens only concentrate on a cross-shaped window with a certain width. Twins\cite{chu2021twins} uses both a local self-attention and a global self-attention implemented on tokens achieved by down-sampling each local window. Sparse Transformer\cite{child2019generating} retains the top-$k$ contributive elements in each attention map. Reformer\cite{kitaev2019reformer} searches and computes only some large attention values via locality sensitive hashing. 
There are also linear methods, such as the Fast Autoregressive Transformers\cite{katharopoulos2020transformers}, Performer\cite{choromanski2020rethinking}, UFO-ViT\cite{song2021ufo}, and SOFT\cite{lu2021soft}. Furthermore, AFT\cite{zhai2021attention} replaces matrix products with Hadamard products. 
Some methods employ self-attention locally. HaloNet\cite{vaswani2021scaling} implements self-attention on non-overlapping blocks internally, and padding the blocks with boundary pixels to compensate for severe global information loss. Swin\cite{liu2021swin} computes self-attention within a local window and exchanges information between different windows by a shifted window method. T2T-ViT\cite{yuan2021tokens} structurize the image gradually by recursively aggregating adjacent tokens. 

\textbf{\ourM vs. other simplification methods}
From a dimension reduction perspective, \ourM not only reduces the dimension for key and value, but also reduces the dimension for query at the same time. Compared with the above ways of selecting bases, we leverage the frequency domain sparsity to generate bases, which has better interpretability. Unlike the current linear attention methods, which first calculate the product of the key and value matrices, we combine dimension reduction and cleverly propose a better order of computation so that the mapping convolution is also simplified. In addition, our method can also be viewed as local self-attention in the frequency domain, but requires no additional operations to compensate for global information.

\section{Methodology}\label{sec:Methodology}
\CheckRmv{
\begin{figure}[t]
  \centering
  \begin{overpic}[width=0.6\linewidth]{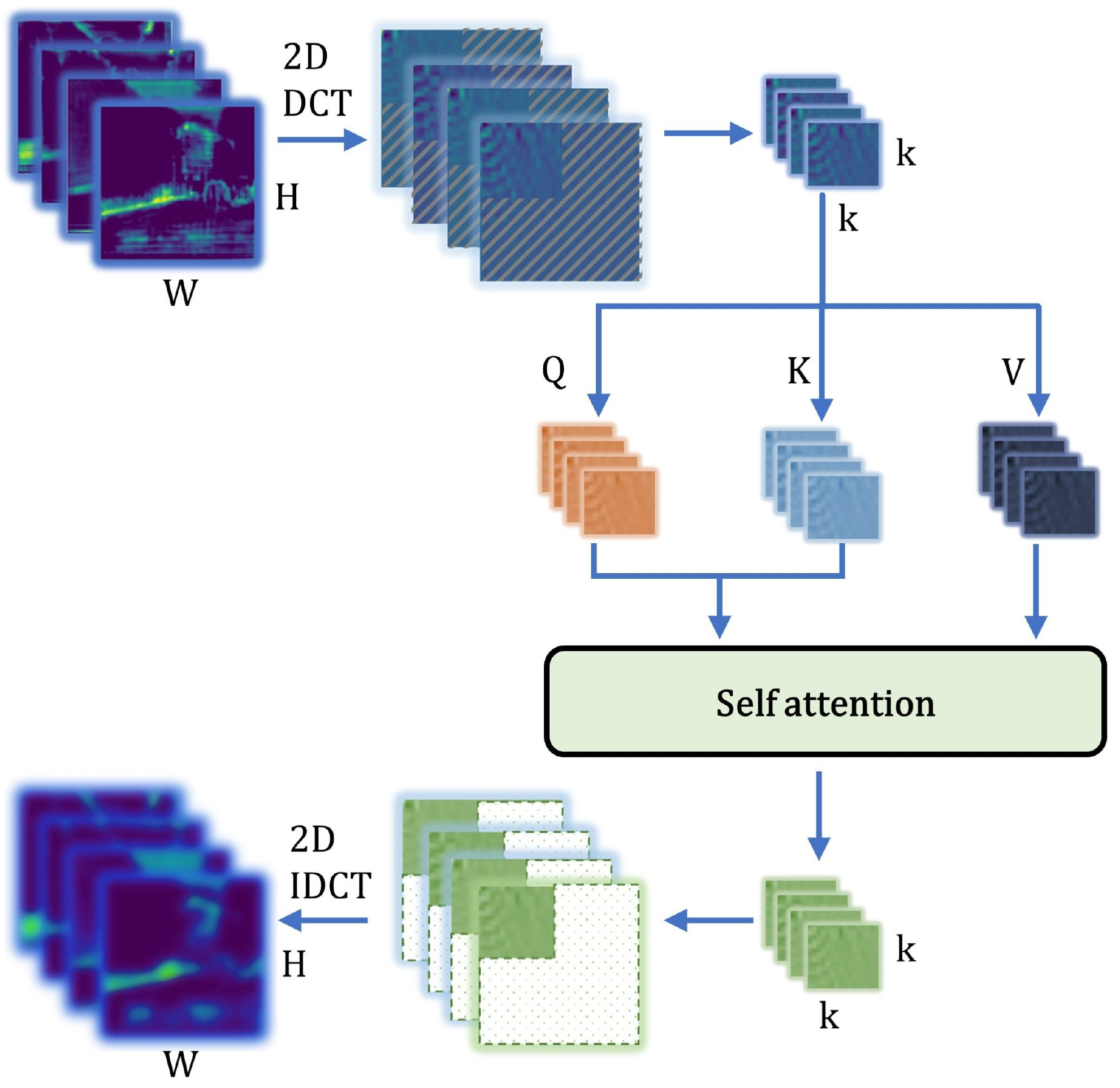}
  \end{overpic}
  \caption{The structure of frequency self-attention}
  \label{fig:networkstruct}
\end{figure}
}
The overall structure of our proposed network is shown in \figref{fig:networkstruct}. Its data processing flow is depicted by the green path in \figref{fig:idealgoal}, which is mathematically equivalent to the more intuitive, but computationally expensive, red path. We explain them and their equivalence.
In our setting, the raw input images are first processed by a backbone network to obtain the feature maps $X\in \bbR^{C\times H\times W}$. In a standard architecture, each 2D map is row-wise expanded (vectorized), yielding the input $X^{'}\in \bbR^{C\times HW}$ of a conventional spatial self-attention unit. Spatial self-attention treats all frequency components indiscriminately, which may cause great distortion on the edges while promoting the consistency of low-frequency components. Therefore, we first remove the high frequency components from $X$ and then feed the result, containing low frequency components, to the self-attention unit. We keep $k$ horizontal low frequency components and $k$ vertical low frequency components, where $k<H$ and $k<W$. The red path in \figref{fig:idealgoal} is the conceptual depiction of this procedure, where each channel in the input $X$ undergoes a 2D-DCT transformation to achieve $F\in \bbR^{C\times H\times W}$. Next, only the specified low frequency block $f\in \bbR^{C\times k\times k}$ is retained, from which new feature maps $X_{f}\in \bbR^{C\times H\times W}$ are reconstructed by an inverse 2D-DCT transform. Finally, $X_{f}$ is row-wise expanded to achieve the input $X_{f}^{'}\in \bbR^{C\times HW}$ of a spatial self-attention unit and the output is $O^{'}\in \bbR^{C\times HW}$. \\
\CheckRmv{
\begin{figure}[t]
  \centering
  \begin{overpic}[width=\linewidth]{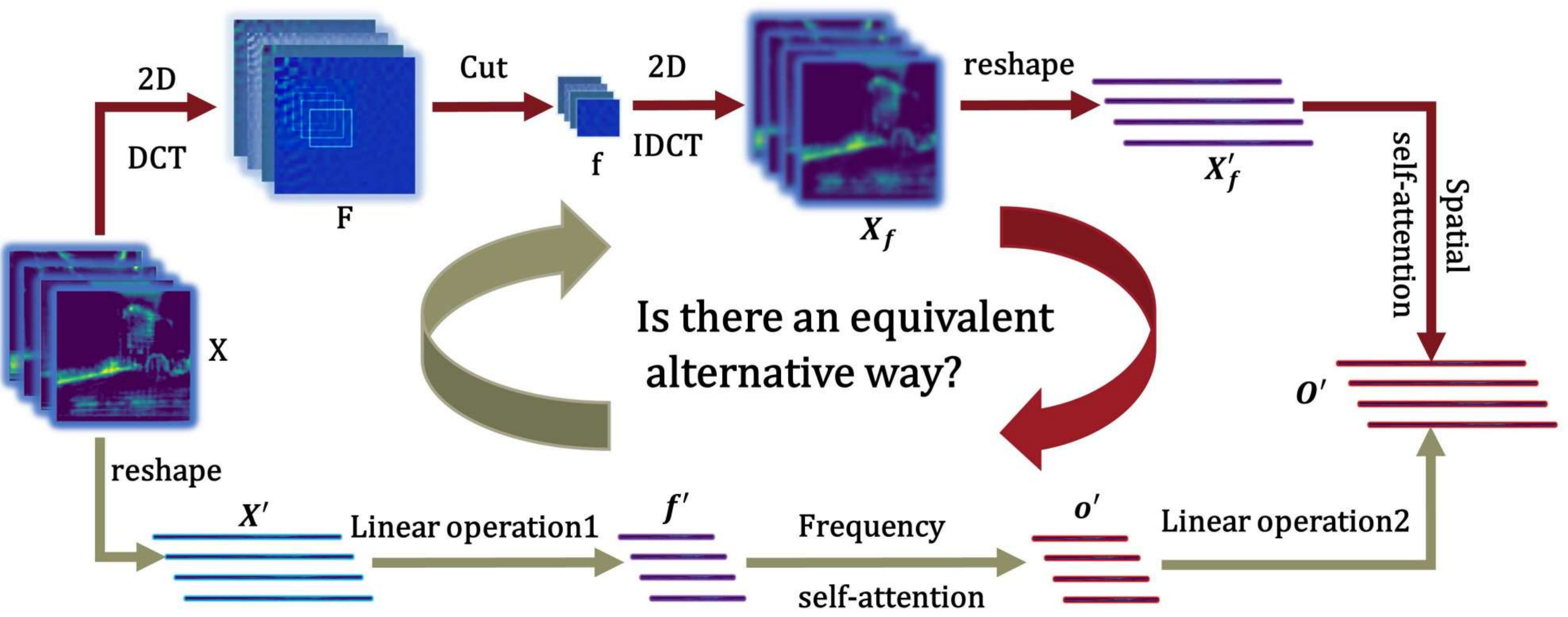}
  \end{overpic}
  \caption{The ideal goal}
  \label{fig:idealgoal}
\end{figure}
}
\newcommand{\addFig}[1]{{\includegraphics[height=.087\textwidth]{#1.pdf}}}
\CheckRmv{
\begin{figure}[t]
  \centering
  \renewcommand{\arraystretch}{0.5}
  \setlength{\tabcolsep}{0.00mm}
  \begin{tabular}{cc}
    \addFig{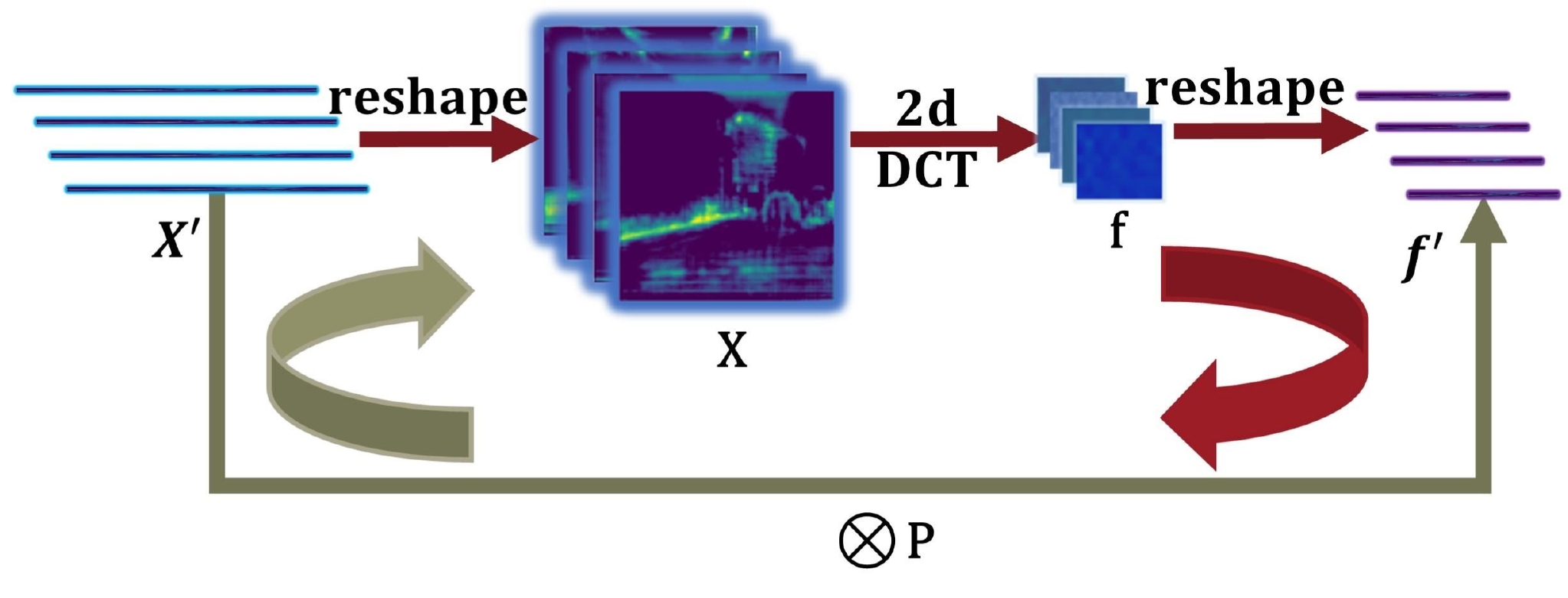}&
    \addFig{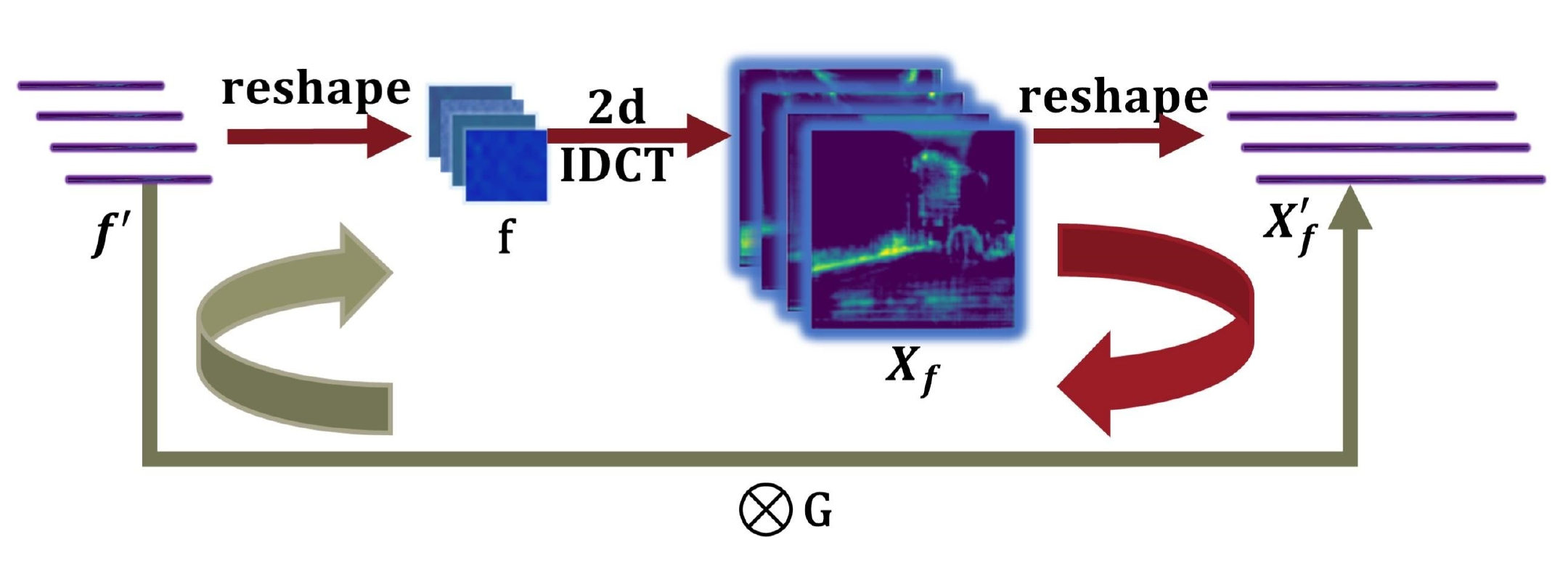}\\
    (a) Equivalence path for $P$ & (b) Equivalence path for $G$\\
  \end{tabular}
  \caption{Conversion between spatial position tokens and frequency tokens}
  \label{fig:assumption23}
\end{figure}
}
The details can be improved when self-attention takes $X^{'}_{f}$ as input. However, the additional processing of frequency domain decomposition and reconstruction in the red path result in even higher complexity than a conventional self-attention layer. Based on the frequency ablation study in \secref{sec:experiments:frequency ablation}, the low frequency block that needs to be preserved tends to be very small. This condition allows us to cut down the computation by designing the equivalent green path in \figref{fig:idealgoal}.
The green path consists of Linear operation 1 that directly calculates the low frequency coefficients (hence reducing dimensionality) and Linear operation 2 that reconstructs the spatial features from the processed frequency coefficients (expanding dimensionality), after a key process called frequency self-attention. Since frequency self-attention acts on extremely low-dimensional features, the green path has very low complexity. 

{\bf Linear Operations 1 and 2:} \figref{fig:assumption23}(a) and (b) constitute the implementation of Linear operations 1 and 2, respectively. Note that in the green path, the input of frequency self-attention  $f{'}\in \bbR^{C\times k^{2}}$ is equal to the row-wise expansion of $f$ in the red path. The input $X$ is also reshaped to $X^{'}$. These facts define Linear operation 1 as a map between $X'$ and $f'$, consisting of reshape and 2D-DCT operations, as depicted in the top path of \figref{fig:assumption23}(a). These operations are represented by a matrix $\mathbf{P}\in \bbR^{HW\times k^2}$ to facilitate the explicit derivation of the formulas. Similarly, we propose a matrix $\mathbf{G}\in \bbR^{k^{2}\times HW}$ to be equivalent to the operations that transform $f'$ to $X_{f}^{'}$ as shown in Fig.3(b). Based on this, The process of generating $X_{f}^{'}$ from $X'$ includes $f'=X^{'}\mathbf{P}$ and $X_{f}^{'}=f'\mathbf{G}$. 
Linear operation 1 corresponds to multiplying $\mathbf{P}$. Linear operation 2 consists of a series of operations related to $\mathbf{G}$, which transform the frequency self-attention output $o{'}\in \bbR^{C\times k^{2}}$ to the spatial self-attention output $O'$.
Now, we may calculate $\mathbf{P},\mathbf{G}$. By setting $X^{'}$ to an $HW\times HW$ identity matrix and processing it according to the red path in \figref{fig:assumption23}(a), we obtain $\mathbf{P}$. By taking $f^{'}$ as a $k^{2}\times k^{2}$ identity matrix and processing it according to the red path in \figref{fig:assumption23}(b), we obtain $\mathbf{G}$. We also find that decomposition and reconstruction with $\mathbf{P}$ and $\mathbf{G}$ have similar properties to the decomposition and reconstruction of 1D-DCT, namely $\mathbf{G=P}^{T}$ and $\mathbf{P}^{T}\mathbf{P}$ is identity. We provide a detailed discussion and pseudo code for $\mathbf{P},\mathbf{G}$ in the appendix.  

{\bf Frequency Self-attention:} Note that frequency self-attention does not have as clear intuition as spatial self-attention and is instead defined by an equivalence principle. 
Hence, we assume that frequency self-attention shares identical learnable parameters with its spatial counterpart. This enables the two paths to exchange parameters without even retraining. 


Spatial self-attention consists of two stages: token mapping and token mixing. Token mapping is common among most self-attention mechanisms. $X^{'}$ is first linearly mapped into matrices $Q$, $K$, $V$. The learnable parameters of self-attention are introduced in the token mapping stage, including $W_{q}\in \bbR^{d_{k},C}$, $B_{q}\in \bbR^{d_{k},1} $, $W_{k}\in \bbR^{d_{k},C}$, $B_{k}\in \bbR^{d_{k},1}$, $W_{v}\in \bbR^{d_{v },C}$, and $B_{v}\in \bbR^{d_{v},1}$. The relation between these parameters can be described by $Q=W_{q}X^{'}+B_{q}$, $K=W_{k}X^{'}+B_{k}$ and $V=W_{v}X^{'}+B_{v}$.
Token mixing refers to the evaluation of the output. Our analysis is based on Non-local Embedding Gaussian self-attention\cite{wang2018non}, in which the process can be described as:
\begin{eqnarray}\label{eq:Non-local Embedded Gaussian}
O^{'}=V\mathrm{Softmax}(K^{T}Q).
\end{eqnarray}

In $Q$, $K$, and $V$, each column corresponds to a q(uery), k(ey), and v(alue) vector, respectively. Each column of $K^{T}Q$ corresponds to the similarity between one $\bq$ vector (a column of $Q$) and all $\bk$ vectors (columns of $K$). Next, Softmax is applied to each column to ensure that the weights are positive and sum to $1$. By doing this, each output vector is computed as a weighted sum of all $\bv$ vectors. It is worth noting that by experiments in \secref{sec:experiments:bias ablation} we found that $B_{q}$, $B_{k}$, and $B_{v}$ can be removed in a well-trained network, with no harm to the performance. Therefore, we removed these biases in subsequent analyses. To take low frequency feature maps as input, we replace $X^{'}$ with $X^{'}_{f}=f'G$. Correspondingly, the $Q$, $K$ and $V$ can be replaced with $Q_f=W_{q}X^{'}_{f}=W_{q}f^{'}G$, $K_f=W_{k}X^{'}_{f}=W_{k}f^{'}G$, and $V_f=W_{v}X^{'}_f=W_{v}f^{'}G$. Further, the replacement corresponds to the red path and yields:
\begin{eqnarray}\label{eq:lowfrequency_attention}
    \begin{small}
    O^{'}=W_{v}f^{'}G\mathrm{Softmax}[(W_{k}f^{'}G)^{T}(W_{q}f^{'}G)]
    \end{small}
\end{eqnarray}

Even if the Softmax operator is nonlinear, the expression in \eqref{eq:lowfrequency_attention} can be viewed as a composition of a series of operations, which can be grouped in different ways based on the associative law of function composition. We consider the following grouping:
\begin{eqnarray}\label{eq:green_path_grouping}
    \begin{small}  O^{'}=\underbrace{(W_{v}f^{'})\underbrace{G\mathrm{Softmax[G^{T}}}\underbrace{(W_{k}f^{'})^{T}(W_{q}f^{'})}}G].
    \end{small}
\end{eqnarray}

 The grouping interprets the output 
 as a frequency self-attention mechanism, corresponding to the green path. However, the proposed grouping does not provide a computational benefit due to the Softmax. Our remedy is to seek a proper spatial self-attention structure by suitably approximating Softmax that allows a more efficient implementation.

\subsection{What's a proper spatial self-attention structure for FsaNet?}
Let us start with the definition of Softmax, which consists of applying an exponential function and the sum normalization:
$\Softmax(x)_i = e^{x_i}/\sum\limits_{j}e^{x_j}$.
Let $e^{K^{T}Q}$ denote the element-wise exponent of $K^{T}Q$ and define $\lambda\in \bbR^{1\times HW}$ as the element-wise inverse of the column-wise sum of $e^{K^{T}Q}$. We may also write  $\lambda=1./\left(\mathbf{1}_{HW}^{T} e^{K^{T}Q}\right)$. Note that $\Softmax(K^{T}Q)$ is the column-wise application of the softmax operator, which corresponds to the sum normalization of each column of $e^{K^{T}Q}$. This can be implemented by multiplying $e^{K^{T}Q}$ from right to a diagonal matrix $\rho$ with the elements of $\lambda$ as its diagonal. Hence, \eqref{eq:Non-local Embedded Gaussian} can be reformulated as:
\begin{eqnarray}
  \begin{aligned}
     O'=V e^{K^{T}Q} \rho =V e^{K^{T}Q} \left(1./\mathrm{diag}\left(\mathbf{1}_{HW}^{T} e^{K^{T}Q}\right)\right).
  \end{aligned}
\end{eqnarray}

Now we observe that the nonlinearity of the exponential function prevents a further simplification, and an approximation is hence sought. It is not difficult to think of approximating the exponential function by its first-order Taylor series at $0$. Then, the entire procedure will include only matrix multiplication and addition, as we elaborate. Based on the approximation of the exponential terms, we obtain an approximation of Softmax, which we refer to as LinSoftmax:
\newcommand{\LinSoftmax}{\mathrm{LinSoftmax}}
\begin{eqnarray}
\LinSoftmax(x)_{i} = (1+x_i)/\sum\limits_{j}(1+x_j)
\end{eqnarray}

In order to reduce the error caused by approximation, we modify the attention module to ensure each element in $K^TQ$ falls in $[-1\ 1]$. In this interval, $1+x$ is close to $e^{x}$. When $x>1$, $e^{x}$ grows much faster than $1+x$. When $x< -1$, $e^{x}$ approaches $0$ but $1+x$ decreases unboundedly. The modification should not introduce substantial computational limitations. Our main strategy is to introduce two affine maps $\rho_{k}\in \bbR^{HW\times HW}$ and $\rho_{q}\in \bbR^{HW\times HW}$, respectively normalizing $K$ and $Q$. To avoid arbitrary matrix operations of size $HW\times HW$, we let $\rho_{k}$ and $\rho_{q}$ be diagonal. Thus, the corresponding matrix multiplications can be replaced by the Hadamard product of their diagonal vectors $\lambda_k\in \bbR^{1\times HW}$ and $\lambda_q\in \bbR^{1\times HW}$, which is significantly faster. In summary, we modify \eqref{eq:Non-local Embedded Gaussian} to: 
\begin{eqnarray}\label{linsoftmax_rho_kq}
O'= V\mathrm{LinSoftmax}((K\rho_k)^T  Q\rho_q).
\end{eqnarray}

In the following, we consider two choices for $\rho_q$ and $\rho_k$ and consequently achieve two kinds of self-attention structures for FsaNet. The first one has a highly simplistic design and the second one is more complex.

\myPara{Non-local Dot product self-attention structure}\label{sec:startegy1}
The first strategy is to choose $\rho_q$ and $\rho_k$ as identity matrices when the self-attention is placed after the CNN backbone. We provide a combination of analytical and empirical evidences to support this choice. We conducted experiments on Cityscapes\cite{cordts2016cityscapes} val dataset based on four trained CNN networks with the module in \eqref{eq:Non-local Embedded Gaussian} provided by a semantic segmentation toolbox MMSegmentation\cite{mmseg2020}. After training, we studied the elements of $K^T Q$ and observed that only about 0.005\% of the elements are out of $[-1\ 1]$. We also observed that the distribution of dot product values for all points is similar, which is visualized in \figref{fig:distribution}. 
\CheckRmv{
\begin{figure}[t]
  \centering
  \begin{overpic}[width=0.7\linewidth]{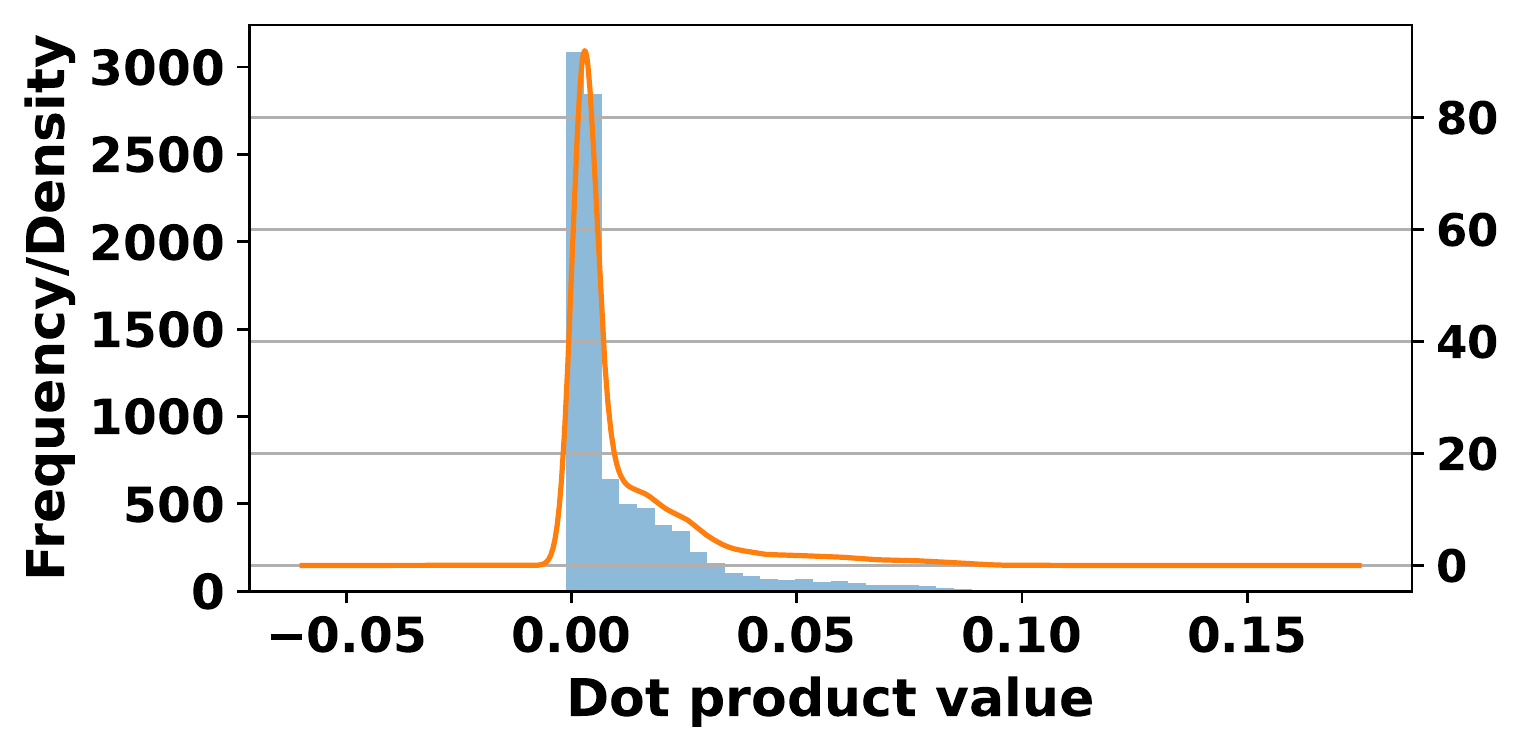}
  \end{overpic}
  \caption{Distribution of dot product values for one point}
  \label{fig:distribution}
\end{figure}
}
As seen, most dot product values are close to $0$. Based on this observation, we may ignore normalization by setting $\lambda_q$ and $\lambda_v$ to all-one vectors. Correspondingly, $\rho_q$ and $\rho_k$ are identity matrices. As a result, we directly replace Softmax with LinSoftmax. Our numerical studies show that the linear replacement will lead to only a 0.1\% performance drop in mIoU. In this case, \eqref{eq:Non-local Embedded Gaussian} can be simply approximated by $O'=V\mathrm{LinSoftmax}(K^T Q)$. With $\lambda=1./(HW\mathbf{1}_{HW}^{T}+\mathbf{1}_{HW}^{T}K^T Q)$ and $\rho=\mathrm{diag}(\lambda)$, we may write:
\begin{eqnarray}\label{eq:nonlocalsimpl}
  \begin{aligned}
   O'=V(\mathbf{11^{T}}+K^T Q)\rho.
  \end{aligned}
\end{eqnarray}

This linear form is similar to Non-local Dot product module:
\begin{eqnarray}\label{eq:Non-local Dot product}
 O'=VK^TQ/(HW).
\end{eqnarray}

The two differences between them are: $\lambda$ includes an additional term $\mathbf{1}_{HW}^{T}K^T Q$ and $O'$ includes a term $V\mathbf{11^{T}}\rho$. To understand the effect of the first term, we conduct experiments where we set $\lambda=1./(HW\mathbf{1}_{HW}^{T})$. We found out that the mIoU fluctuates just about 0.02\%, which means that altering $\lambda$ is not critical for the performance. Based on this, a simpler approximation expression can be obtained:
\begin{eqnarray}\label{eq:attention_linear_1}
   O' =V\mathbf{11^{T}}/(HW)+V K^T Q/(HW).
\end{eqnarray}

Next, the global average pooling term $V\mathbf{11^{T}}/(HW)$ is removed. In this case, we observe that mIoU will decrease by $3.3\%$. However, this reduction is a result of the learned parameters being trained with the module in \eqref{eq:Non-local Embedded Gaussian} rather than \eqref{eq:Non-local Dot product}. In fact, noting that removing the global average pooling term $V\mathbf{11^{T}}/(HW)$ simply leads to the non-local dot product module in \eqref{eq:Non-local Dot product} and according to the previous studies on the latter \cite{wang2018non}, we know that retraining the network with the model in \eqref{eq:Non-local Dot product} will recover the loss. In summary, all observations indicate that \eqref{eq:Non-local Dot product} can be used as a base to obtain our framework in \eqref{eq:green_path_grouping} by a series of empirically justified simplifications. 
As a final note, the purely multi-linear form in \eqref{eq:Non-local Dot product} brings another benefit. It can be simultaneously viewed as both spatial attention and channel attention depending on the order of operations. Computing the product of the leftmost matrices first leads to channel self-attention, while computing the product of the rightmost matrices first leads to spatial self-attention. This observation makes the non-local dot product module adaptable to different forms of attention by parameter adjustment.
\myPara{Normalized LinSoftmax self-attention structure}
\label{sec:strategy2}
Although the empirical performance of the non-local dot product form is often satisfactory, its stability in a network other than CNN may not be guaranteed. For example, we found that stacking too many non-local dot product forms in ViT\cite{dosovitskiy2020image} leads to non-convergent training. To overcome this limitation, we consider another strategy of selecting $\lambda_k$ and $\lambda_q$ as the element-wise inverse of the $\ell_{2}$ norms of the $\bq$ and $\bk$ vectors. In this case, we make sure that $\mathbf{q}\cdot \mathbf{k}$ is not only strictly restricted to $[-1\ 1]$, but also represents a cosine similarity metric. This form of self-attention has also been considered in Linear attention\cite{li2020linear}. Based on the pre-implemented $\ell_2$ normalization, self-attention can also be expressed as the product of matrices, as we elaborate.
First, we denote by $\|Q\|_2,\|K\|_2$ the vectors respectively containing the $\ell_2$ norms of the columns of $Q$ and $K$ (i.e. along the channel dimension)\footnote{Note that $\|\ldotp\|_2$ does not denote the operator norm of a matrix and its output is a vector.}, and then define:
\newcommand{\diag}{\mathrm{diag}}
\begin{eqnarray}
\lambda_{q}=1./\left\|Q\right\|_2,\quad \lambda_{k}=1./\left\|K\right\|_2.
\end{eqnarray}

We further define $\rho_{q}=\diag(\lambda_{q}), \quad \rho_{k}=\diag(\lambda_{k})$. In order to implement summation normalization, we remind:
\begin{eqnarray}
\rho=\diag(1./\left(\mathbf{1}_{HW}^{T}[\mathbf{1}\mathbf{1}^{T}+(K\rho_{k})^{T}(Q\rho_{q})])
\right) 
\end{eqnarray}

Through empirical experiments, we found that replacing dividing by sum with dividing by the length of the key sequence does not cause performance differences of more than 0.02\%. We may also simply take $\rho=\mathrm{diag}(1./(HW\mathbf{1}_{HW}^{T}))$. Then, we can calculate the output by:
\begin{eqnarray}\label{eq:LinSoftmax_l2_reg}
\begin{aligned}
   O^{'}=V[\mathbf{1}\mathbf{1}^{T}+(K\rho_{k})^{T}(Q\rho_{q})]/(HW).
\end{aligned}
\end{eqnarray}

\subsection{Replacing input with low frequency components}
Based on the above, two structures for frequency self-attention are obtained. Next, we need to replace the input by low frequency components. This is obtained by using $\mathbf{G}=\mathbf{P}^T$ and replacing $Q$, $K$ and $V$ with their low-pass filtered version $Q_f=W_{q}f'\mathbf{P}^T$, $K_f=W_{k}f'\mathbf{P}^T$ and $V_f=W_{v}f'\mathbf{P}^T$.
Finally, applying $\mathbf{P}^{T}\mathbf{P}=E$, where $E$ is the identity matrix, to the Non-local Dot-product structure in \eqref{eq:Non-local Dot product}, a series of procedures depicted in the green path in \figref{fig:idealgoal} is obtained, which is written as:
\begin{eqnarray}\label{eq:fsa_dot_PG}
O'=(W_{v}f')(W_{k}f')^{T}(W_{q}f')/(HW) \mathbf{P}^T.
\end{eqnarray}

For this situation, $f'=X'\mathbf{P}$ corresponds to Linear operation 1, $(W_{v}f')(W_{k}f')^{T}(W_{q}f')/(HW)$ corresponds to the frequency self-attention and the operation of multiplying $\mathbf{P}^T$ corresponds to Linear operation 2. We refer to this version of frequency self-attention as FsaNet-Dot, which is shown in \figref{fig:flowchartnonlocal}.

Similarly, we replace the $Q$, $K$, and $V$ in the Normalized LinSoftmax self-attention structure in \eqref{eq:LinSoftmax_l2_reg}, which leads to a sequence of steps depicted in the green path in \figref{fig:idealgoal}, and given by:
\begin{eqnarray}\label{eq:lowfre_spation_attention_lin}
\begin{small}
\begin{aligned}
&\rho_{q}=\diag(1./\left\|W_{q}f^{'}\mathbf{P}^T\right\|_2),\quad \rho_{k}=\diag(1./\left\|W_{k}f^{'}\mathbf{P}^T\right\|_2)\\
&\rho=1/(HW)\\
&O^{'}=(W_{v}f^{'}(\mathbf{P}^{T}\rho_{k}\mathbf{P})(W_{k}f^{'})^{T}(W_{q}f^{'})\mathbf{P}^{T}\rho_{q}+W_{v}f^{'}\mathbf{P}^{T}\mathbf{1}\mathbf{1}_{HW}^{T})\rho.
\end{aligned}
\end{small}
\end{eqnarray}

In this case, $f'=X'\mathbf{P}$ corresponds to Linear operation 1, $W_{v}f^{'}(\mathbf{P}^{T}\rho_{k}\mathbf{P})(W_{k}f^{'})^{T}(W_{q}f^{'})$ corresponds to the frequency self-attention unit and the operations of multiplying $\mathbf{P}^{T}\rho_{q}$, adding $W_{v}f^{'}\mathbf{P}^T\mathbf{1}\mathbf{1}_{HW}^{T}$, and dividing $HW$ correspond to Linear operation 2. We refer to this version of frequency self-attention as FsaNet-Lin, and its entire process is shown in \figref{fig:flowchartlinsoftmax}.
\CheckRmv{
\begin{figure}[t]
  \begin{overpic}[width=0.83\linewidth]{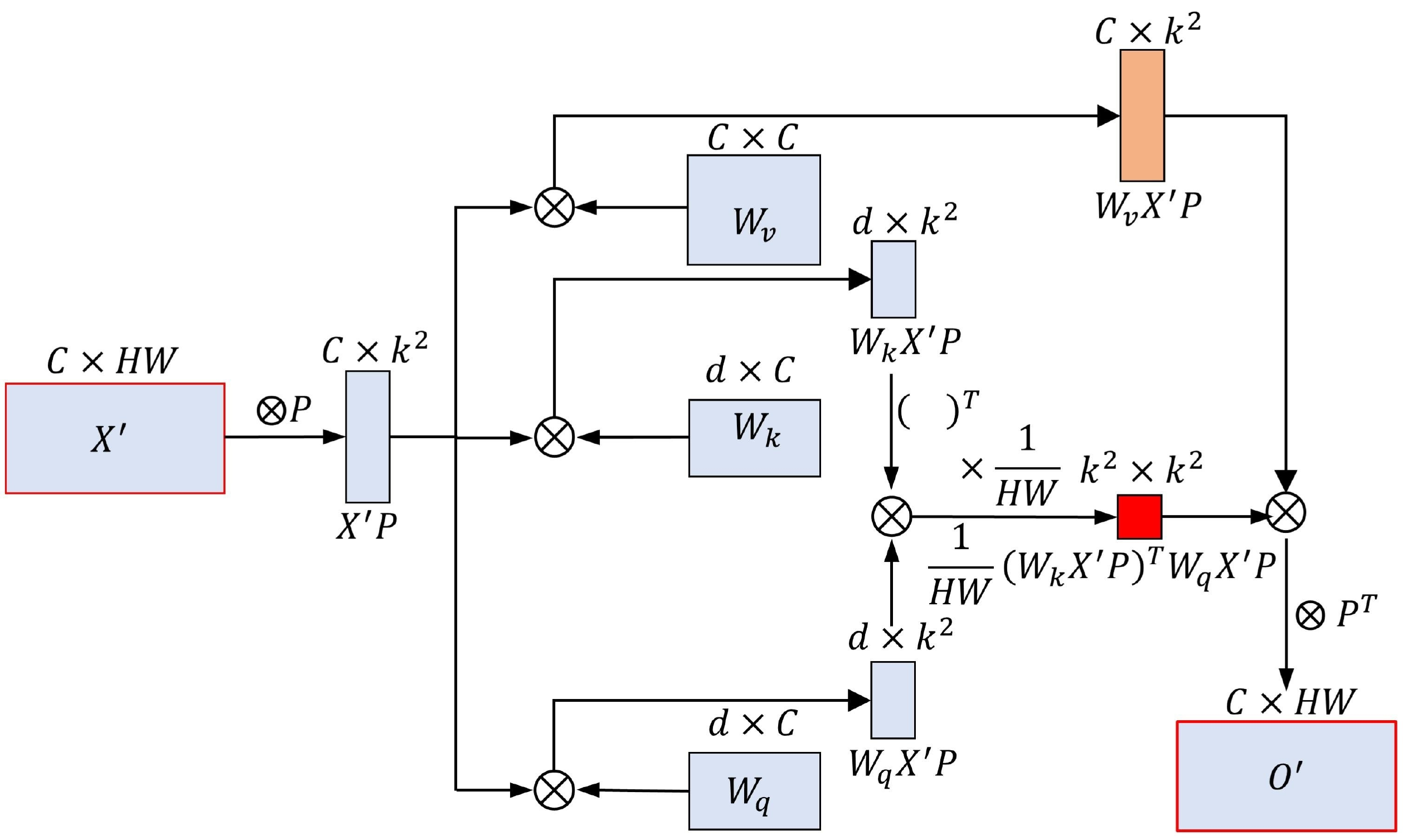}
  \end{overpic}
  \caption{Calculation process for FsaNet-Dot}
  \label{fig:flowchartnonlocal}
\end{figure}
}
\CheckRmv{
\begin{figure}[t]
  \begin{overpic}[width=\linewidth]{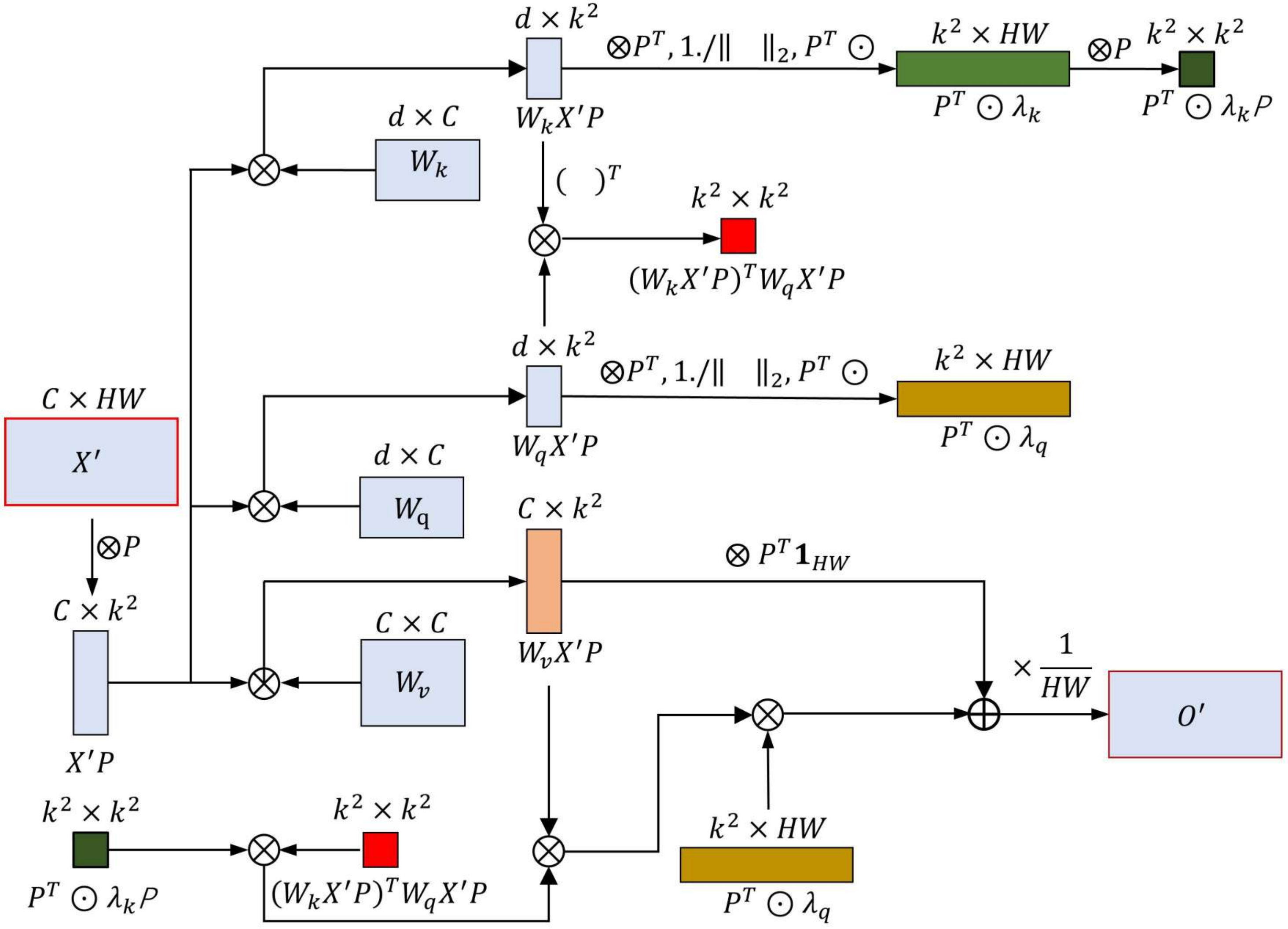}
  \end{overpic}
  \caption{Calculation process for FsaNet-Lin}
  \label{fig:flowchartlinsoftmax}
\end{figure}
}

\section{Experiments}\label{sec:experiments}
In the following subsections, we first introduce the datasets and a semantic segmentation toolbox. Then, we introduce ablation experiments on the high-frequency components to justify our methodology and further determine the required number $k$ of low-frequency components. Finally, we show the implementation details and performance experiments.
\subsection{Datasets and Toolbox}
We adopt mIoU (mean intersection over union) for Cityscapes~\cite{cordts2016cityscapes}, ADE20K \cite{zhou2019semantic} and VOCaug\cite{pascal-voc-2012}, and AP (Average Precision) for COCO\cite{lin2014microsoft} as evaluation metrics. MMSegmentation~\cite{mmseg2020} toolbox is used to maintain a fair performance comparison. \\
• Cityscapes is a dataset of urban street scenes with 34 categories. However, existing studies mainly focus on only 19 categories. There are 5,000 finely annotated images, which are divided into 2,975/500/1,525 images for training, validation, and testing, respectively. Besides, extra 19997 images are provided with only a coarse annotation.\\
• ADE20K is a scene dataset containing dense labels of 150 stuff/object categories. The scenes are relatively rich: indoor, outdoor, natural scenes, etc. The dataset includes 20k/2k images for training and validation, respectively. \\
• VOCaug (PASCAL VOC 2012 Augmented Dataset) mainly contains complex daily scenes with 20 categories. Most recent works on Pascal VOC 2012 usually exploit extra augmentation data. The augmented dataset includes 10,582/1,449 images for training and validation, respectively. \\
• COCO is a dataset for instance segmentation that contains 115k images over 80 categories for training, 5k images for validation, and 20k images for testing.\\
• The Mmsegmentation toolbox provides a unified framework for comparing \sArt and popular methods. For fairness, all hyperparameters remain the same in all supported methods, except for the difference in their own structural characteristics. Note that the Cityscapes suggest a slightly different approach for calculating the mIoU than the conventional method. There is often a small gap ($\sim0.1\%$) between the two, and we adopt the Cityscapes instructions.

\subsection{Ablation study for self-attention/Non-local module}\label{sec:experiments:ablation}

\myPara{Ablation study on biases for Non-local module}\label{sec:experiments:bias ablation}
Here we establish that the biases in the three $1\times 1$ convolutions to obtain $Q$, $K$ and $V$ have no substantial effect, which is also the reason why they are removed from the previous formulas. In order to verify the above points, we implemented an ablation study based on a series of trained Non-local models officially provided by MMSegmentation\cite{mmseg2020}. These twelve models are all trained in the Embedded Gaussian mode with different datasets, backbones and learning rate update schemes. Usually, we need to retrain the model to perform the ablation study, because even if useless parameters or frequency components are removed from the model, the performance may be severely degraded without retraining. On the other hand, if some parameters or frequency components are removed but have little effect on performance without retraining, they can be removed. All ablation studies in this paper are performed without retraining, still demonstrating the ineffectiveness of the 3 biases and high-frequency components.
We directly remove the three biases at the same time, and the comparison with the original results is shown in \tabref{biastab}. We observe that the performance of most of the models degrades by no more than 0.03, only one model drops by 0.05, and two models even improve their performance.
\CheckRmv{
    \begin{table}[tbp]
    \tiny
    \scriptsize
    \tabFormat
    \setlength{\tabcolsep}{2.2mm}
    \caption{Ablation study on Biases.
    }
	\begin{tabular}{lcccc}
	\toprule
	Dataset & Backbone & Schedule & Bias & Nobias\\
	\midrule
	\multirow{4}{*}{Cityscapes} & ResNet-50 & 40k & 78.21 & 78.23$\textcolor{red}{\uparrow}_{0.02}$ \\
	~ & ResNet-50 & 80k &78.86 &78.88$\textcolor{red}{\uparrow}_{0.02}$ \\
	~ & ResNet-101 & 40k &78.51 &78.49$\textcolor{green}{\downarrow}_{0.02}$ \\
	~ & ResNet-101 & 80k &79.35 &79.35$\textcolor{red}{\uparrow}_{0.00}$\\
	\midrule
	\multirow{4}*{ADE20k} & ResNet-50 & 80k & 40.75 & 40.72$\textcolor{green}{\downarrow}_{0.03}$ \\
	~ & ResNet-50 & 160k & 42.03 & 42.01$\textcolor{green}{\downarrow}_{0.02}$ \\
	~ & ResNet-101 & 80k & 42.90 & 42.86$\textcolor{green}{\downarrow}_{0.04}$ \\
	~ & ResNet-101 & 160k & 44.63 & 44.62$\textcolor{green}{\downarrow}_{0.01}$ \\
	\midrule
	\multirow{4}*{VOCaug} & ResNet-50 & 20k & 76.20 & 76.15$\textcolor{green}{\downarrow}_{0.05}$ \\
	~ & ResNet-50 & 40k & 76.65 & 76.62$\textcolor{green}{\downarrow}_{0.03}$ \\
	~ & ResNet-101 & 20k & 78.15 & 78.12$\textcolor{green}{\downarrow}_{0.03}$ \\
	~ & ResNet-101 & 40k & 78.27 & 78.24$\textcolor{green}{\downarrow}_{0.03}$ \\
	\bottomrule
    \end{tabular}
    \label{biastab}	
    \end{table}
}
\myPara{Ablation study on input frequency components for Non-local module}\label{sec:experiments:frequency ablation}
\renewcommand{\addFig}[1]{{\includegraphics[width=.33\linewidth]{#1.pdf}}}
\CheckRmv{
\begin{figure*}[t]
  \centering
  \tiny
  \renewcommand{\arraystretch}{0.5}
  \setlength{\tabcolsep}{0.2mm}
  \begin{tabular}{ccc}
    \addFig{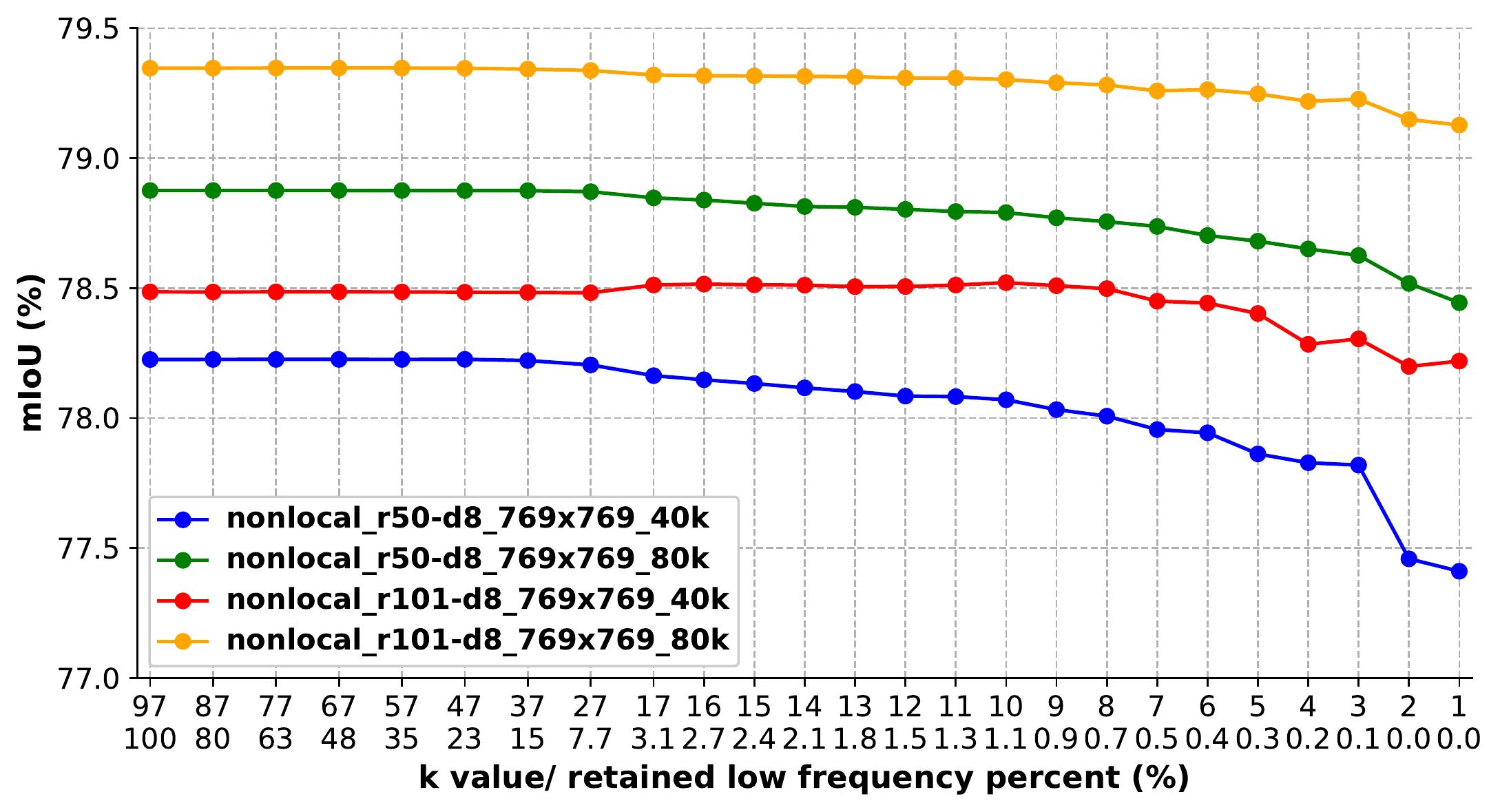}&
    \addFig{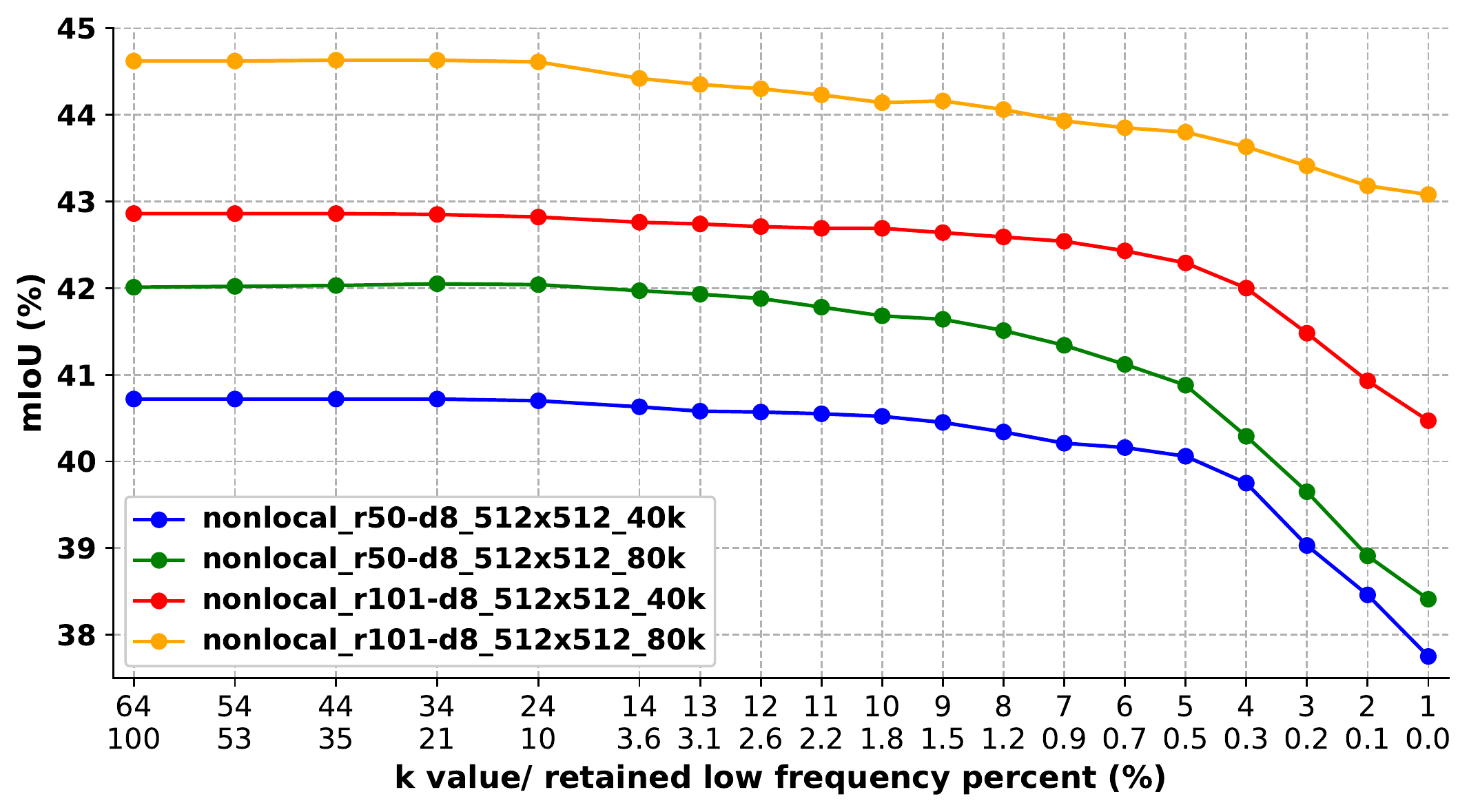}&
    \addFig{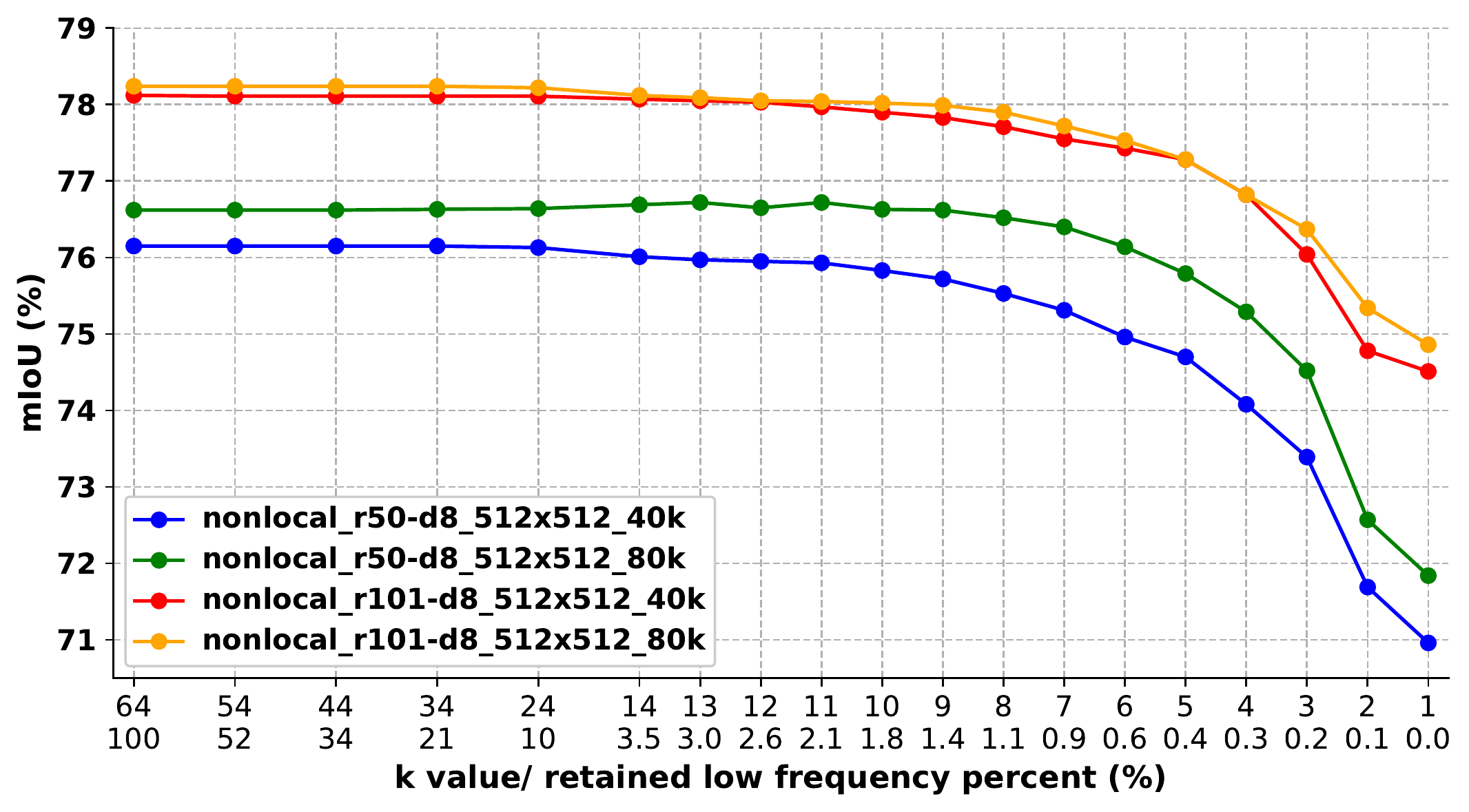}\\
  \end{tabular}
  \caption{mIoU varies with preserved low frequencies on Cityscapes, ADE20K and VOCaug.}
  \label{fig:cityfre}
\end{figure*}
}
We conducted follow-up experiments with biases removed. To analyze the effect of retaining different low-frequency components on the performance, we transform the input to the frequency domain, take the low frequency blocks of size $(k, k)$ and reconstruct it as the new input for self-attention. The curves of performance versus $k$ value on three different datasets are shown in \figref{fig:cityfre}. We use two scales to display the curves and marked the percentage of retained low-frequency components corresponding to different $k$ values.
With the reduction of the retained frequency components, the performance drops very slowly, and sometimes even slightly rises. This implies that the learned parameters tend to cater to the low-frequency components, even though the network is trained with full frequency. Moreover, we observe that the curves with high performance declined more slowly than the curves with low performance. This also shows that the parameters with good performance tend to adapt to low frequencies, so the performance is easier to maintain when a few low-frequency components are retained. As seen in \figref{fig:cityfre}, there is considerable redundancy in the calculation of spatial self-attention, so it is feasible to reduce the number of tokens from $HW$ to $k^2$.

To observe the effect of self-attention on the processing of different frequency components, one channel of the feature maps before and after processing is randomly selected as shown in \figref{fig:highlowfreprocess}. For aesthetic purposes, single-channel images are displayed with a perceptually uniform sequential colormap. We observe that the coherence of the subject is enhanced after the self-attention processing of the low-frequency components, while the edges of the high-frequency components become more blurred after the identity processing.
\renewcommand{\addFig}[1]{{\includegraphics[width=.115\textwidth]{#1.png}}}
\CheckRmv{
\begin{figure}[t]
  \centering
  \tiny
  \renewcommand{\arraystretch}{0.3}
  \setlength{\tabcolsep}{0.03mm}
  \begin{tabular}{lcccc}
    \rotatebox[origin=l]{90}{before on Cityscapes}&
    \addFig{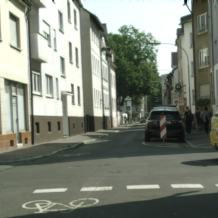}&
    \addFig{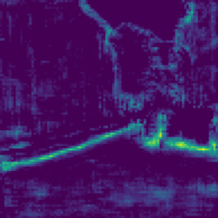}&
    \addFig{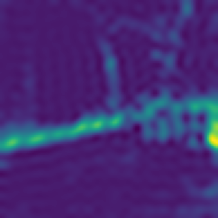}&
    \addFig{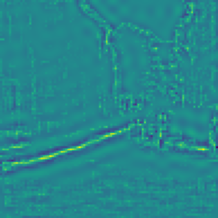}\\
    \rotatebox[origin=l]{90}{after on Cityscapes}&
    \addFig{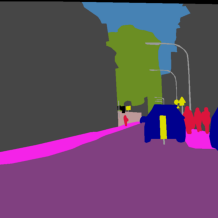}&
    \addFig{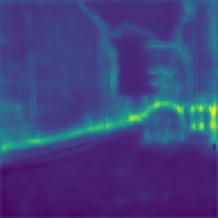}&
    \addFig{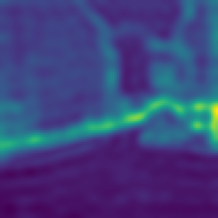}&
    \addFig{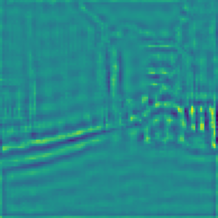}
    \\
    \rotatebox[origin=l]{90}{before on ADE20K}&
    \addFig{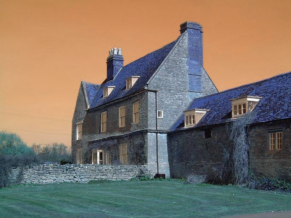}&
    \addFig{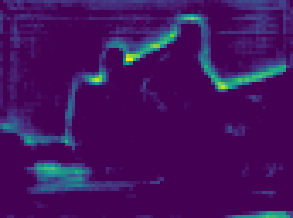}&
    \addFig{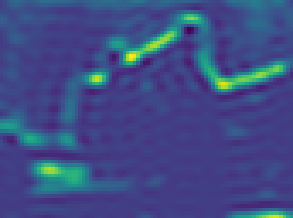}&
    \addFig{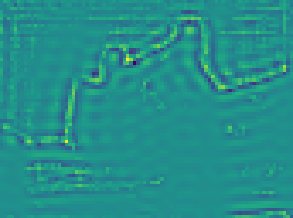}\\
    \rotatebox[origin=l]{90}{after on ADE20K}&
    \addFig{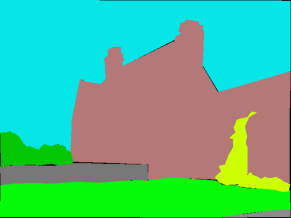}&
    \addFig{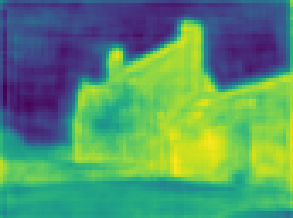}&
    \addFig{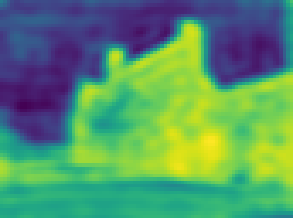}&
    \addFig{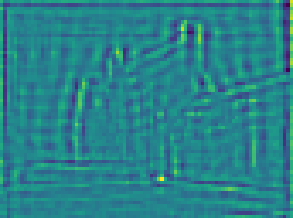}
    \\
    \rotatebox[origin=l]{90}{before on VOCaug}&
    \addFig{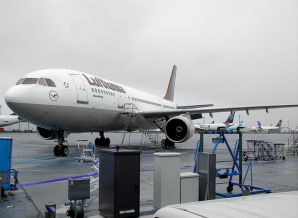}&
    \addFig{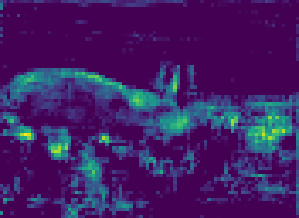}&
    \addFig{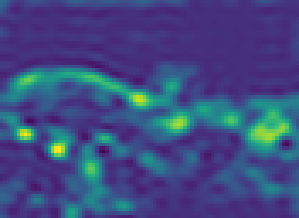}&
    \addFig{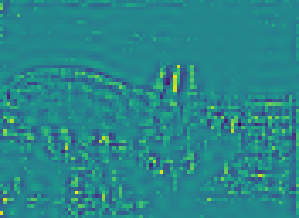} \\
    \rotatebox[origin=l]{90}{after on VOCaug}&
    \addFig{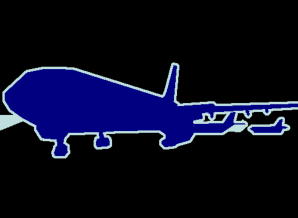}&
    \addFig{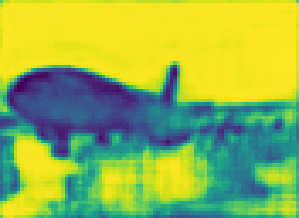}&
    \addFig{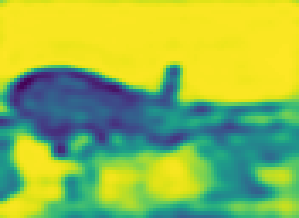}&
    \addFig{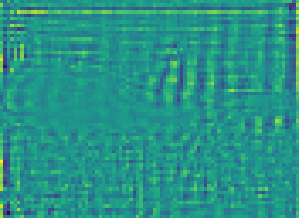}\\
   &Image and GT & Full frequency & Low frequency & High frequency \\
  \end{tabular}
  \caption{High and low frequencies are separately processed.}
  \label{fig:highlowfreprocess}
\end{figure}
}

\subsection{Visualization of differences caused by linearization}
\renewcommand{\addFig}[1]{{\includegraphics[width=.24\textwidth]{#1.pdf}}}
\CheckRmv{
\begin{figure}[t]
  \setlength{\tabcolsep}{0.0mm}
  \begin{tabular}{ccc}
  \addFig{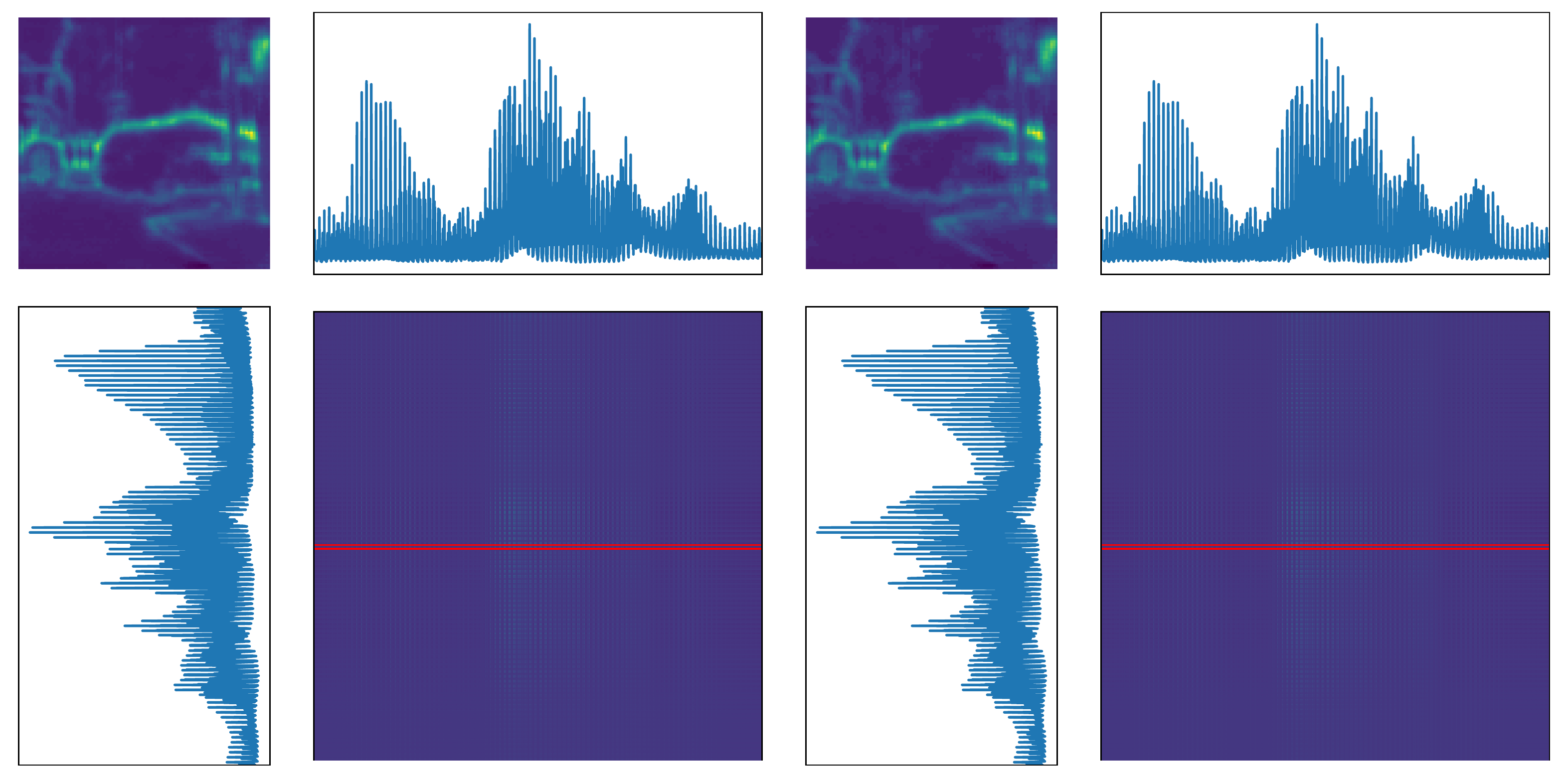} &  &
  \addFig{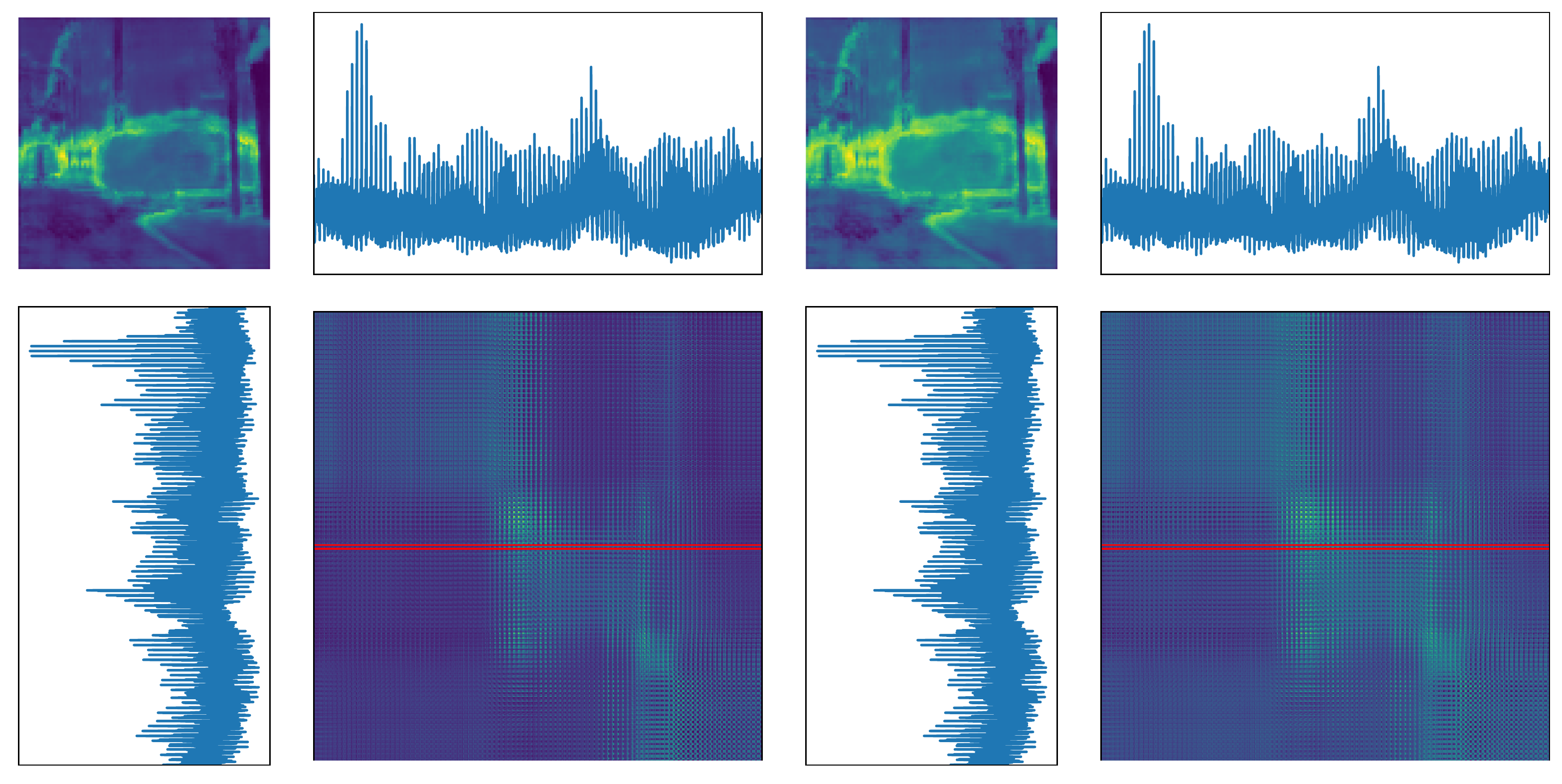}\\
  (a) & \; & (b)
  \end{tabular}
  \caption{For each figure, the attention maps of all points are displayed in the lower right corner, where each row represents the attention map of one single point against all other points. The attention map of the point marked by the red line is displayed in the upper left corner. The $\ell_2$ norm of the $HW$ columns in $Q$ is shown in the lower left corner. The $\ell_2$ norm of the $HW$ columns in $K$ is shown in the upper right corner. 
  }
  \label{fig:Softmax_normLinSoftmax}
\end{figure}
}
In the derivation of FsaNet-Dot, we approximated $\mathrm{Softmax}(K^TQ)$ with $(\mathbf{11^{T}}+K^{T}Q)/(HW)$. To further understand the difference in such linearized approximation, we trained a CNN network with self-attention whose attention map is calculated by $\mathrm{Softmax}(K^TQ)$. From this well-trained network, we achieved a sample of $Q$ and $K$. Based on the same $Q$ and $K$, two attention maps are visualized in \figref{fig:Softmax_normLinSoftmax}(a). The attention map in the left figure is calculated by $Softmax(K^TQ)$, and the attention map in the right figure is calculated by $(\mathbf{11^{T}}+K^{T}Q)/(HW)$. It is seen that there is a slight change among the entire attention map, but for each point, the attention map is almost unchanged. This suggests that even if the linear structure is not adopted by self-attention placed after CNN backbone, it will still become linear adaptively.

To understand the differences in the linearized approximation in FsaNet-Lin, we also trained a CNN network with the attention map computed by $\mathrm{Softmax}((K\rho_{k})^{T}(Q\rho_{q}))$. Based on the same $Q$, $K$ achieved from this well-trained network, two kinds of attention maps calculated by $\mathrm{Softmax}((K\rho_{k})^{T}(Q\rho_{q}))$ and $(\mathbf{11^{T}}+(K\rho_{k})^{T}(Q\rho_{q}))/(HW)$ are visualized in \figref{fig:Softmax_normLinSoftmax}(b) from left to right. It is seen that the weight distribution of attention is basically unchanged, although the difference between the weights has narrowed somewhat.

Even when different global attention modules are used in the case of independent training, there is a high similarity between \figref{fig:Softmax_normLinSoftmax}(a) and (b), indicating a consistent focus on global attention. 

\subsection{Implementation Details}
All experiments are implemented on 4 Tesla V100 GPUs with ResNet101 as the backbone. We realize \ourM by replacing the non-local module with our frequency self-attention module in Non-local neural network implemented by MMSegmentation\cite{mmseg2020}. The poly learning rate is adopted and calculated by
\newcommand{\baselr}{\mathrm{lr}_\mathrm{base}}
\newcommand{\minlr}{\mathrm{lr}_\mathrm{min}}
\newcommand{\iter}{\mathrm{iter}}
\newcommand{\iterm}{\mathrm{iter}_\mathrm{max}}
$(\baselr - \minlr)*(1-\frac{\iter}{\iterm})^p+ \minlr$ with $\baselr=1e{-2}$, $p=0.9$ and $\minlr=1e{-4}$. For Cityscapes and ADE20K, We set $\iterm=8000$, and set $\iterm=4000$ for VOC. The momentum parameter is 0.9 and the weight decay is 0.0005. The crop size is $769\times 769$ for Cityscapes and $512\times 512$ for ADE20K and VOC. The batch size is 8 for Cityscapes, and is 16 for ADE20K and VOC. We follow \CCNet~so that the input is cyclically processed by one self-attention module $R$ times. To confirm whether $P$ and $G$ can be gained in a learning way, we also added an experiment based on FsaNet-Dot, setting $P$ and $G$ as learnable parameters with $R=1$.
Further, we perform an  experiment on Cityscapes test set. To improve our method compared to other ResNet101-based methods, we adopt similar procedural steps to \CCNet. For convenience, we replaced the  criss-cross module in \CCNet~with our frequency self-attention module. In this way, the continuity loss in \CCNet~is applied in FsaNet. The backbone is still ResNet-101 but its last two down-sampling operations are removed, and dilated convolutions are employed in the subsequent convolutional layers. We also employ coarsely labeled data, increase the crop size to $849\times 849$ and set $R=2$ and $k=8$. The poly learning rate is calculated by $\baselr*(1-\frac{\iter}{\iterm})^{p}$ with $\baselr=1e{-2}$, $p=0.9$. After 6000 steps of training on the finely labeled data set, the ohem(Online Hard Example Mining) is utilized, and then the coarsely labeled data is added for mixed training for 30k steps. The momentum parameter is 0.9 and the weight decay is 0.0005. Notably, we add a set of learnable biases to the output of self-attention, as we discovered this modification to be beneficial to performance. In the implementation of MMSegmentation\cite{mmseg2020}, this bias already exists since the output of self-attention is processed by an extra output convolution. 

\subsection{FsaNet Performance on Cityscapes}\label{sec:Experiments_City}
\textbf{Performance on Cityscapes val set}
Results of \ourM and other methods on Cityscapes val set in MMSegmentation\cite{mmseg2020} are summarized in \tabref{citymiouTab}. We refer to FsaNet-Dot at $R=1$ and $R=2$ as FsaNet-Dot-R1 and FsaNet-Dot-R2, respectively, and FsaNet-Lin at $R=1$ and $R=2$ as FsaNet-Lin-R1 and FsaNet-Lin-R2, respectively. We achieved better results with $R=2$. Considering the influence of random factors, FsaNet-Dot and FsaNet-Lin have small differences, which is consistent with the conclusion that different versions of the non-local modules achieve similar results. The $P$ and $G$ derived from the DCT show a notable superiority over the learned $P$ and $G$, with the latter showing a 3.79\% performance drop relative to the former. It may be attributed to the fact that the semantic segmentation dataset is not large enough to train $P$ and $G$ adequately. Since sparsity in the frequency domain is a common pattern shared by all images, using this prior, there is no need to capture a low-rank pattern by large-scale training. Compared to the \FCN~without context module, \ourM has an improvement of 4.62$\sim$5.29\% in mIoU, which fully demonstrates its effectiveness. \ourM can easily outperform \NonlocalNetwork~by a large margin of 0.62$\sim$1.29\%. \ourM also showed apparent advantages against \DNLNet~by 0.69$\sim$1.36\%, which is also a \NonlocalNetwork~improved method by decoupling boundary and the body.
\begin{table}[!tbp]
        \scriptsize
        \setlength{\tabcolsep}{0mm}
	\begin{minipage}[t]{0.28\linewidth}
        \vspace{0pt}
		\makeatletter\def\@captype{table}\makeatother
		\caption{Performance on Cityscapes (val).} 
		\begin{tabular}{lc}
        \toprule
         Method & mIoU(\%)\\
        \midrule
        \FCN	 & 75.35\\
        \EncNet & 75.97\\
        \APCNet & 78.32\\
        \ANN & 78.61\\
        \GCNet & 79.05\\
        \DMNet & 79.06\\
        \DNLNet & 79.28\\
        \CCNet & 79.31\\
        Non-local\cite{wang2018non} & 79.35\\
        \EMANet & 79.48\\
        \PSANet& 79.49\\
        \DeepLabVthree & 79.59\\
        \PSPNet & 79.69\\
        \DANet & 80.29\\
        \OCRNet & 80.33\\
        \ISANet & 80.47\\
        \midrule
        \ourM-LearnPG & 76.18\\
        \ourM-Dot-R1 & 79.97\\
        \ourM-Lin-R1 & 80.04\\
        \ourM-Dot-R2 & 80.36\\
        \ourM-Lin-R2 & \textbf{80.64}\\
        \bottomrule
	\end{tabular}
	\label{citymiouTab}
	\end{minipage}
        \quad\begin{minipage}[t]{0.28\linewidth}
        \vspace{0.17em}
        \makeatletter\def\@captype{table}\makeatother
		\caption{Performance on ADE20k.} 
		\begin{tabular}{lc}
        \toprule
        Method  & mIoU(\%)\\
        \midrule
        \FCN & 39.61\\
        \EncNet & 42.11\\
        \GCNet & 42.82\\
        \OCRNet & 42.87\\
        \DNLNet & 42.88\\
        Non-local\cite{wang2018non} & 42.90\\
        \UPerNet & 42.91\\
        \ANN	 & 42.94\\
        \ISANet	 & 43.51\\
        \PSPNet	 & 43.57\\
        \DANet	 & 43.64\\
        \CCNet	 & 43.71\\
        \PSANet	 & 43.80\\
        \DeepLabVthree & 44.08\\
        \midrule
        FsaNet-LearnPG & 42.24\\
        FsaNet-Dot-R1 & 43.04\\
        FsaNet-Lin-R1 & 43.05\\
        FsaNet-Dot-R2 & 43.53\\
        FsaNet-Lin-R2 &\textbf{44.10}\\
	\bottomrule
	\label{ADEmiouTab}
		\end{tabular}
	\end{minipage}
        \quad\begin{minipage}[t]{0.28\linewidth}
        \vspace{0.16em}
        \makeatletter\def\@captype{table}\makeatother
		\caption{Performance on VOCaug.}
		\begin{tabular}{lc}
        \toprule
        Method & mIoU(\%)\\
        \midrule
        \FCN& 69.91\\
        \DANet & 76.51\\
        \ANN& 76.70\\
        \UPerNet & 77.43\\
        \PSANet & 77.73\\
        \GCNet & 77.84\\
        \CCNet & 77.87\\
        \DeepLabVthree & 77.92\\
        \ISANet & 78.12\\
        \DNLNet & 78.18\\
        Non-local\cite{wang2018non} & 78.27\\
        \midrule
        FsaNet-LearnPG & 75.18\\
        FsaNet-Dot-R1 & 78.31\\
        FsaNet-Lin-R1 & 78.33\\
        FsaNet-Dot-R2 & 78.67\\
        FsaNet-Lin-R2 & \textbf{78.70}\\
        \bottomrule
        \label{VOCmiouTab}
		\end{tabular} 
	\end{minipage}
\vspace{-15pt}
\end{table}
We visualize the segmentation results of Non-local Network, \OCRNet, and our approaches on the Cityscapes validation set in \figref{fig:cityseg}. In the first row, Some pixels of the truck were wrongly attributed to bus or car in \NonlocalNetwork, \OCRNet, and FsaNet-R1. However, when \ourM is applied with $R=2$, the consistency within the truck is better promoted, and the whole truck is correctly identified. In the second row, the person and rider are marked in dark red and bright red, respectively. In \NonlocalNetwork~and \OCRNet, the head, chest and arms of the rider on the far right are misclassified as person. In FsaNet-R1, only a few pixels of the face and chest are misclassified. In FsaNet-R2, all the pixels of the rider are correctly identified. In the third row, \ourM is less prone to omissions in identifying yellow traffic signs. This shows that on small objects, \ourM better maintains  the details.
\renewcommand{\addFig}[1]{{\includegraphics[height=.06\textwidth]{#1.png}}}
\CheckRmv{
\begin{figure}[t]
  \centering
  \tiny
  \renewcommand{\arraystretch}{1}
  \setlength{\tabcolsep}{0.0mm}
  \begin{tabular}{cccc}
   \\
   \addFig{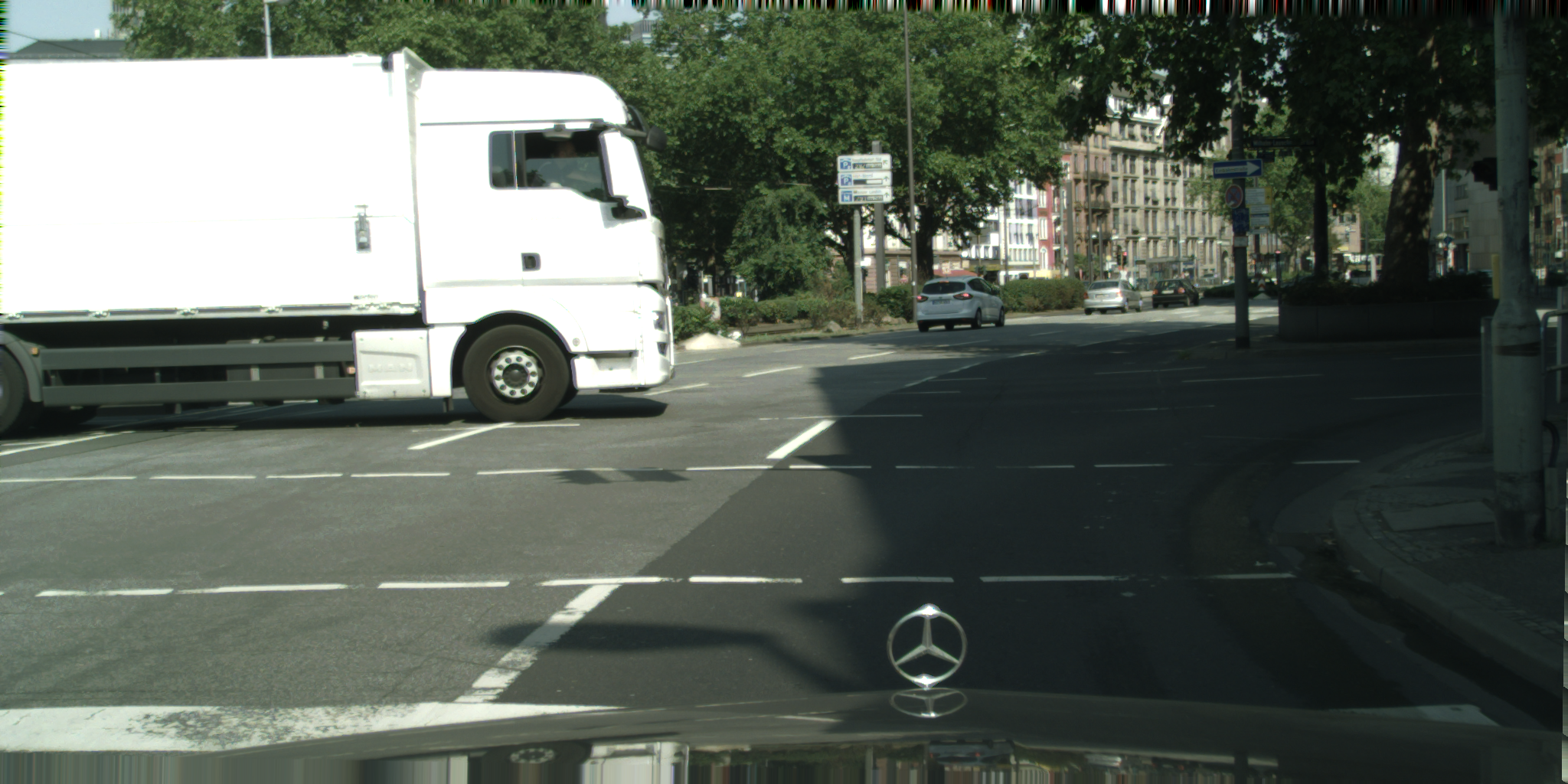}&
   \addFig{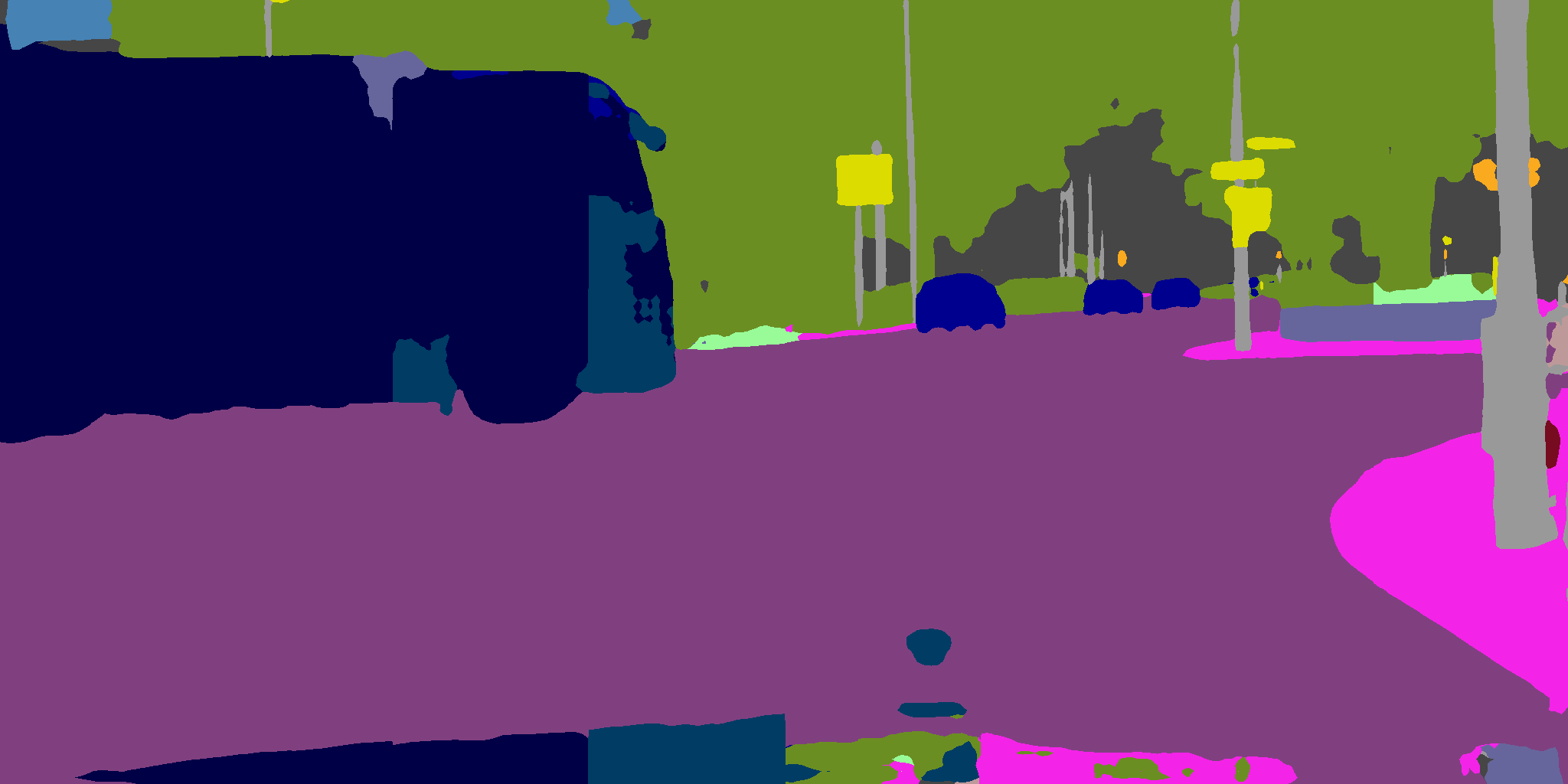}&
   \addFig{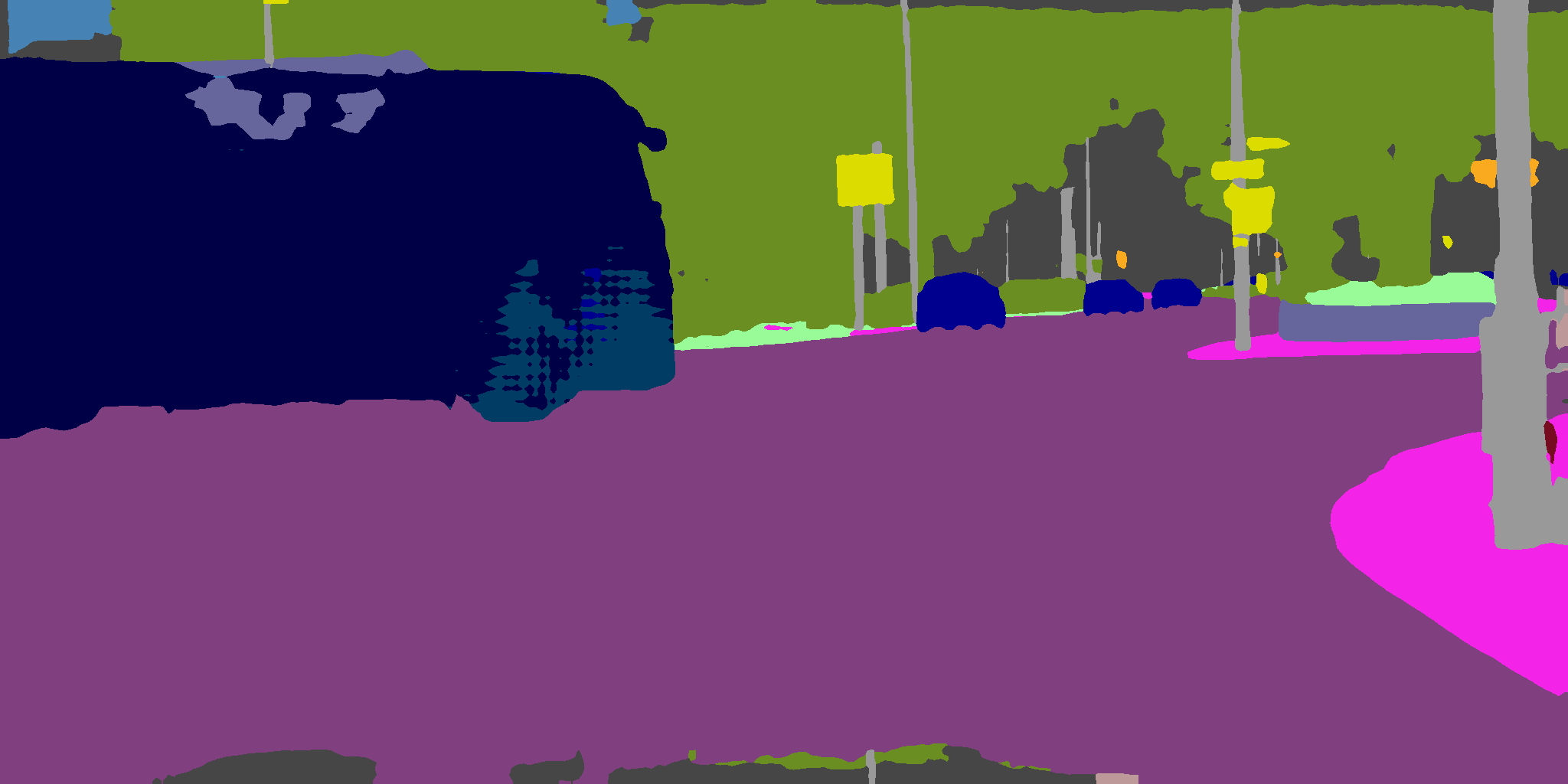}&
   \addFig{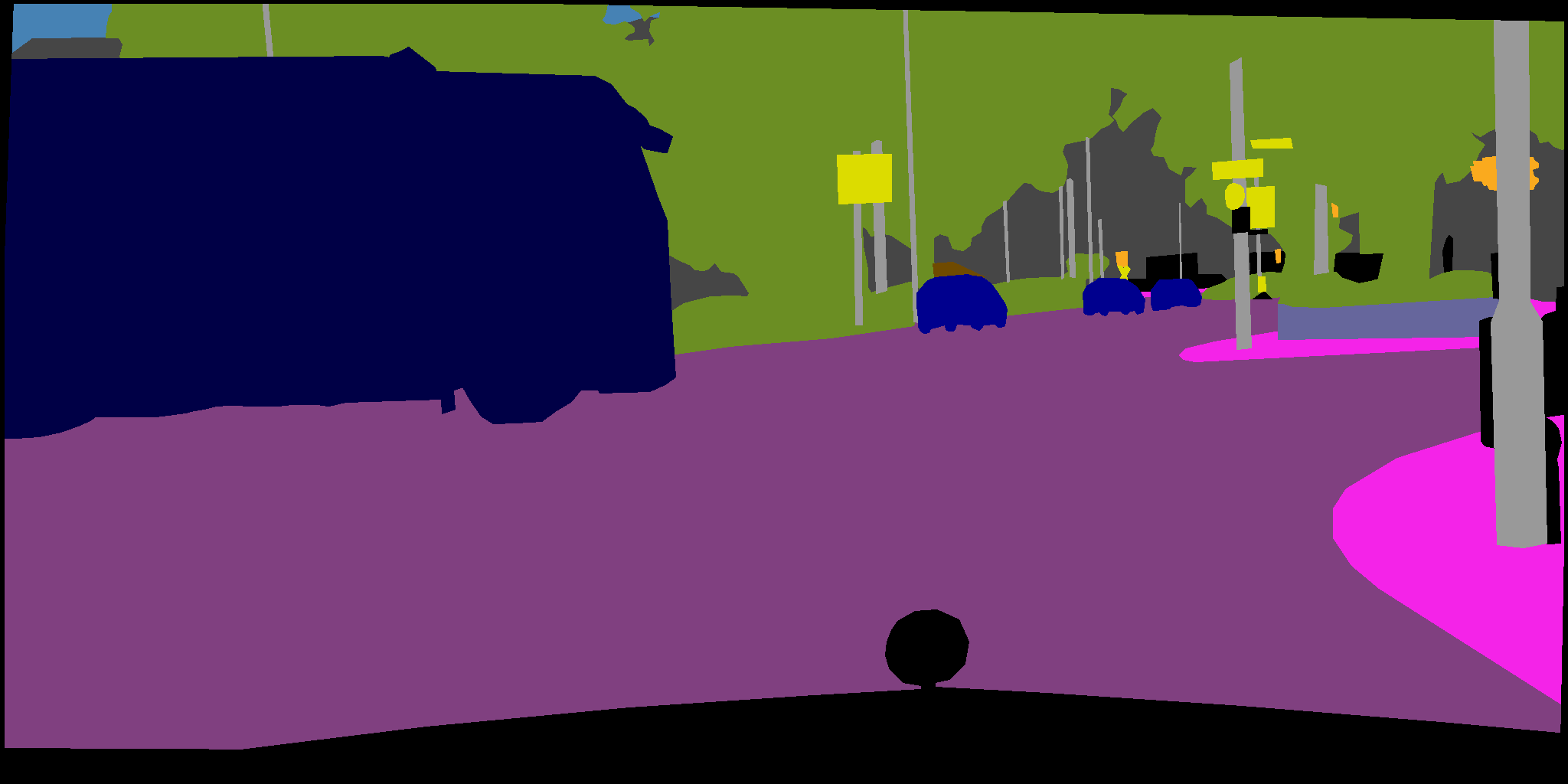}
   \\
   \addFig{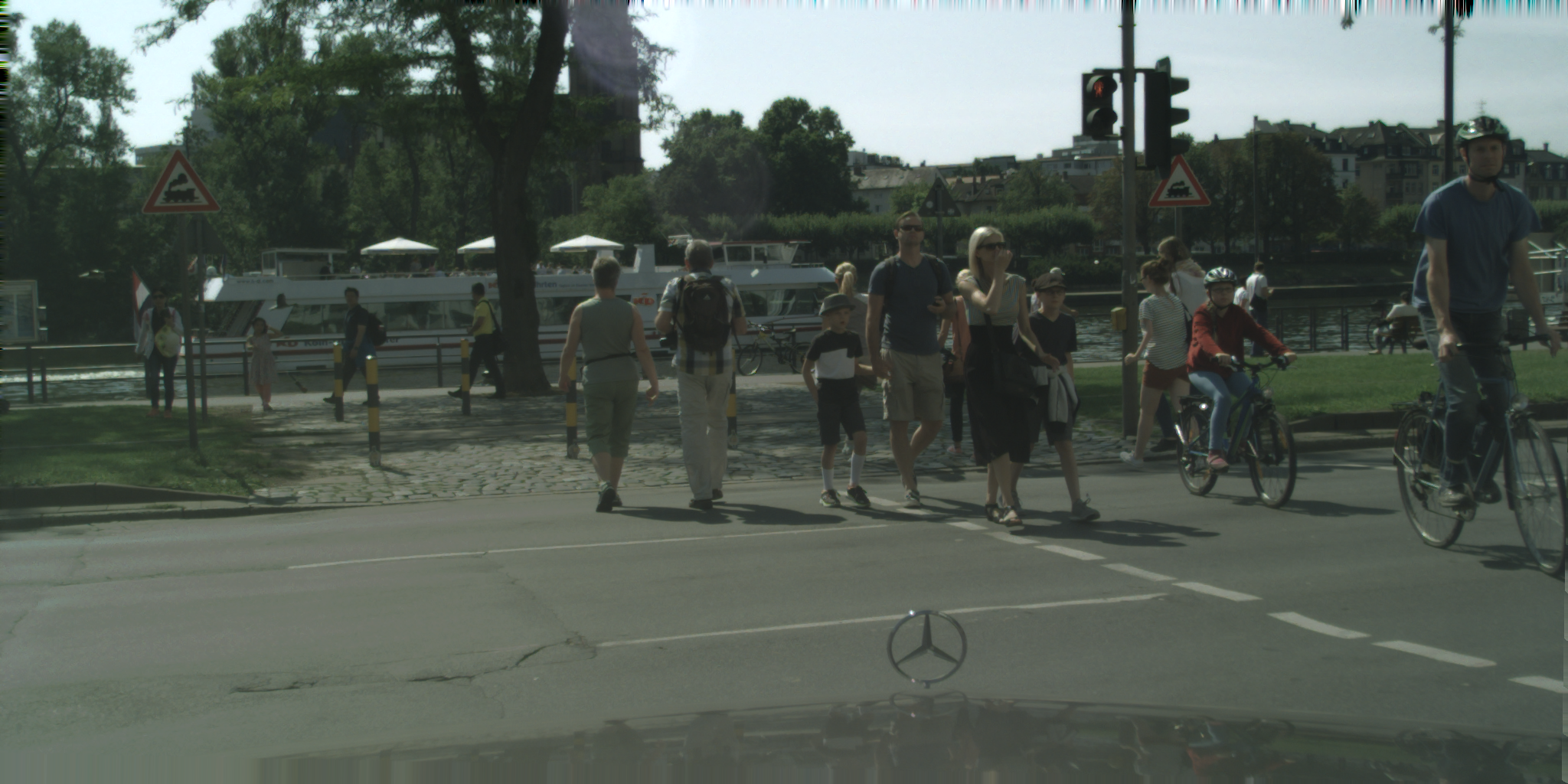}&
   \addFig{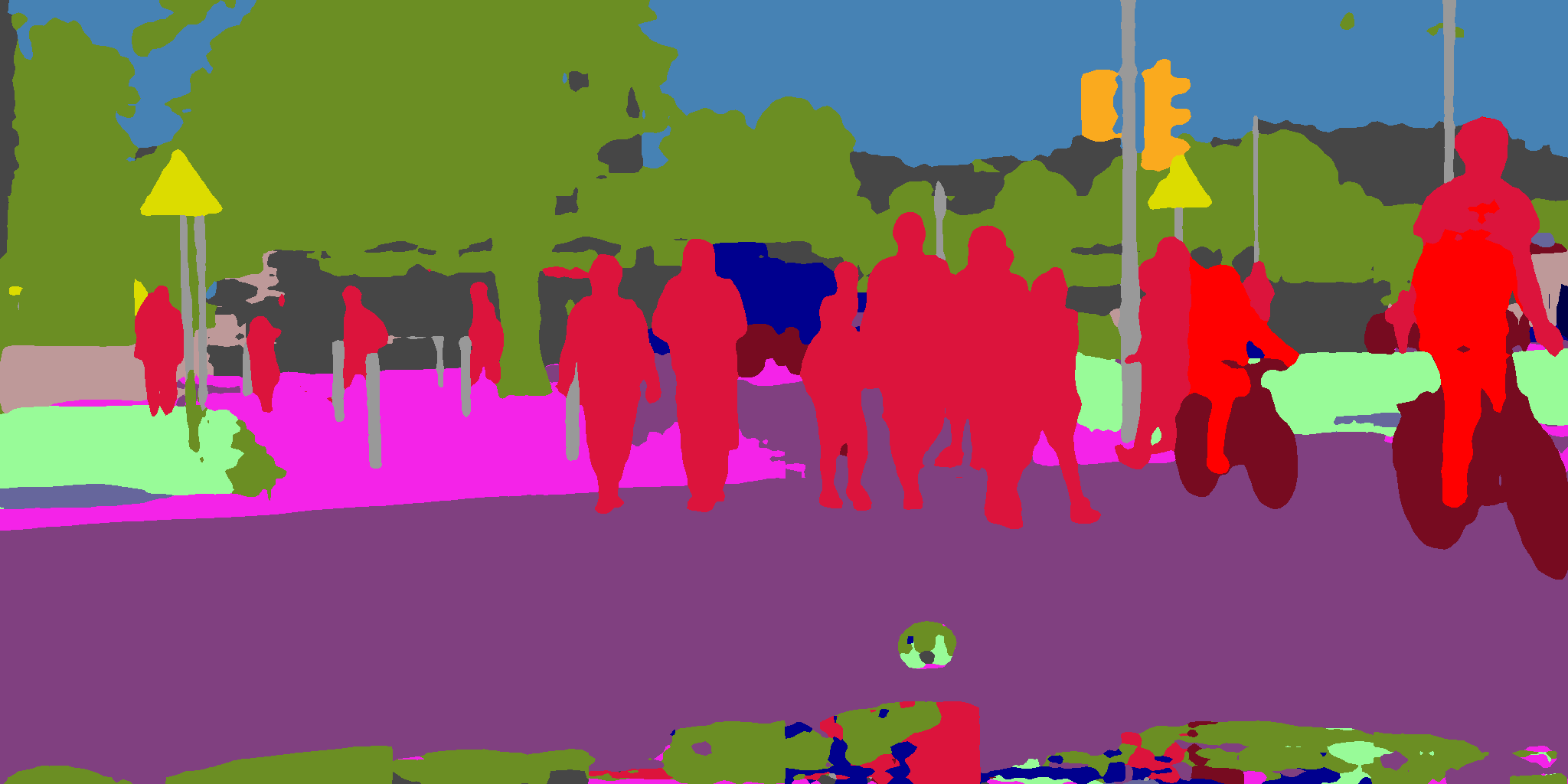}&
   \addFig{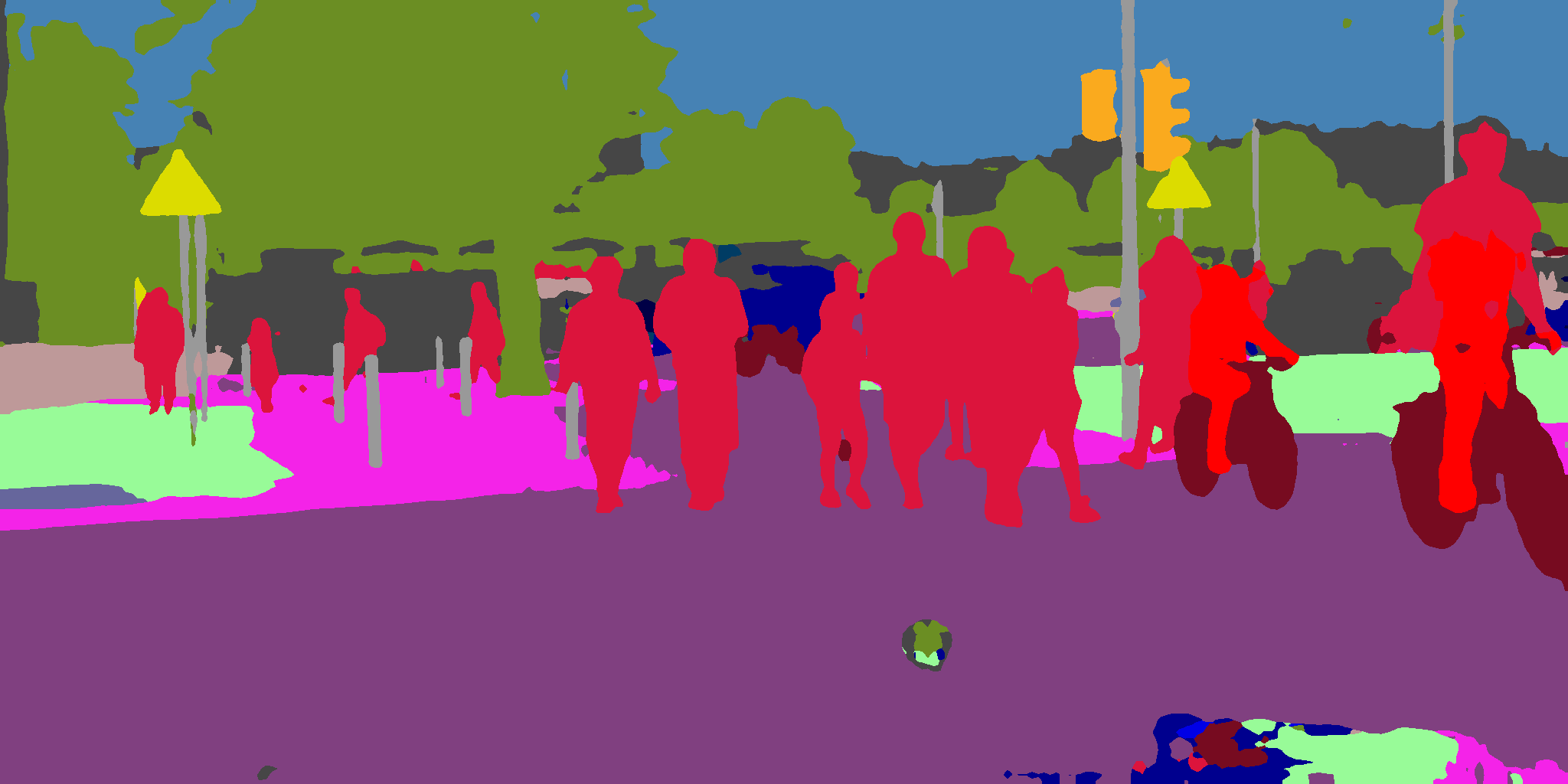}&
   \addFig{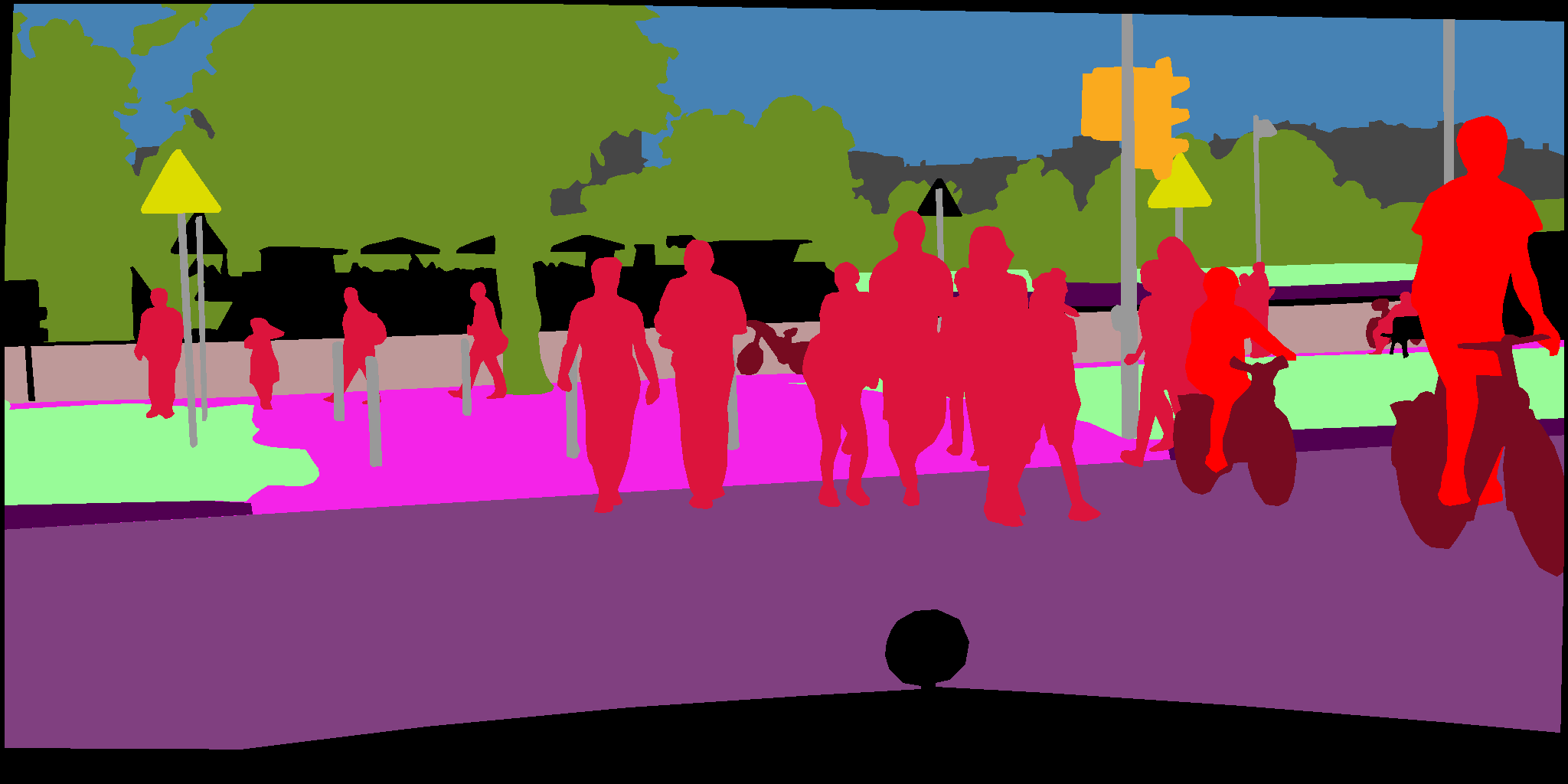}
   \\
   \addFig{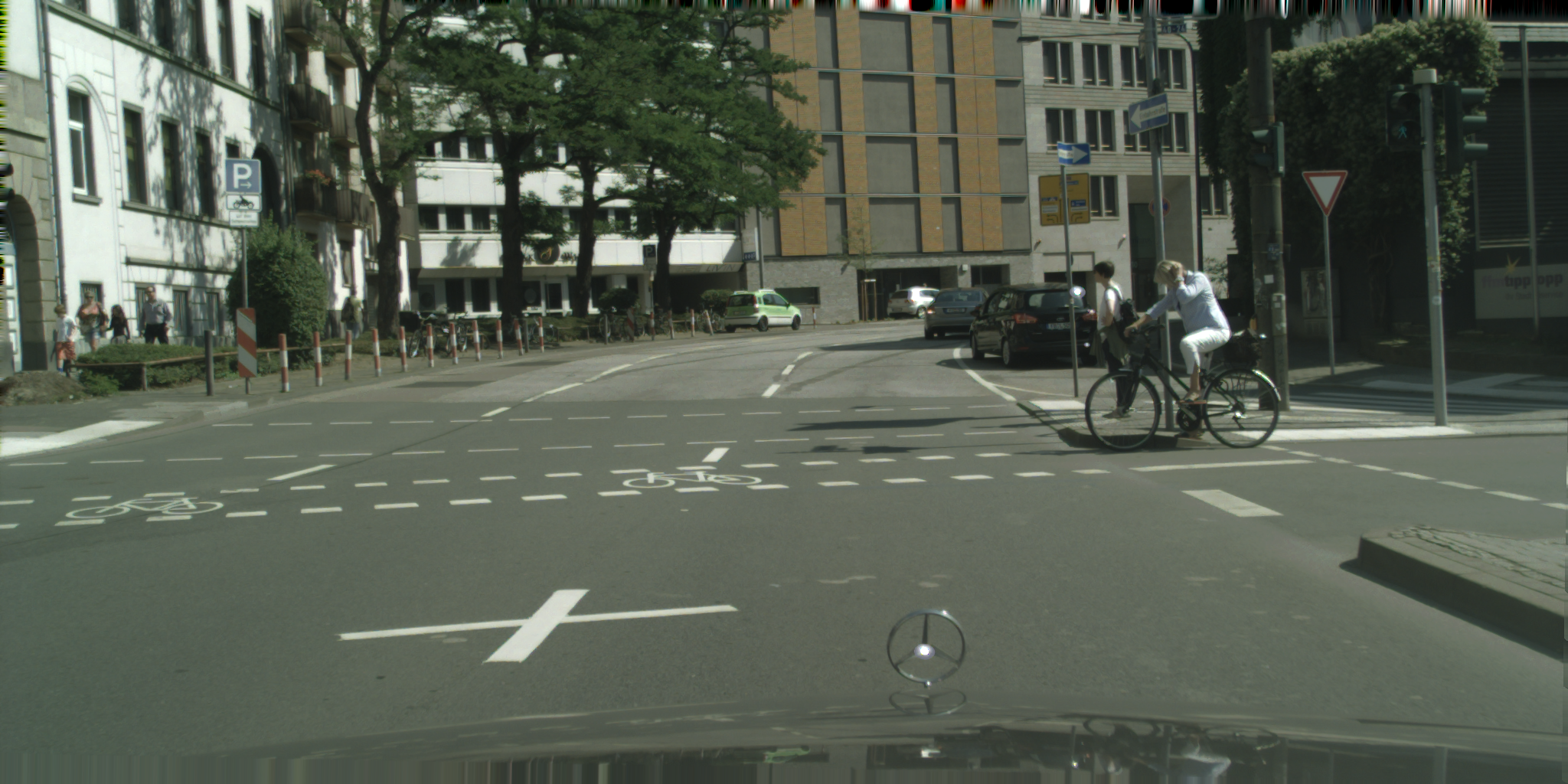}&
   \addFig{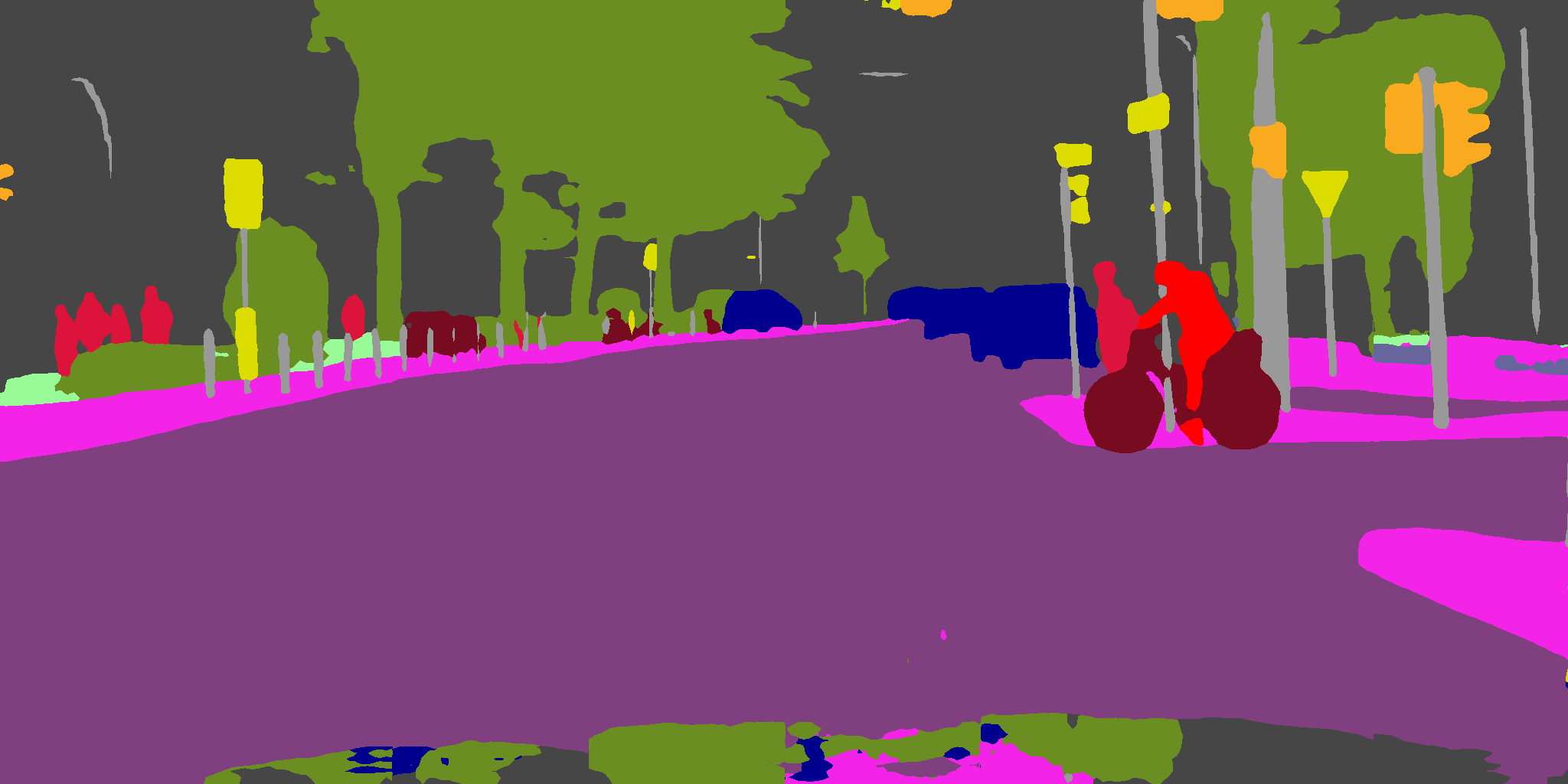}&
   \addFig{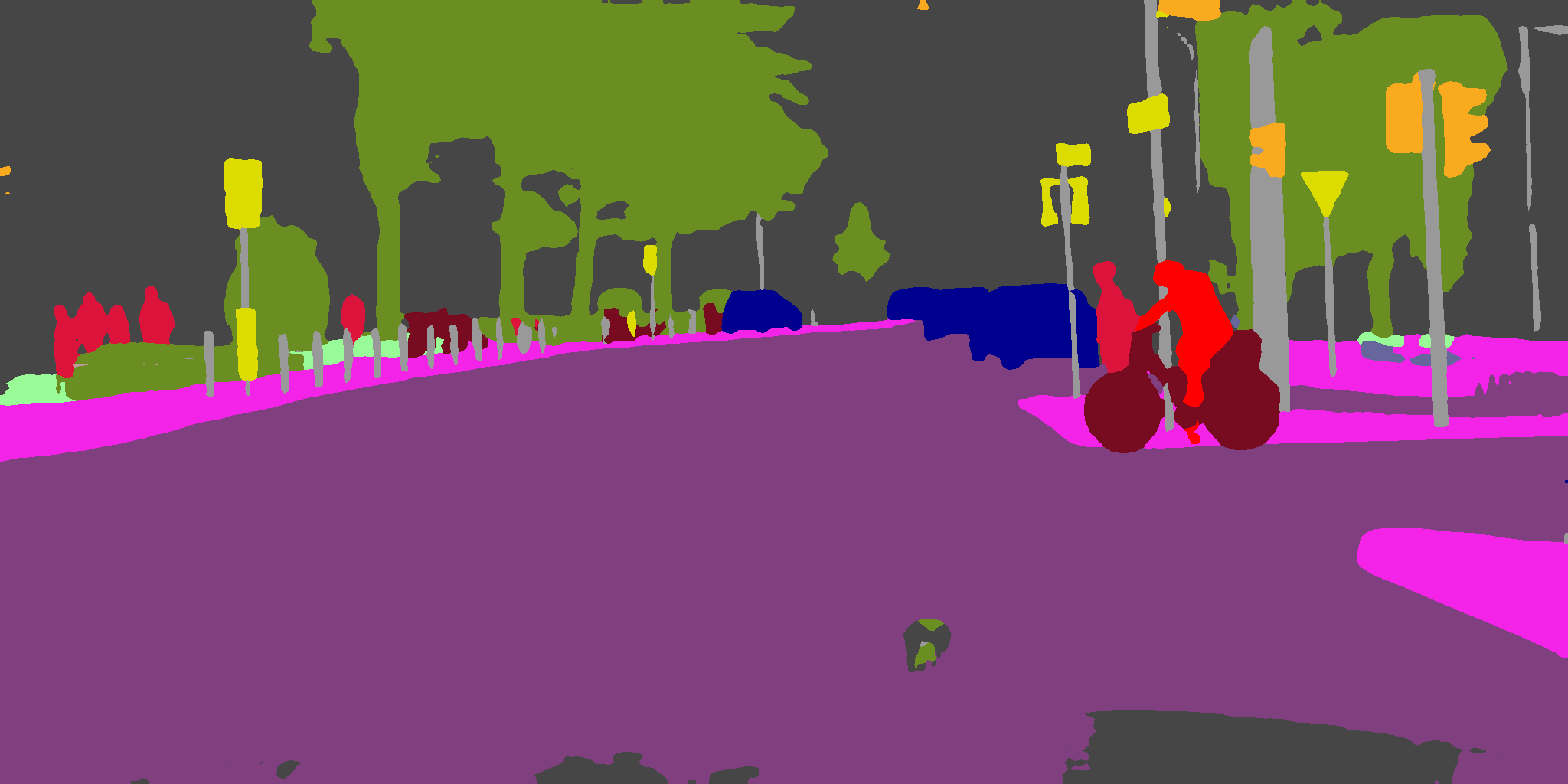}&
   \addFig{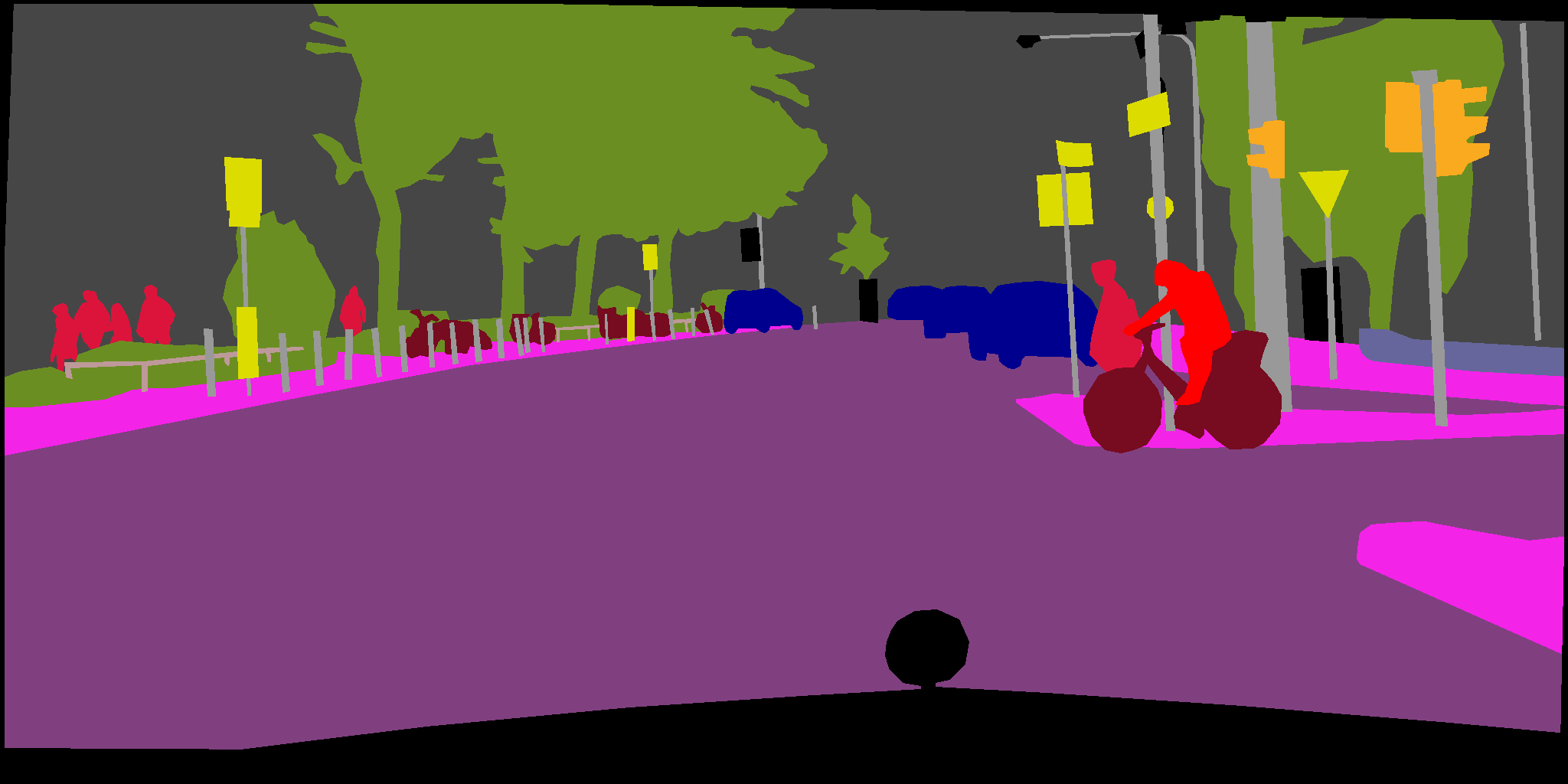}
   \\
   Image & Non-local & OCRNet & Ground Truth
   \\
   \addFig{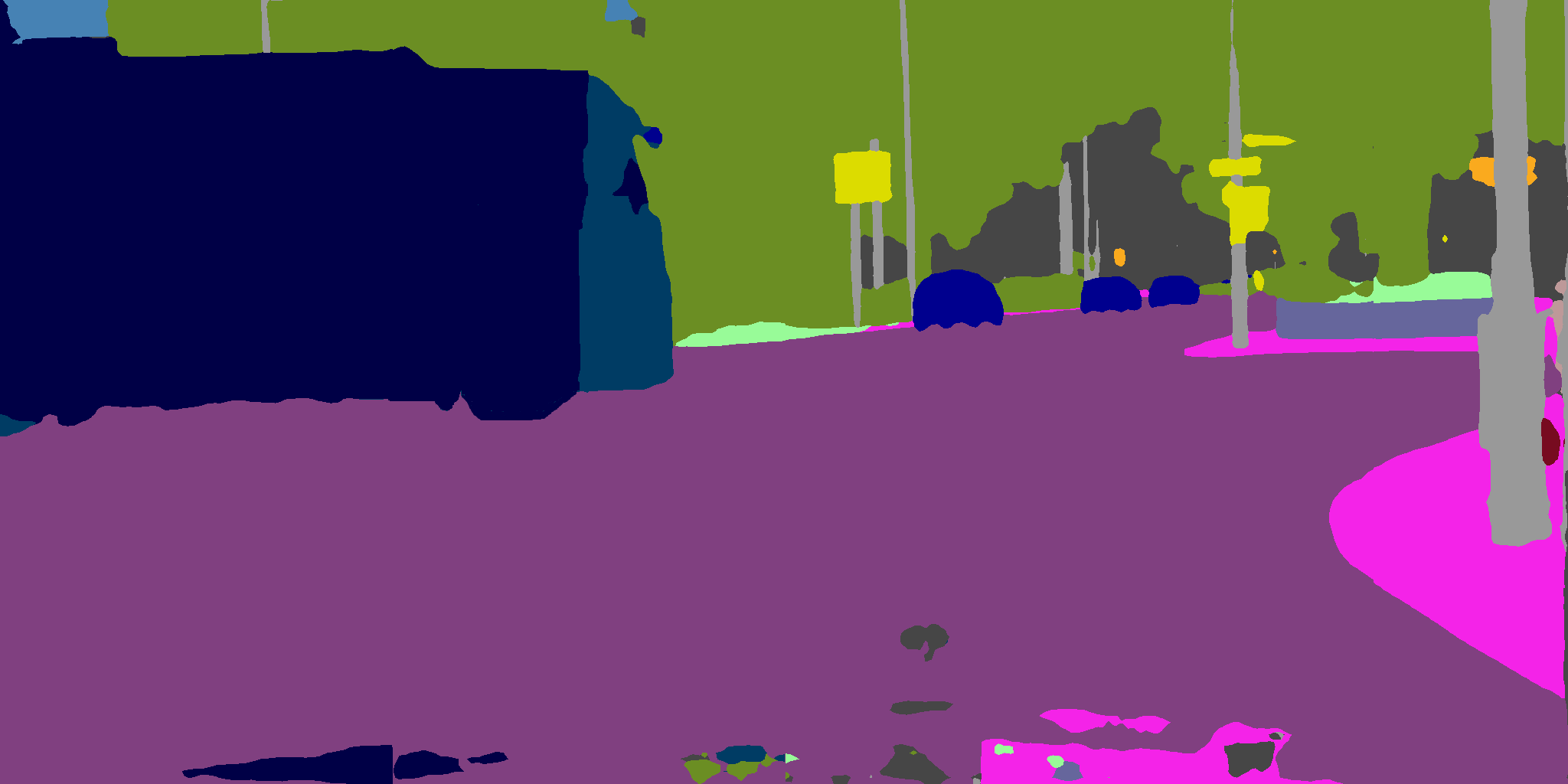}&
   \addFig{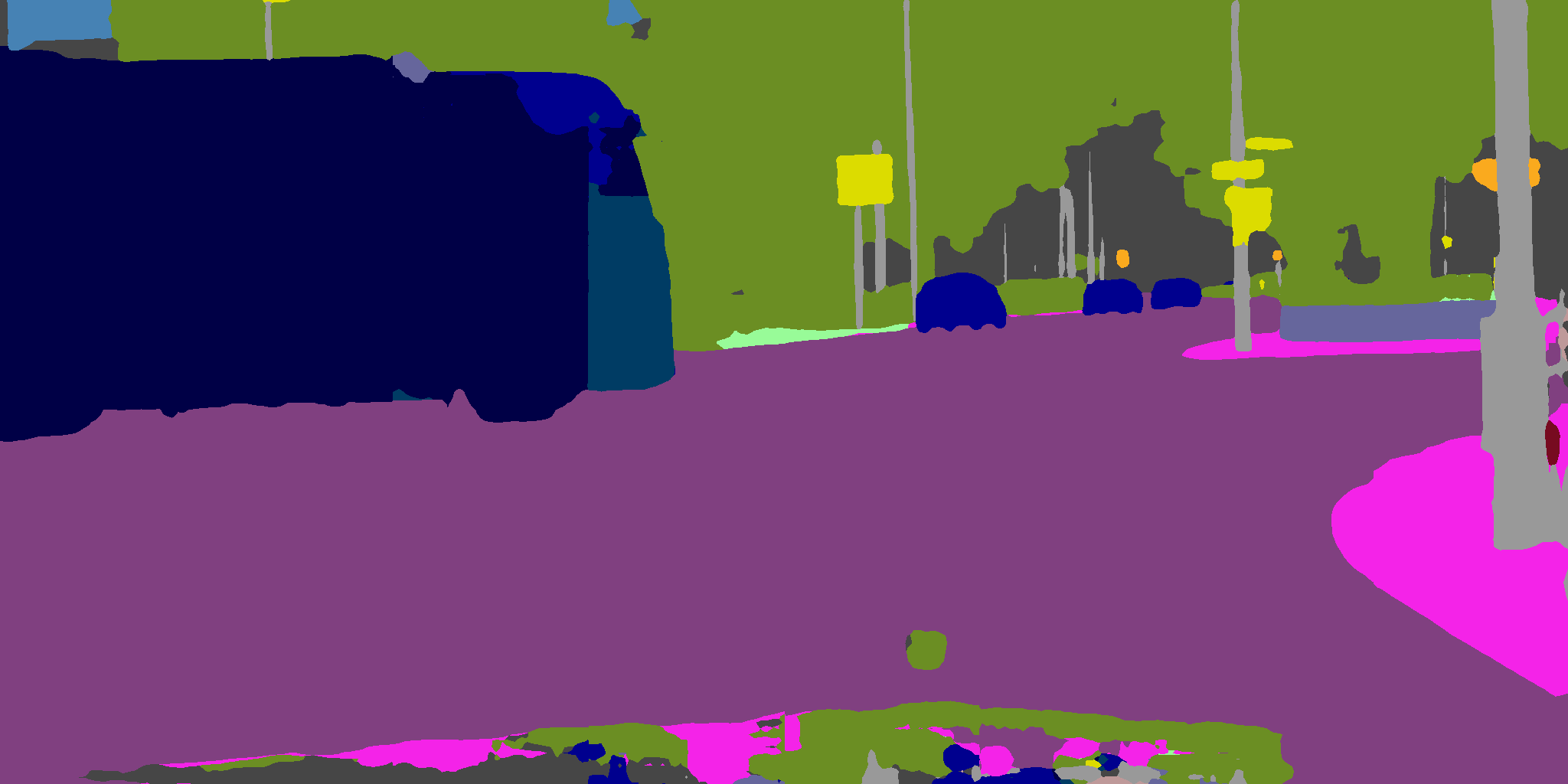}&
   \addFig{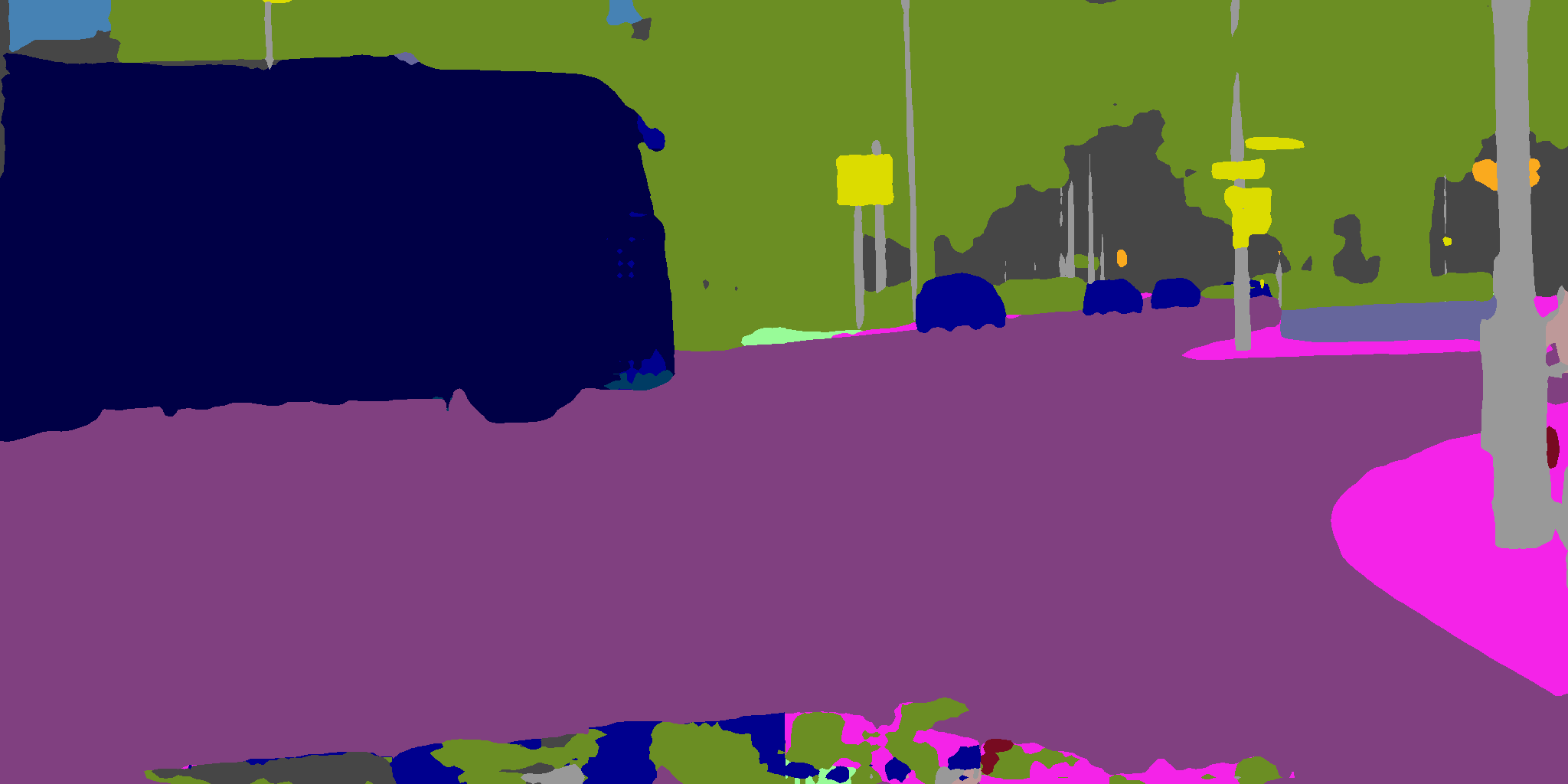}&
   \addFig{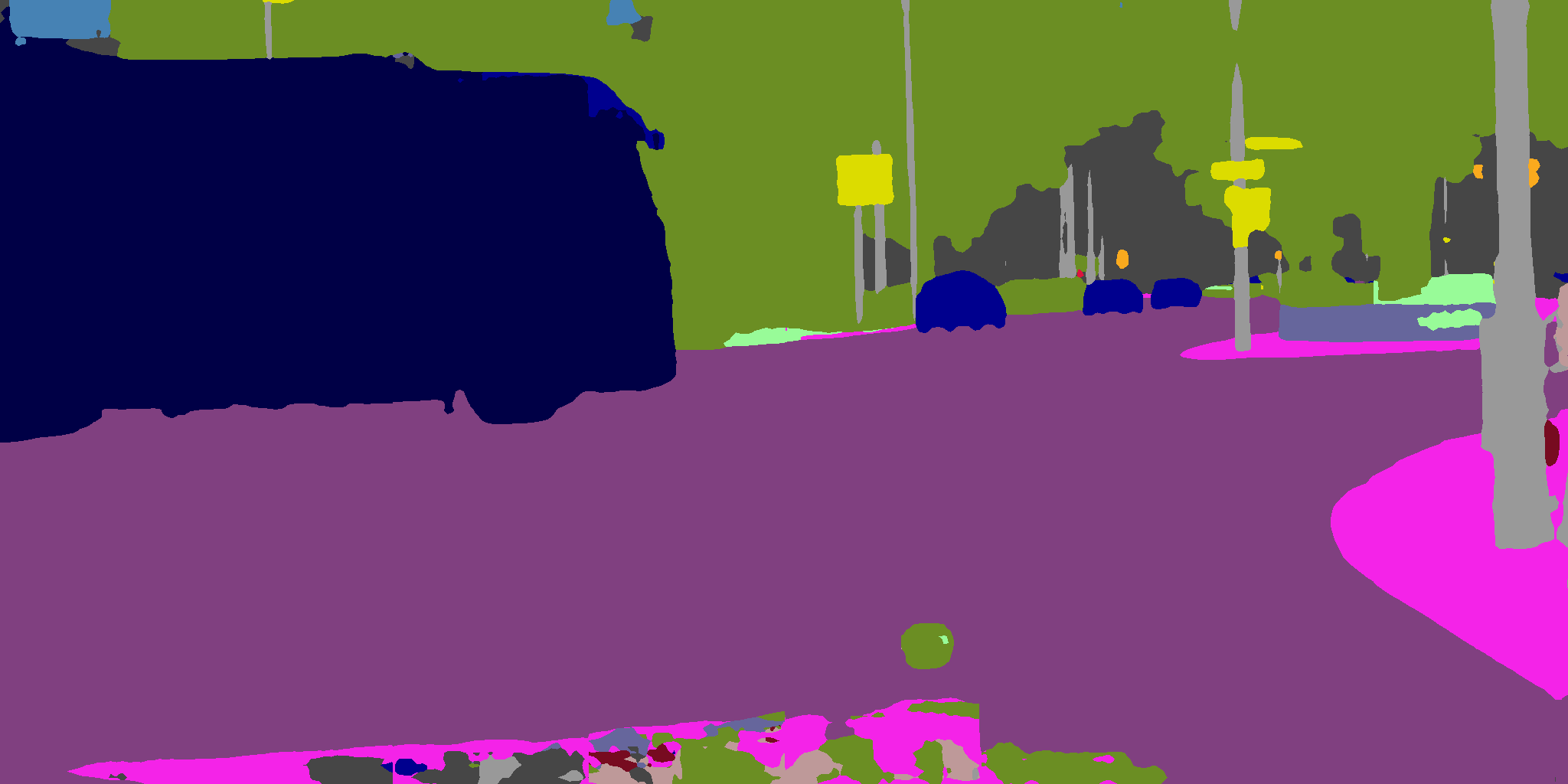}
   \\
   \addFig{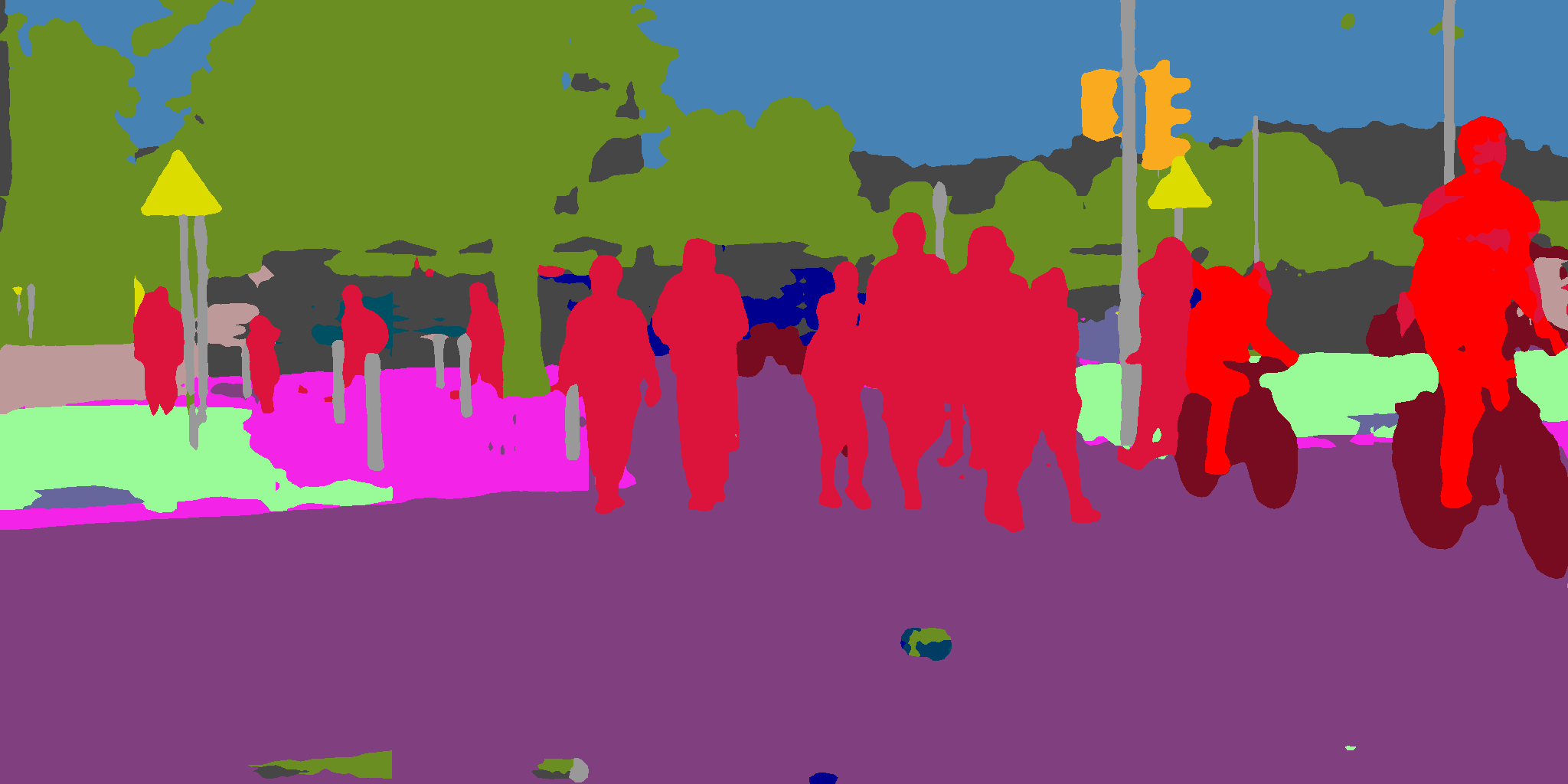}&
   \addFig{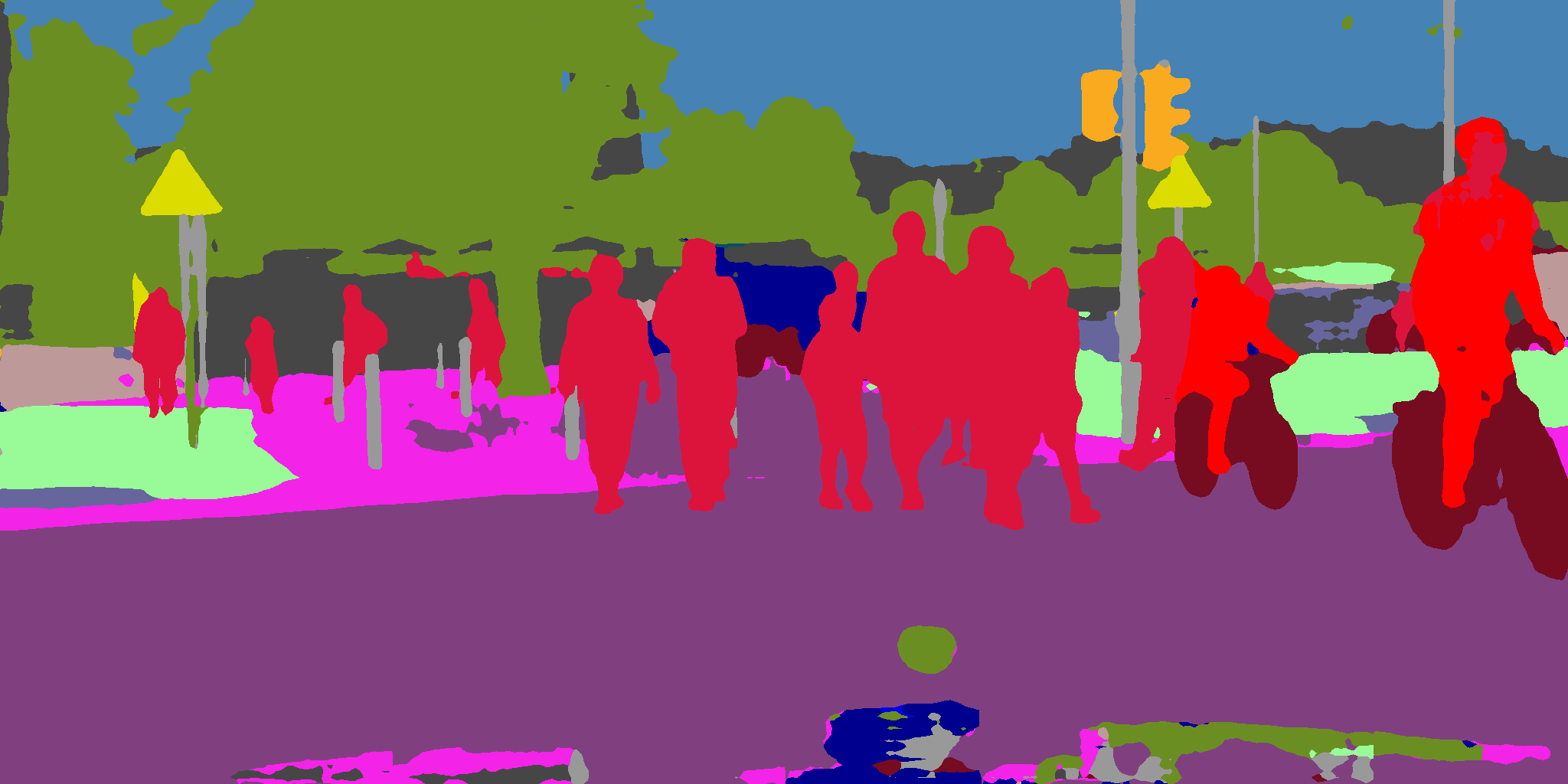}&
   \addFig{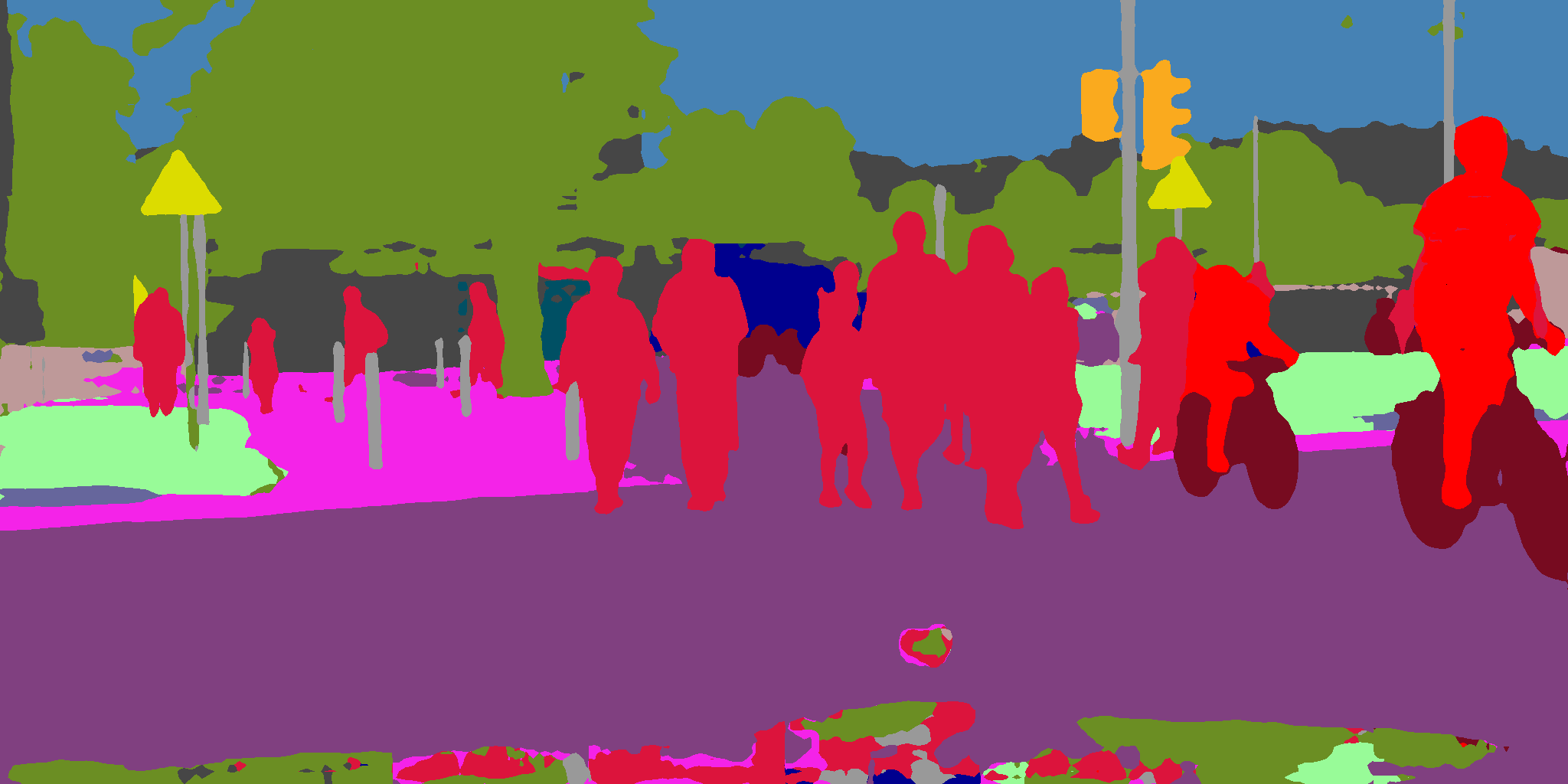}&
   \addFig{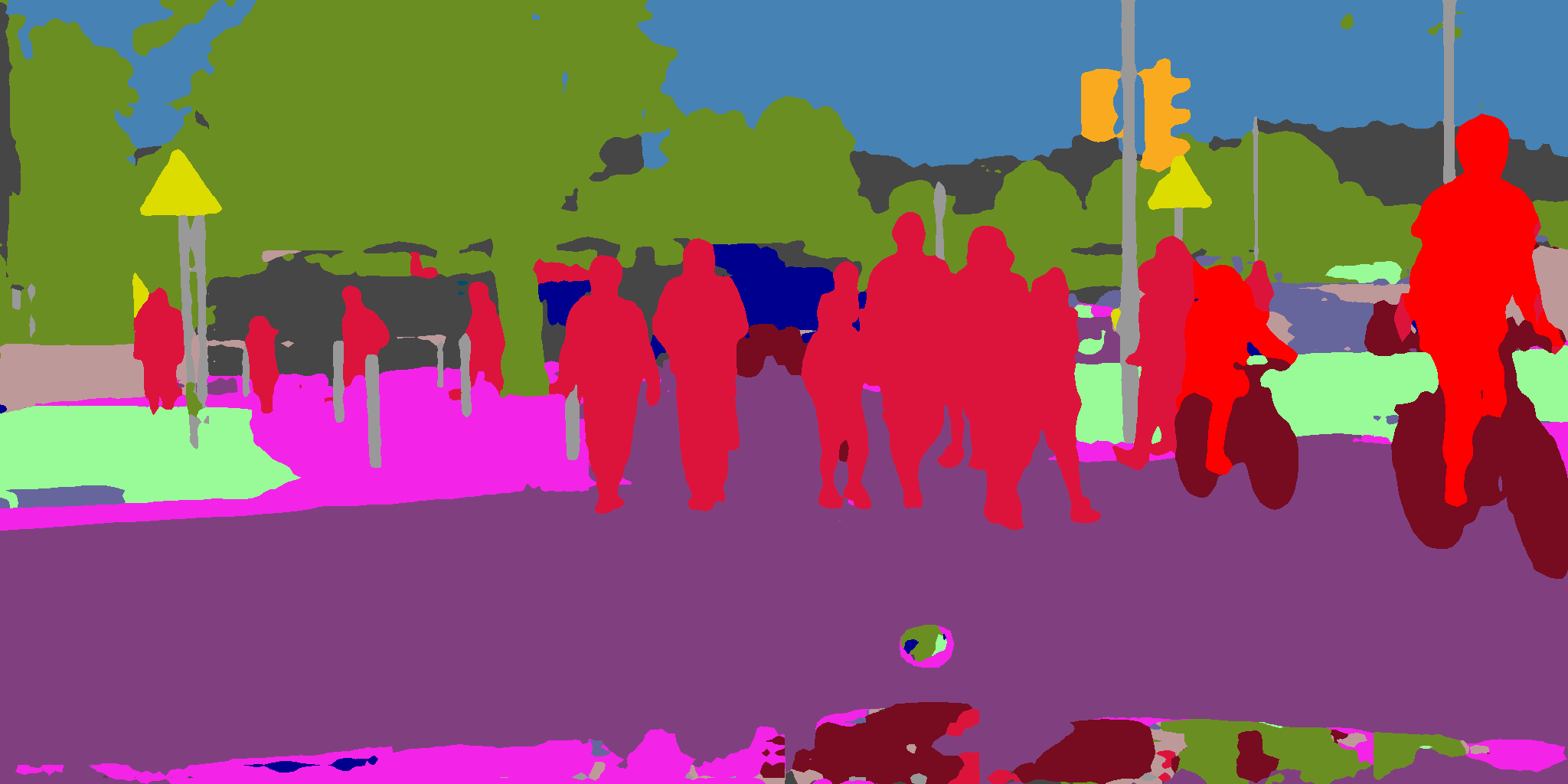}
   \\
   \addFig{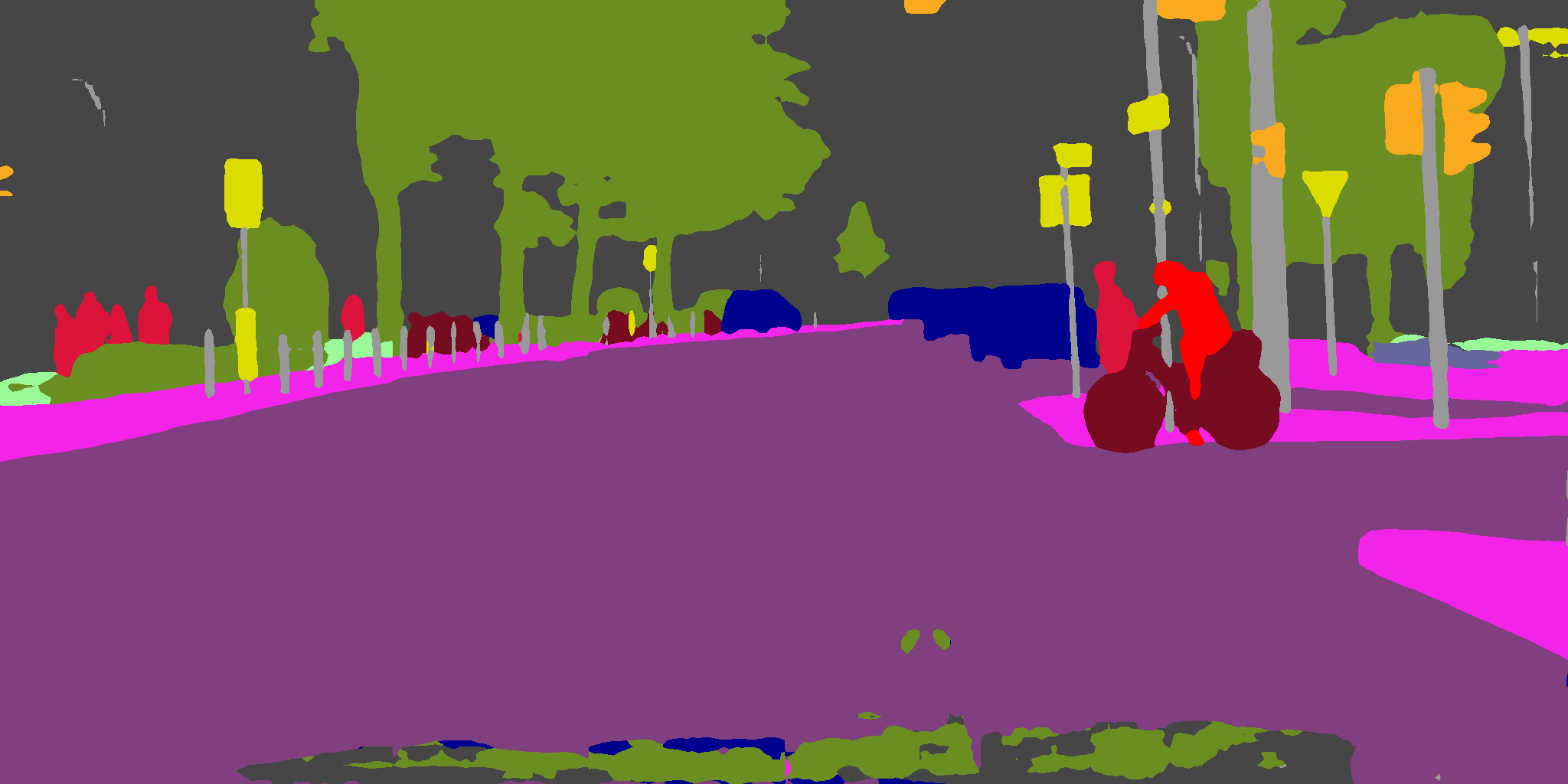}&
   \addFig{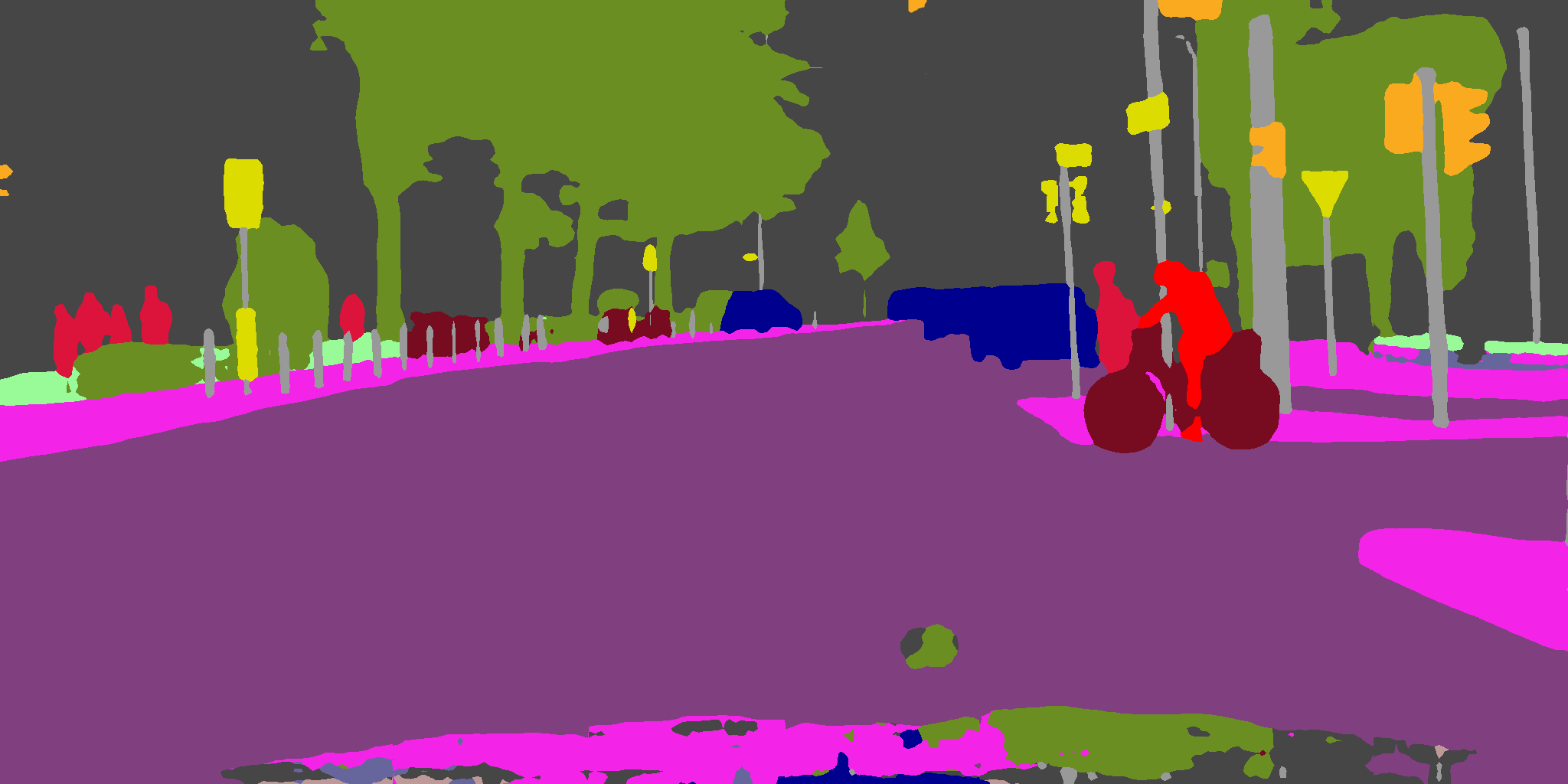}&
   \addFig{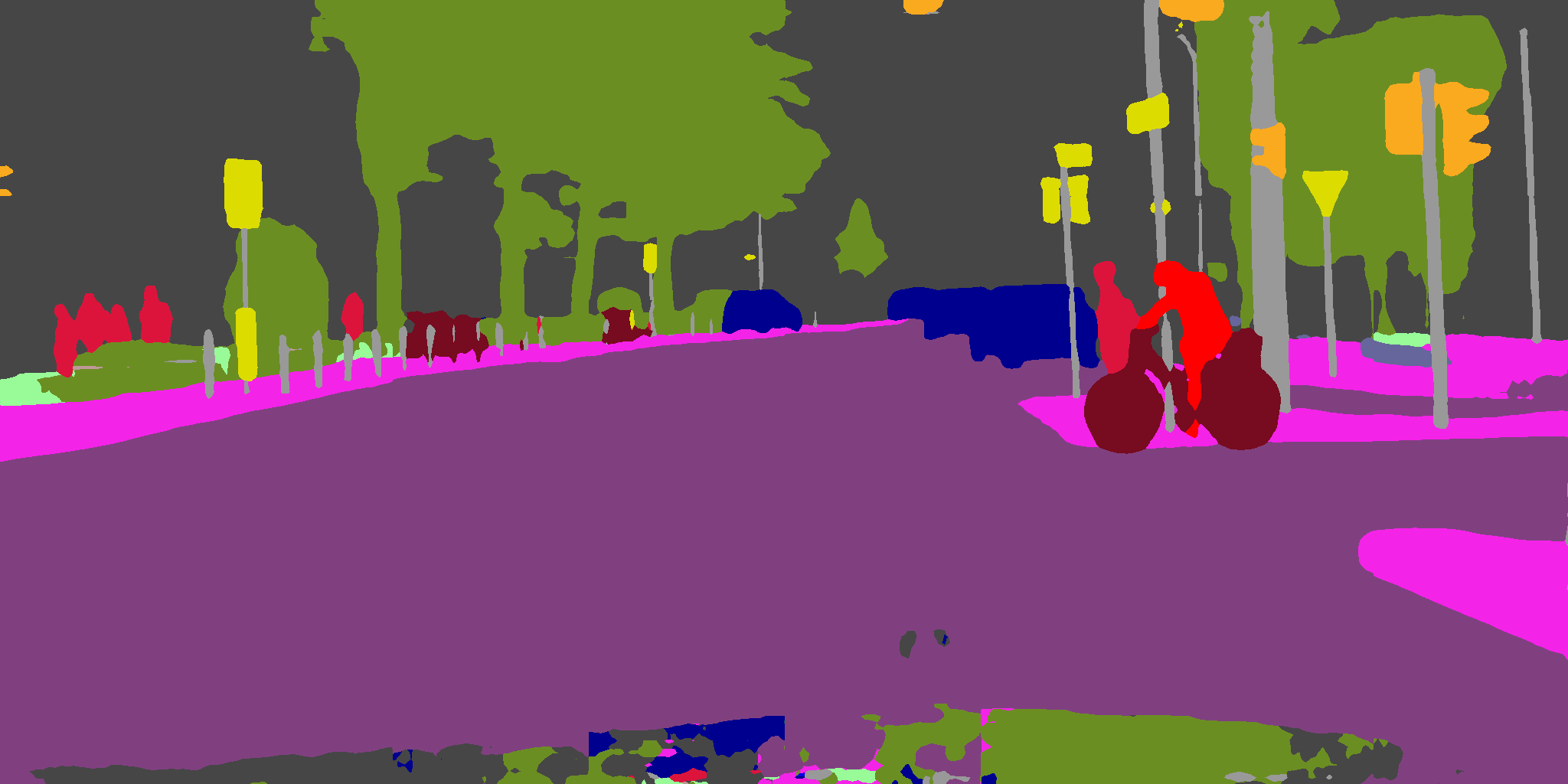}&
   \addFig{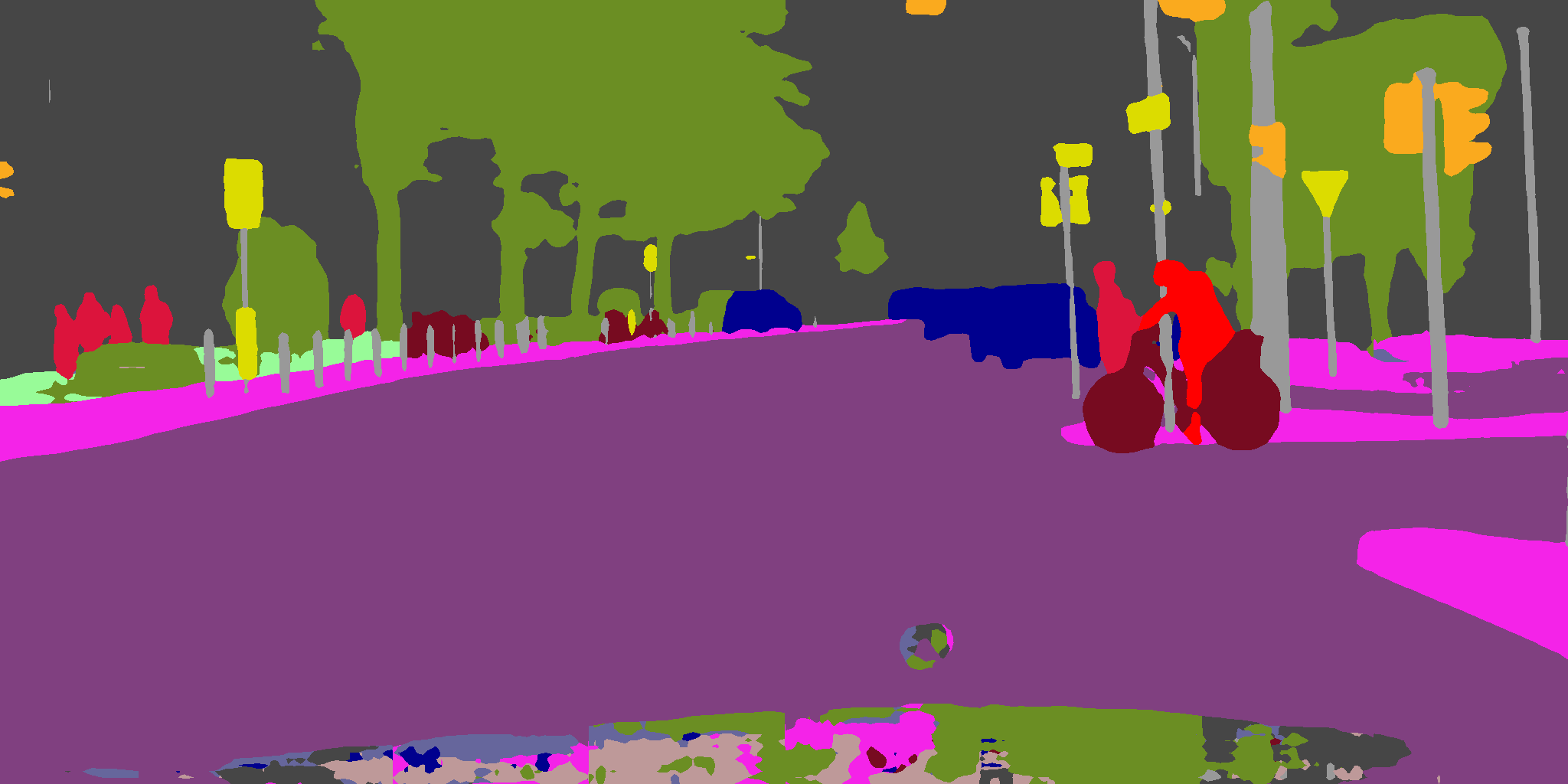}
   \\
   FsaNet-Lin-R1 & FsaNet-Dot-R1 & FsaNet-Lin-R2 & FsaNet-Dot-R2\\
  \end{tabular}
  \caption{Visualization of segmentation for the validation
set of Cityscapes.}
  \label{fig:cityseg}
\end{figure}
}

\textbf{Performance on Cityscapes test set}
We also compare \ourM based on ResNet-101 with other recent approaches on the Cityscapes test set. All approaches share dilated ResNet101 as the backbone but adopt different enhanced strategies. We enhance FsaNet by continuous loss, ohem training, extra data, and bigger crop size to get better results. \CCNet~adopts continuous loss and ohem training. \PSANet~adopts a bigger batch size and extra data. \DANet~and \DGCNet~adopts both spatial and channel attention. \HANet~introduces structural priors with respect to the height in urban-scene data. \OCRNet~adopts extra data. \DecoupledSegNets~adopts a context module in DeepLabV3+\cite{chen2018encoder}, decouples the body and edge feature by supervising separately, increases crop size, integrates multi-stage information, uses class uniforms sampling to enhance data, and uses ohem training. As seen from \tabref{citytestmIou}, FsaNet-Lin-R2 and FsaNet-Dot-R2 achieve the mIoU of 82.81\% and 83.05\%, which surpasses all the other well-designed models with ResNet-101 of previous \sArt work.\\
\begin{table}[tbp]
    \scriptsize
    \tabFormat
    \setlength{\tabcolsep}{1mm}
    \caption{Comparison with state-of-the-arts on Cityscapes (test). ‡: trained on fine and coarse annotated data.
    }
    \begin{tabular}{lcccc}
    \toprule
    Method & Backbone &Batch size& Crop size & mIoU(\%)\\
    \midrule
    \CCNet &  ResNet101& 8 & 769 & 81.40\\
    \PSANet‡ & ResNet101 & 16 & 705 & 81.40\\
    \DANet & ResNet101 & 8 & 768 & 81.50\\
    \DGCNet & ResNet101& 8 & 769 & 82.00\\
    \HANet & ResNet101 & 8 & 768 & 82.10\\
    \OCRNet‡ & ResNet101 & 8 & 769 &82.40\\
    \DecoupledSegNets & ResNet101 & 8 & 832 & 82.80\\
    FsaNet-Lin-R2‡ & ResNet101 & 8 & 769 & 82.31\\
    FsaNet-Dot-R2‡ & ResNet101 & 8 & 769 & 82.25\\
    FsaNet-Lin-R2‡ & ResNet101 & 8 & 849 & 82.81\\
    FsaNet-Dot-R2‡ & ResNet101 & 8 & 849 & \textbf{83.05}\\
    \bottomrule
    \end{tabular}
    \label{citytestmIou}
\end{table}

\subsection{FsaNet Performance on ADE20K}\label{sec:Experiments_ADE}
We conduct experiments on the AED20K and show the results in \tabref{ADEmiouTab}. The FsaNet achieve significant performance gains over the \FCN~with 3.43$\sim$4.49\% by mIoU. This significant improvement demonstrates the effectiveness of proposed frequency self-attention modules on ADE20K. Compared with the \NonlocalNetwork, \ourM has increased mIoU by 0.14$\sim$1.2\%. When $R=1$, \ourM can slightly outperform the \NonlocalNetwork. When $R=2$, \ourM can achieve performance that exceeds or is comparable to previous \sArt methods based on ResNet101.
We also show \ourM segmentation results compared to \NonlocalNetwork~and \OCRNet~on ADE20K in \figref{fig:adeseg}. In the first row it can be observed that the four versions of FsaNet can better maintain the integrity and internal consistency of the wall than \NonlocalNetwork~and \OCRNet. The climbing plants on the house in FsaNet are also more completely segmented. Although FsaNet mistakes plants for trees, such misunderstandings may also occur in human perception. In the second row, we observe that FsaNet better promotes consistency inside lamps and pool tables. In the result of \NonlocalNetwork, some parts of the table lamp are identified as chandeliers, and in \OCRNet~some pixels on the left leg of the pool table are misattributed to person. In the third row, in the result of \NonlocalNetwork, some pixels of the bed are misattributed to pillow, and in \OCRNet~the box and paintings were ignored on the chest of drawers. This indicates that \ourM can better promote the internal consistency of large objects and is also beneficial to the recognition of small objects. 
 \renewcommand{\addFig}[1]{{\includegraphics[width=.06\textwidth]{#1.jpg}}}
\CheckRmv{
\begin{figure}[t]
  \centering
  \tiny
  \renewcommand{\arraystretch}{0.5}
  \setlength{\tabcolsep}{0.0mm}
  \begin{tabular}{cccccccc}
   \addFig{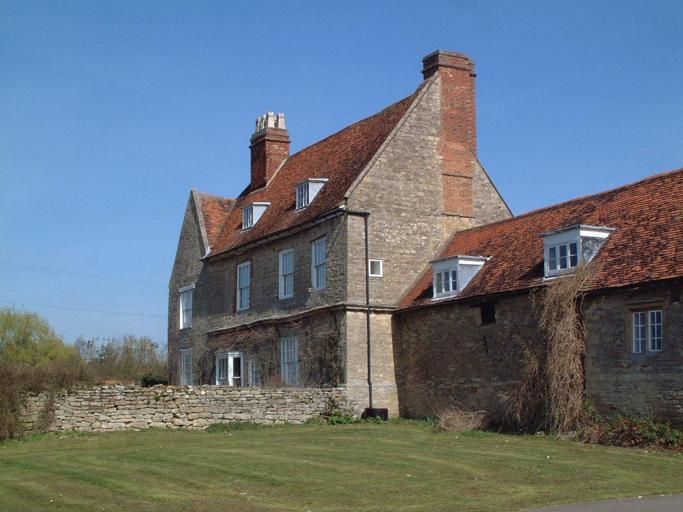}&
   \addFig{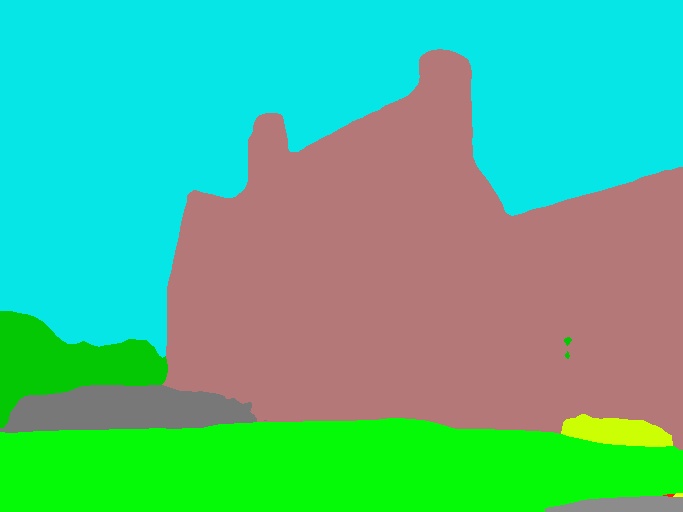}&
   \addFig{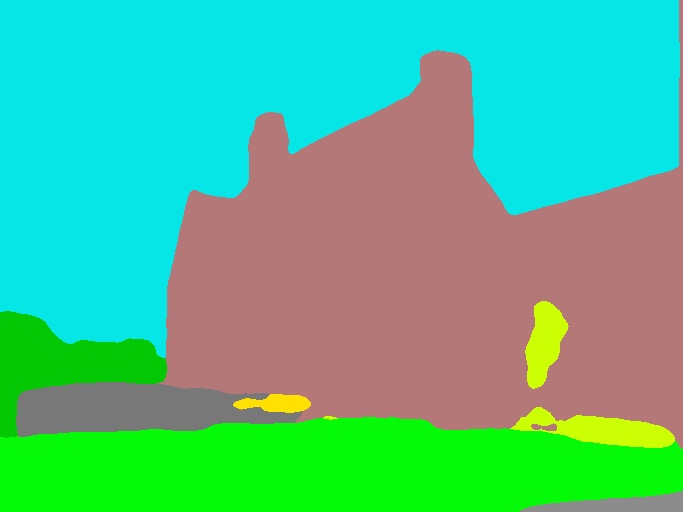}&
   \addFig{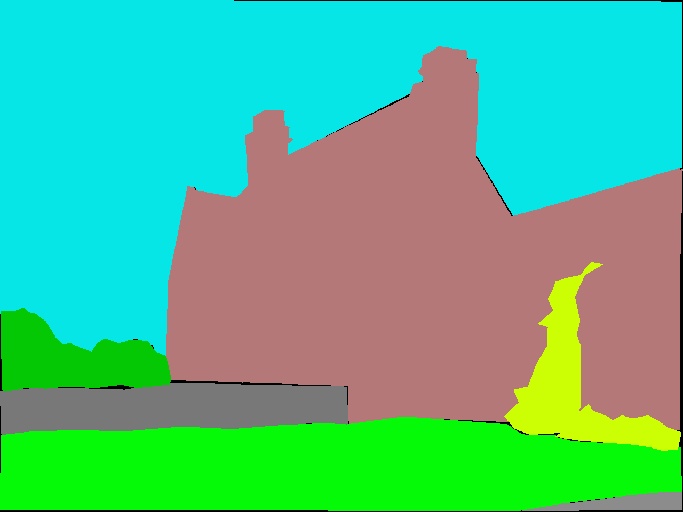}&
   \addFig{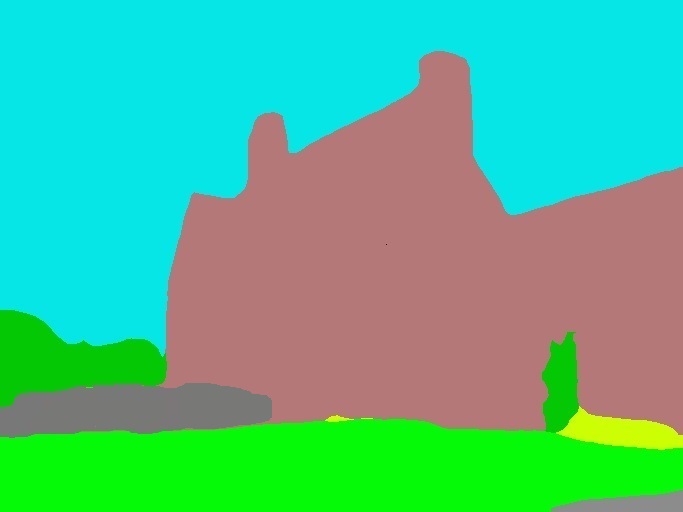}&
   \addFig{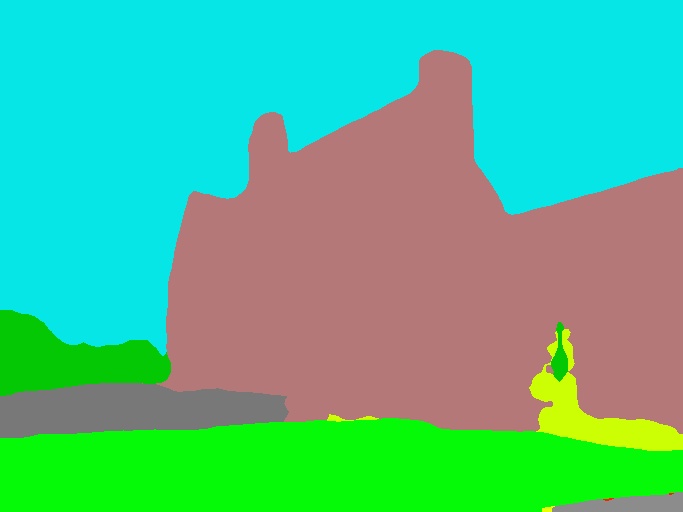}&
   \addFig{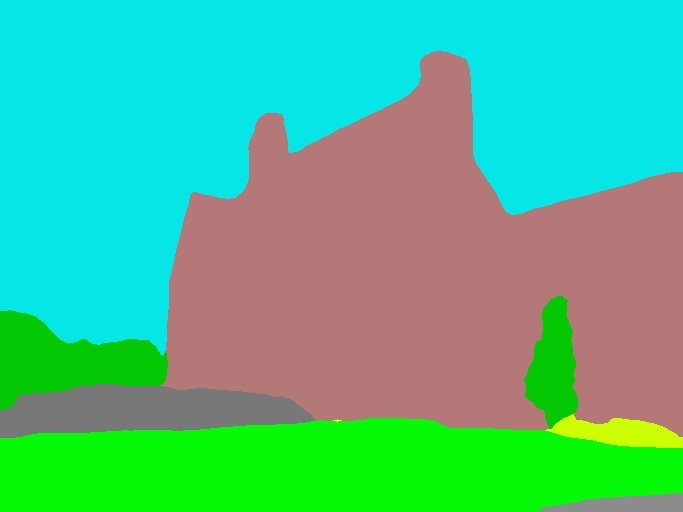}&
   \addFig{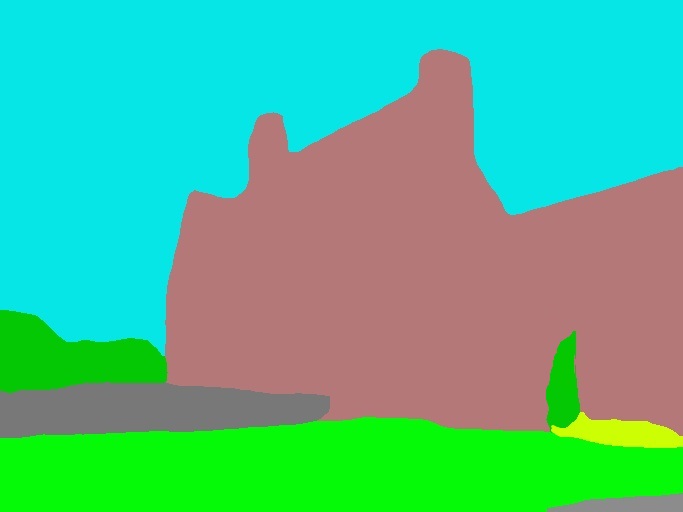}
   \\
   \addFig{adeseg/00000988/Image}&
   \addFig{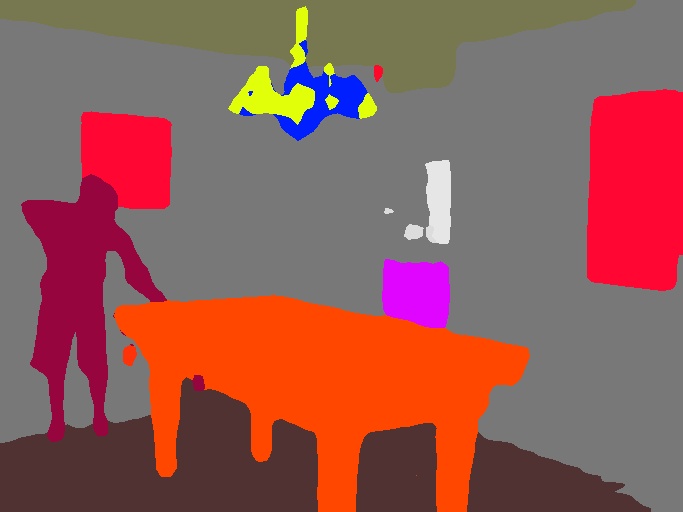}&
   \addFig{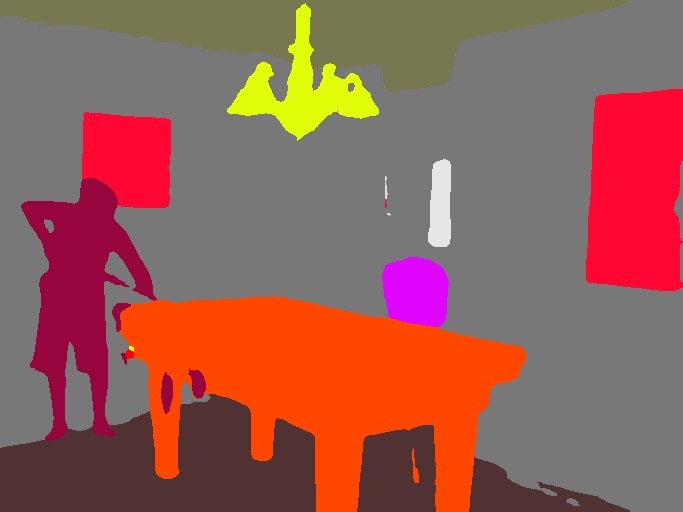}&
   \addFig{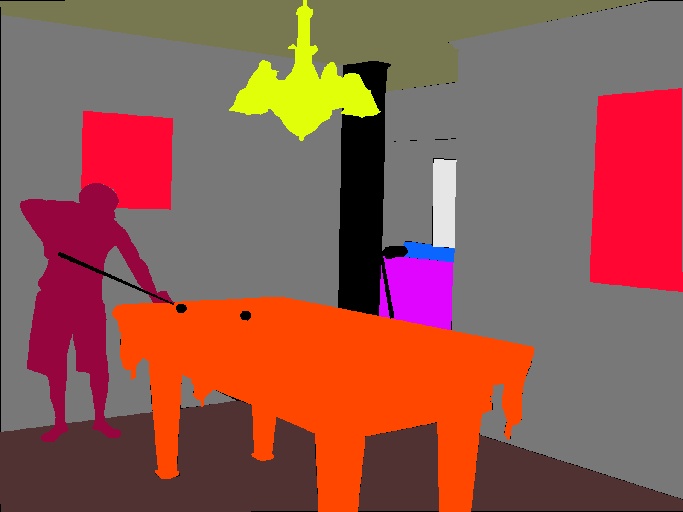}&
   \addFig{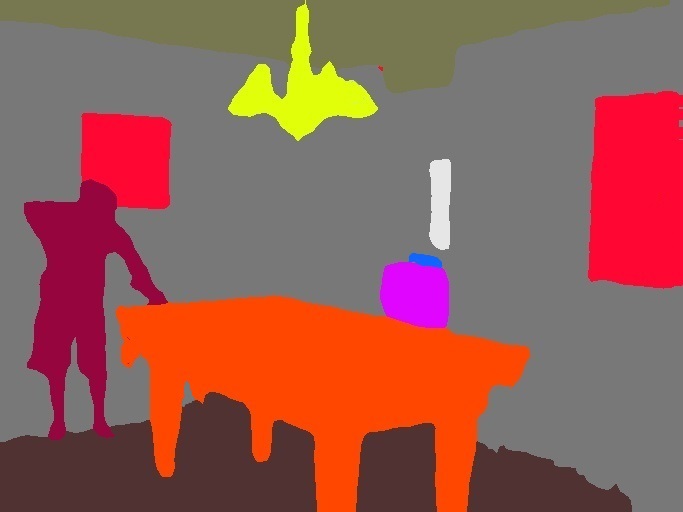}&
   \addFig{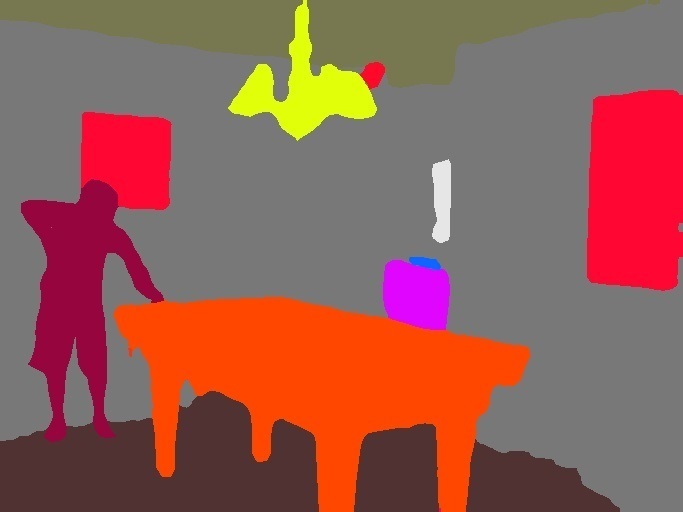}&
   \addFig{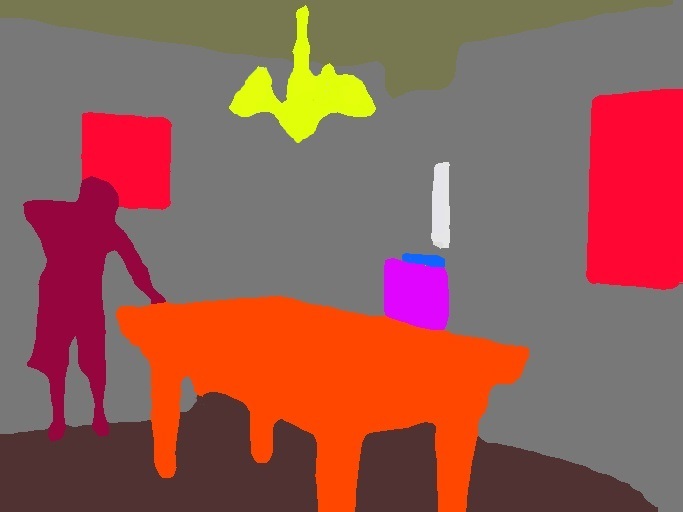}&
   \addFig{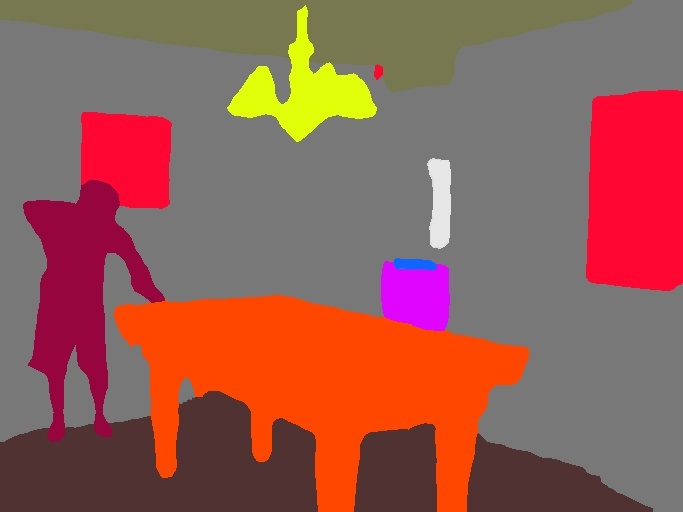}
   \\
   \addFig{adeseg/00000152/Image}&
   \addFig{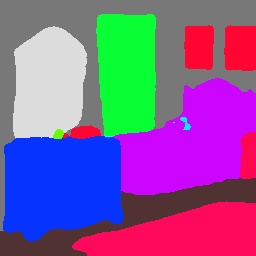}&
   \addFig{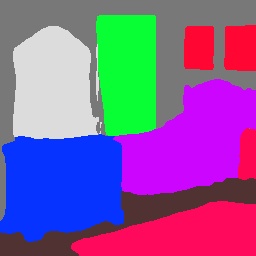}&
   \addFig{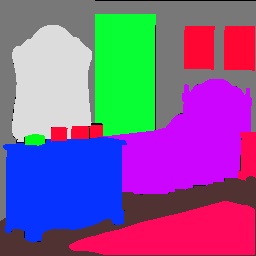}&
    \addFig{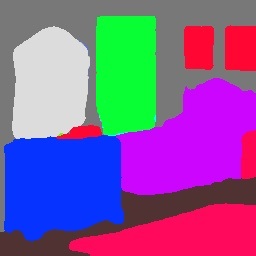}&
   \addFig{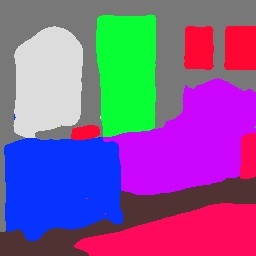}&
   \addFig{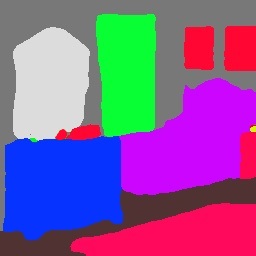}&
   \addFig{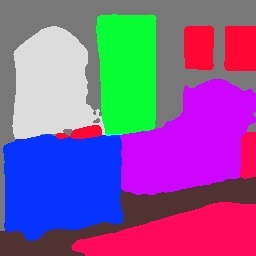}
   \\
  Image & Non-local & OCRNet & Ground Truth& FsaNet-Lin-R1 & FsaNet-Dot-R1 & FsaNet-Lin-R2 & FsaNet-Dot-R2
  \end{tabular}
  \caption{Visualization of segmentation for ADE20K.}
  \label{fig:adeseg}
\end{figure}
}
\subsection{FsaNet Performance on VOCaug}\label{sec:Experiments_VOC}
We also compare \ourM on the VOCaug\cite{pascal-voc-2012} with \sArt methods in \tabref{VOCmiouTab}. The \ourM has an improvement of 7.73$\sim$9.07\% in mIoU over \FCN, demonstrating the effectiveness of our modules. We observe that \NonlocalNetwork~outperforms previous approaches and \ourM achieves better results with less computational consumption.
Visual comparisons of semantic segmentation results on VOC are illustrated in \figref{fig:vocseg}. In the three rows of images, we observe that \ourM tends to accurately preserve the outline details for bicycles, cars and airplanes, separately. In the second line, the number of cars is hard to be counted because the segmentation of \NonlocalNetwork~and \OCRNet~is not fine.
 \renewcommand{\addFig}[1]{{\includegraphics[width=.06\textwidth]{#1.jpg}}}
 \newcommand{\addFigpng}[1]{{\includegraphics[width=.06\textwidth]{#1.png}}}
\CheckRmv{
\begin{figure}[t]
  \centering
  \tiny
  \renewcommand{\arraystretch}{0.5}
  \setlength{\tabcolsep}{0.1mm}
  \begin{tabular}{cccccccc}
   \addFig{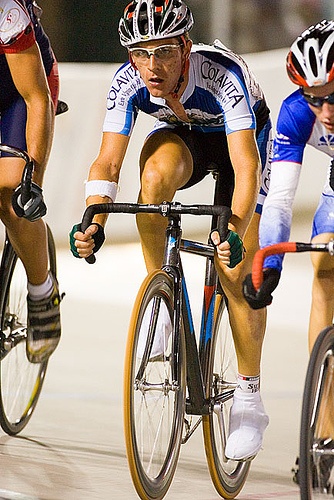}&
   \addFig{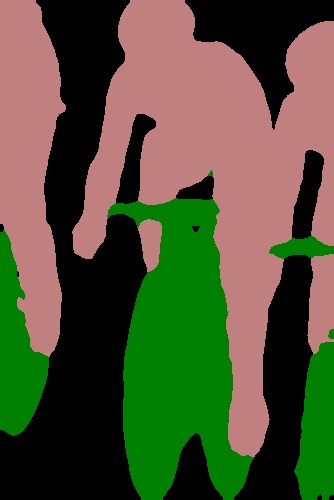}&
   \addFig{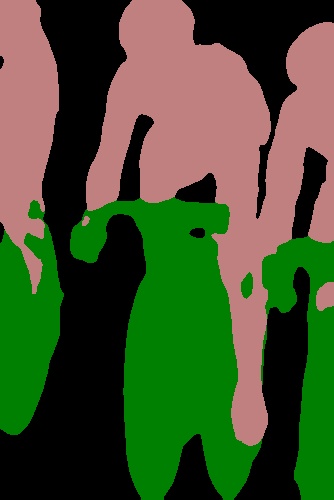}&
   \addFigpng{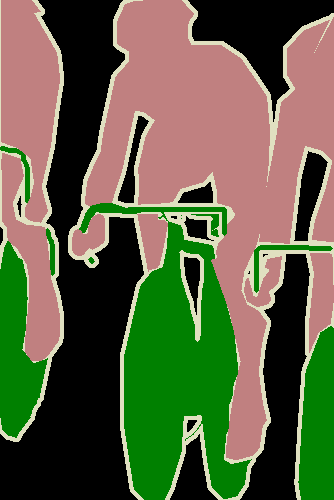}&
   \addFig{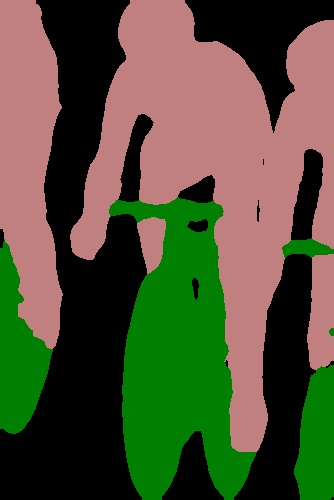}&
   \addFig{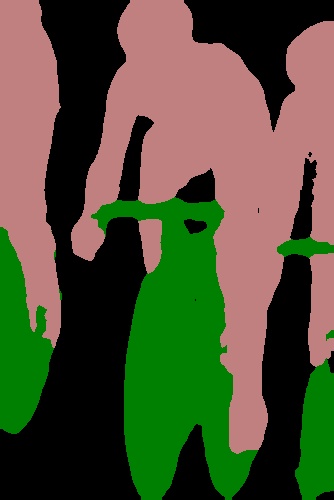}&
   \addFig{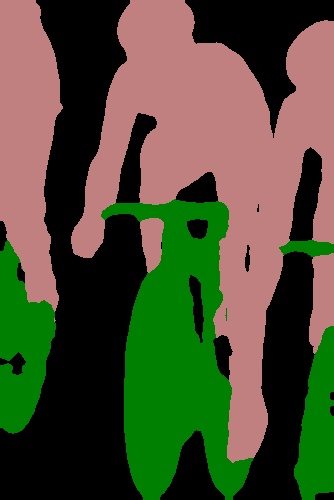}&
   \addFig{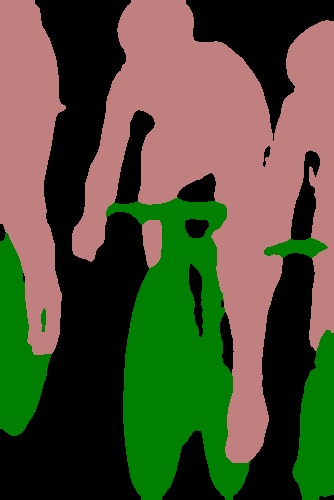}
   \\
   \addFig{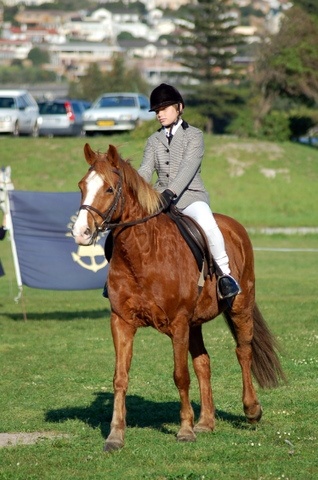}&
   \addFig{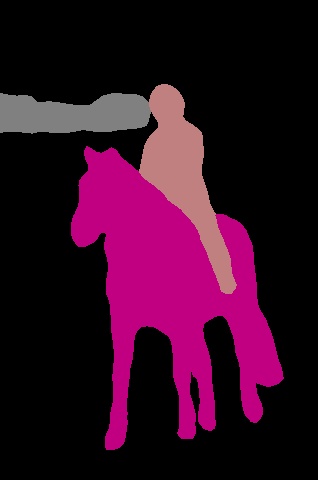}&
   \addFig{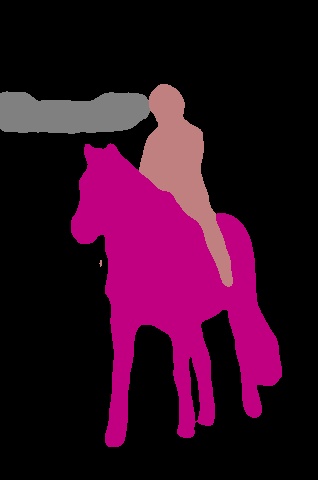}&
   \addFigpng{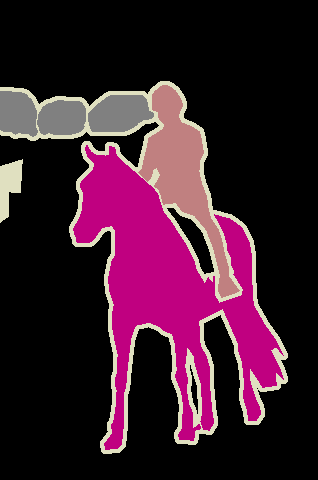}&
   \addFig{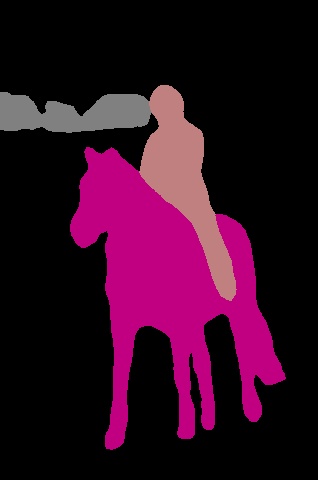}&
   \addFig{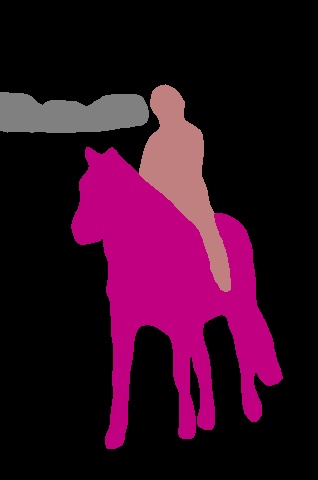}&
   \addFig{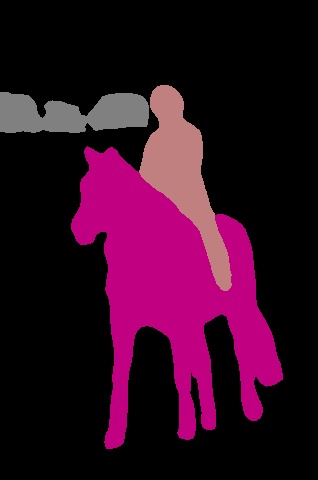}&
   \addFig{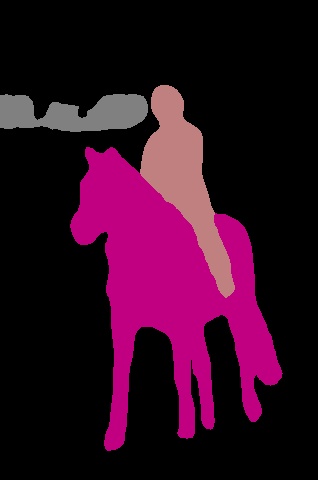}
   \\
   \addFig{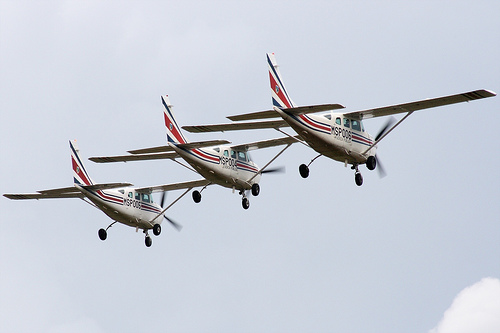}&
   \addFig{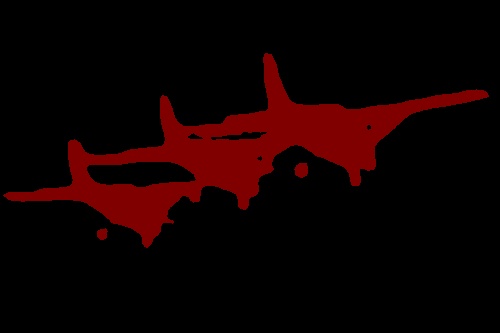}&
   \addFig{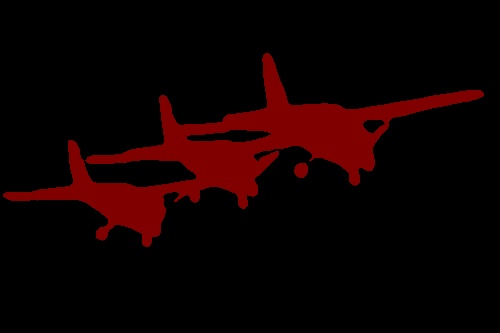}&
   \addFigpng{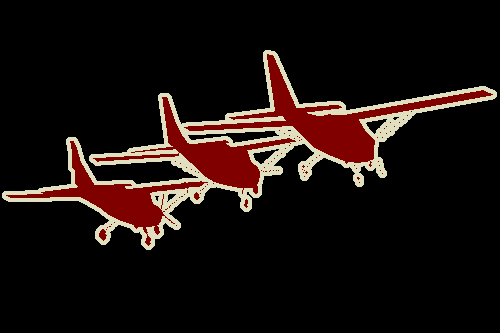}&
   \addFig{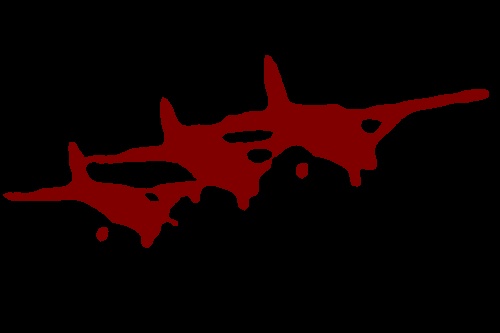}&
   \addFig{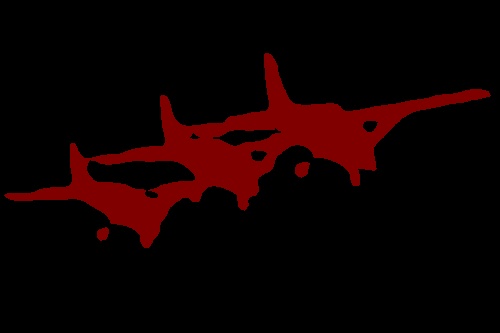}&
   \addFig{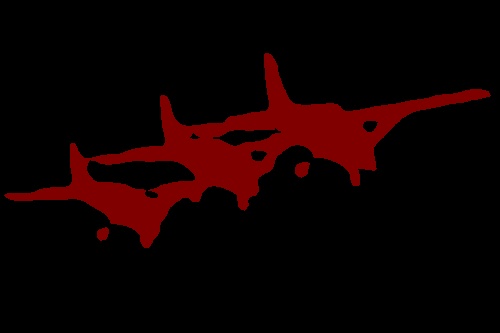}&
   \addFig{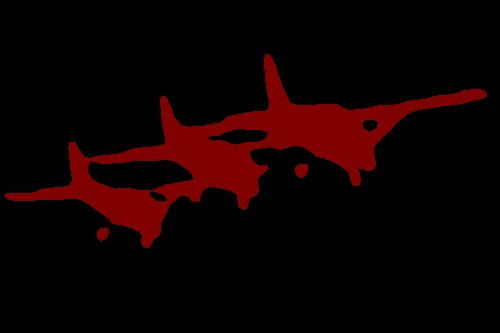}
   \\
  Image & Non-local & OCRNet & Ground Truth& FsaNet-Lin-R1 & FsaNet-Dot-R1 & FsaNet-Lin-R2 & FsaNet-Dot-R2\\
  \end{tabular}
  \caption{Visualization of segmentation for VOCaug.}
  \label{fig:vocseg}
\end{figure}
}

\subsection{FsaNet Performance on COCO}
To further demonstrate the generality of FsaNet, we conduct the instance segmentation task on COCO\cite{lin2014microsoft} using the competitive Mask R-CNN\cite{he2017mask} as the baseline. We modify the Mask R-CNN\cite{he2017mask} backbone by adding the frequency self-attention module and evaluate a standard baseline of ResNet-50. All models are fine-tuned from ImageNet pre-training.
The results in terms of box AP and mask AP are reported in \tabref{cocoAP}, which demonstrate that our method substantially outperforms the baseline in all metrics. Some segmentation results for comparing the baseline with “+FsaNet” are given in \figref{fig:cocoseg}.
\CheckRmv{
\begin{table}[tbp]
    \scriptsize
    \tabFormat
    \setlength{\tabcolsep}{2.2mm}
    \caption{Performance on COCO.}
    \begin{tabular}{lccc}
    \toprule
    Method &           $AP^{box}$ & $AP^{mask}$\\
    \midrule
    R50 baseline       &39.2 & 35.4\\
    R50+Fsa-Dot-R1     &39.9$\textcolor{red}{\uparrow}_{0.7}$ & 36.1$\textcolor{red}{\uparrow}_{0.7}$\\
    R50+Fsa-Dot-R2     &39.9$\textcolor{red}{\uparrow}_{0.7}$ & 36.1$\textcolor{red}{\uparrow}_{0.7}$\\
    R50+Fsa-Lin-R1     &40.2$\textcolor{red}{\uparrow}_{1.0}$ & 36.3$\textcolor{red}{\uparrow}_{0.9}$\\
    R50+Fsa-Lin-R2     &40.1$\textcolor{red}{\uparrow}_{0.9}$ & 36.3$\textcolor{red}{\uparrow}_{0.9}$\\
    \bottomrule
    \end{tabular}
    \label{cocoAP}
\end{table}
}

 \renewcommand{\addFig}[1]{{\includegraphics[width=.081\textwidth]{#1.jpg}}}
\CheckRmv{
\begin{figure}[t]
  \centering
  \tiny
  \renewcommand{\arraystretch}{0.5}
  \setlength{\tabcolsep}{0.0mm}
  \begin{tabular}{lccccc}
   \addFig{cocoseg/cocoimg}&
   \addFig{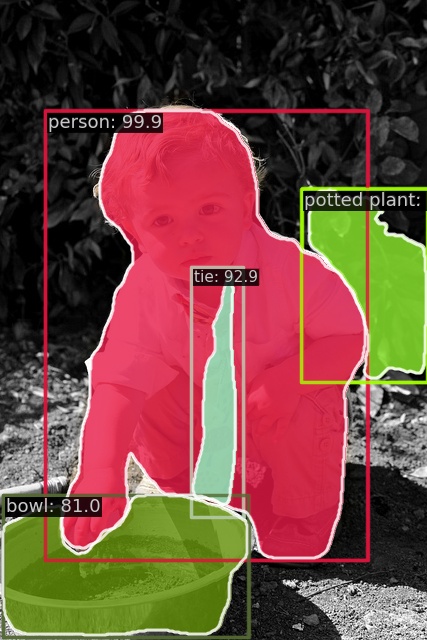}&
   \addFig{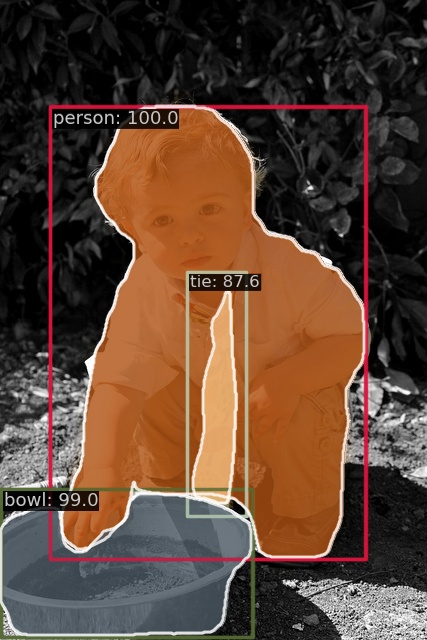}&
   \addFig{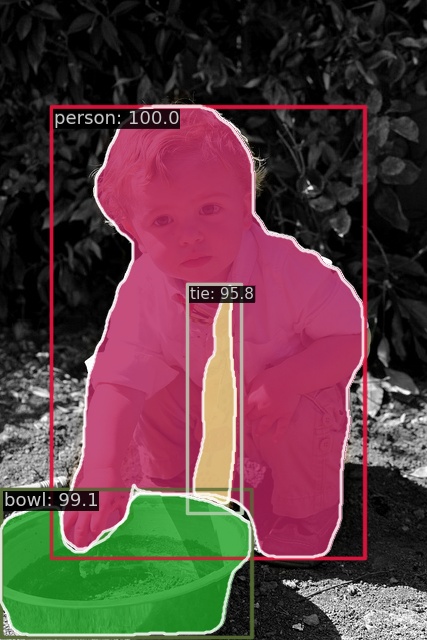}&
   \addFig{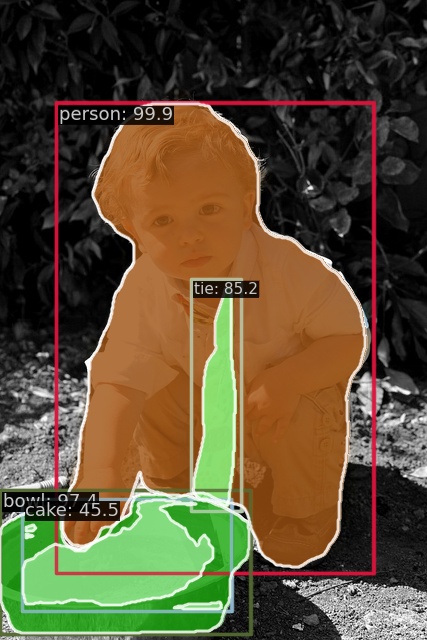}&
   \addFig{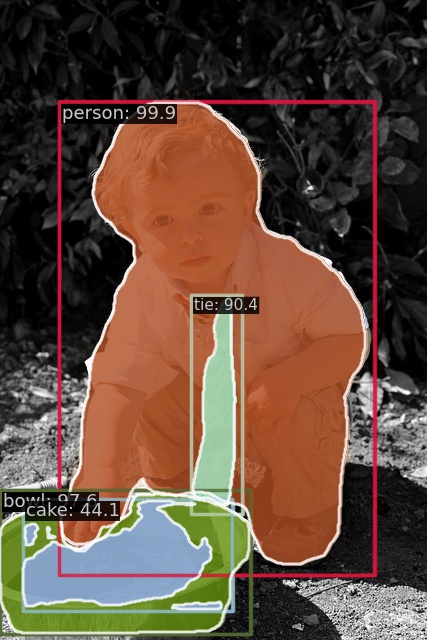}
   \\
   Image & Baseline & +Fsa-Dot-R1 &+Fsa-Lin-R1 &+Fsa-Dot-R2 & +Fsa-Lin-R2\\
  \end{tabular}
  \caption{Visualization of instance segmentation for COCO.}
  \label{fig:cocoseg}
\end{figure}
}

\subsection{Performance on Transformer architectures}
We have verified the validity of our technique on CNN architecture. Next, we conducted an analysis for applying frequency self-attention to Transformer architecture. Our frequency self-attention can serve as an effective substitute for global self-attention. There are a large number of self-attention blocks served as the basic components for building the Transformer backbone. Although they share the same form of expression, some focus on the global area, while others focus on the local area. In each layer of ViT\cite{dosovitskiy2020image}, there are parallel global and local self-attention. Even if replacing global self-attention with frequency self-attention, it is difficult to accelerate the entire layer. Fortunately, the current hybrid architecture of CNN and Transformer, such as Segformer\cite{xie2021segformer}, allows some self-attention layers to only play the role of global attention through reasonable structural design. We conducted experiments based on the trained Segformer models provided by MMSegmentation\cite{mmseg2020} on ADE20K\cite{zhou2019semantic}. FsaNet-Lin version is used in the following experiments. We adopt slide mode with crop size of $(512,512)$ and stripe of $(384, 384)$ in the inference stage. The backbone of Segformer\cite{xie2021segformer} is named MiT and has 6 versions from small to large, B0 to B5. We found that all self-attention layers of Stage 1 in MiT focus on the global area. This shows that Segformer\cite{xie2021segformer} obtains the global information in the previous layers just as the original Transformer ViT\cite{dosovitskiy2020image} does. Due to the down-sampling of key and value in Segformer\cite{xie2021segformer}, we reduce the dimension by $P1$ and $P2$ for query and (key, value) respectively. Due to the highest resolution of the input feature map in Stage 1, applying frequency self-attention in this stage is the most helpful in saving computational cost. Our method can compress the number of query from $128\times 128$ to $16\times 16$ by $P1$ and the number of (key, value) from $16\times 16$ to $8\times 8$ by $P2$. We directly replaced all self-attention layers in Stage 1 with FsaNet-Lin and presented the results in the situation of no retraining and fine-tuning, respectively.

From \tabref{tab:Segformer}, it can be observed that the larger the network, the more suitable it is to use frequency self-attention. Without retraining, the performance of MiT-B5 after replacement is even better than before. After a small amount of fine-tuning, the performance of all networks replaced by FsaNet-Lin in Stage 1 has been improved. The above experiments sufficiently demonstrate the potential performance of FsaNet on Transformer.
\CheckRmv{
\begin{table}[tbp]
    \caption{Performance of Segformer with FsaNet on ADE20K.}
    \label{tab:Segformer}
    \scriptsize
    \tabFormat
    \setlength{\tabcolsep}{0.0mm}
    \begin{minipage}[t]{0.48\linewidth}
    \vspace{0pt}
	\makeatletter\def\@captype{table}\makeatother
    \begin{tabular}{lcc}
    \toprule
     & No retraining &\\
    \midrule
    Replace layers &          Before &  After\\
    of Stage 1 &          mIoU(\%) & mIoU(\%)\\
    \midrule
    MiT-B0 & 38.38 & 37.74$\textcolor{green}{\downarrow}_{0.64}$\\
    MiT-B1  &43.23 & 41.91$\textcolor{green}{\downarrow}_{1.32}$\\
    MiT-B2 &46.37 & 46.33$\textcolor{green}{\downarrow}_{0.04}$\\
    MiT-B3 &48.39 & 48.26$\textcolor{green}{\downarrow}_{0.13}$\\
    MiT-B4 &49.03 & 48.97$\textcolor{green}{\downarrow}_{0.06}$\\
    MiT-B5 &49.75 & 49.78$\textcolor{red}{\uparrow}_{0.03}$\\
    \bottomrule
    \end{tabular}
    \end{minipage}
    \quad\begin{minipage}[t]{0.48\linewidth}
    \vspace{0em}
    \makeatletter\def\@captype{table}\makeatother
    \begin{tabular}{lcc}
    \toprule
     & Fine-tuning &\\
    \midrule
    Replace layer &    Before  & After\\
    of Stage 1     &    mIoU(\%)& mIoU(\%)\\
    \midrule
    MiT-B0 &38.38 & 38.44$\textcolor{red}{\uparrow}_{0.06}$\\
    MiT-B1 &43.23 & 43.27$\textcolor{red}{\uparrow}_{0.04}$\\
    MiT-B2 &46.37 & 46.40$\textcolor{red}{\uparrow}_{0.03}$\\
    MiT-B3 &48.39 & 48.46$\textcolor{red}{\uparrow}_{0.07}$\\
    MiT-B4 &49.03 & 49.14$\textcolor{red}{\uparrow}_{0.11}$\\
    MiT-B5 &49.75 & 49.94$\textcolor{red}{\uparrow}_{0.19}$\\
    \bottomrule
    \end{tabular}
    \end{minipage}
\end{table}
}

\section{Complexity Analysis}\label{sec:Experiments_Complex}
\subsection{Theoretical complexity Comparison}
The original implementation of self-attention includes two parts: token mapping ($1\times 1$ convolution) and token mixing. To simplify self-attention, some methods add dimension reduction operations. In order to compare the complexity of different self-attention simplification methods, we give the overall $N$-dependent complexity and a more precise representation, where $N=HW$. To start, we denote the input channel dimension as $C$ and the channel dimension of query and key as $d$. The channel dimension of value can be set to $C$ or $d$. When set to $d$, an additional $1\times 1$ convolution is needed to restore the channel to $C$. 

We first calculate the complexity for FsaNet. For token mapping, the specified frequency block of size $(k, k)$ is used as the input and the complexity is $O(4k^{2}Cd)$. For token mixing, we do not change the calculation order like other linear structure methods because $k$ is already very small, and the complexity is $O(2k^{4}d)$. For dimension reduction, due to different implementations for frequency domain transformation, there exist three algorithms with different complexities of $O(HWk^{2}C)$, $O(HWkC+\min(H,W)k^2C)$ and $O(HW\log_{2}(HW)C)$. First, we may directly multiply the vectorized input to the matrix $P$. Second, multiply each channel respectively from left and right to $D^T_{H,k}$ and $D_{W,k}$, and then reshape it into a vector. Third, obtain the complete frequency block of size $(H, W)$ by a fast 2D-DCT, retain the specified frequency block, and then reshape it into a vector. For large $k$ value, fast DCT is the most efficient. However, this might not be the case for smaller dimensions, since fast DCT, unlike the other two strategies, calculates the entire frequency components. Hence, we conduct empirical studies for selecting the fastest algorithm. Taking $H=W=97$, we test the running time of these algorithms with different $k$, as shown in \figref{fig:dcttime}.
\CheckRmv{
\begin{figure}[t]
  \centering
  \begin{overpic}[width=0.65\linewidth]{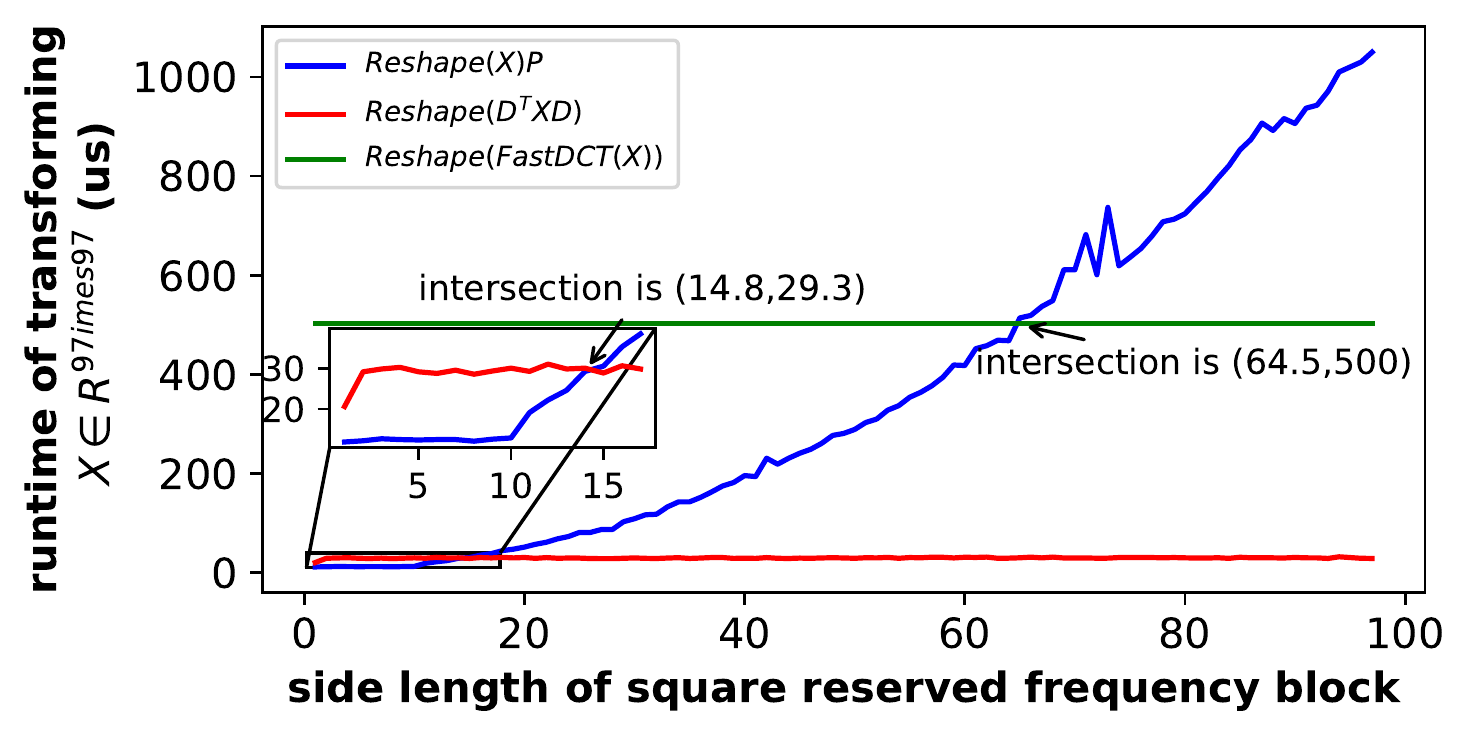}
  \end{overpic}
  \caption{The running time of the three methods under different $k$ values.}
  \label{fig:dcttime}
\end{figure}
}
The running time of the third algorithm has the longest running time when k is less than 64. The main reason is that GPU parallelization is considered, which is highly effective for matrix multiplication but has a marginal effect on fast DCT.
When $k$ is less than 14, the first algorithm runs the fastest since there are no time-consuming permutations. Therefore, we choose the first one and the complexity is $O(Nk^{2}C)$.

Then we calculate the complexity of the other self-attention simplification methods. \CCNet~only collects information from pixels in the same row or column. The complexity of token mixing is $O(2N(H+W)d)$, in which $H=W=\sqrt{N}$ is usually taken. For \ISANet, assume that the local window size is $(p_{h}, p_{w})$, and note that $P=p_{h}p_{w}$, then the complexity of token mixing complexity is $O(2dNN/P+2dNP)=O(2N(N/P+P)d) \ge 4N\sqrt{N}d$. \ANN~has two global blocks, AFNB and APNB. The query in AFNB is mapped by high level features with $C_{h}$ channels, and key and value are mapped by low level features with $C_{l}$ channels. In APNB, query, key and value are all mapped by high level features with $C_{h}$ channels. They both adopt pyramidal pooling to reduce the number of bases to $S$. The $N$-dependent complexity are same and we count on APNB. Taking pool scales as $(1, 3, 6, 8)$, $S$ will be $1^2+3^2+6^2+8^2=110$. \EMANet~introduces learnable parameters with two $1\times 1$ convolutions, and the channel number is always maintained at $C$. It generates $K$ Gaussian bases and corresponding attention maps through $T$ iterations, where $T=3$, $K=64$. \OCRNet~generates $K$ soft object region based on the pre-segmentation results, and $K$ is the class number. In particular, \ANN~and \OCRNet~fuse context by concatenating the output with the input instead of adding them directly, which adds $O(2NC_{h}^{2})$ and $O(2NC^{2})$ to the $1\times 1$ Conv part. \ourM is the only method that reduces the dimension of query, key and value at the same time by applying $1\times 1$ convolution after dimension reduction. This not only reduces the complexity of the token mixing part sharply, but also reduces the complexity of convolution, hence a sharp decline in computation. 
\CheckRmv{
    \begin{table}[tbp]
    \tiny
    \scriptsize
    \centering
    \tabFormat
    \setlength{\tabcolsep}{0.3mm}
    \caption{Comparision of complexity. 
    }
    \begin{tabular}{ lcccc }
        \toprule
        Method    & Overall & $1\times 1$ Conv + Reduce dim + Token mix\\
        \midrule
        Nonlocal\cite{wang2018non} & $O(N^{2})$     & $O(4NCd)$+$O(0)$+$O(2N^{2}d)$\\
        \CCNet     & $O(N\sqrt{N})$ & $O(4NCd)$+$O(0)$+$O(4N\sqrt{N}d)$\\
        \ISANet    & $O(N\sqrt{N})$ & $O(8NCd)$+$O(0)$+$O(2N(N/P+P)d)$\\
        \ANN       & $O(N)$         & $O(2NC_{h}d+2NC_{l}d+2NC_{h}^{2})$+$O(4Nd)$+$O(2NSd)$\\
        \EMANet    & $O(N)$         & $O(2NC^{2})$+$O(2NKTC)$+$O(NKC)$\\
        \OCRNet    & $O(N)$         & $O(2NCd+2KCd+2NC^{2})$+$O(NKC)$+$O(2NKd)$\\ 
        FsaNet     & $O(N)$         & $O(4k^{2}Cd)$+$O(Nk^{2}C)$+$O(2k^{4}d)$\\
        \bottomrule 
    \end{tabular}
    \label{tab:Complex}  
    \end{table}
}
\subsection{Memory, FLOPs and Run time}
For more precise comparison, we also report the memory, FLOPs (Floating Point Operations) and run time in \tabref{Tab:FLOPs} on Cityscapes with $769\times 769$ crop size. We counted FLOPs for all operations including matrix multiplication, Hadamard product, 'fro' norm, summation along a dimension, and Softmax. A multiply/add/exponential/square operation is counted as a FLOP. The input is a feature map of size $512\times 97\times 97$ and the embedding dimension is $64$. 
\ISANet~implements global operations by combining intra-window self-attention and inter-window self-attention, and the sum of FLOPs for both self-attention is counted. The AFNB and APNB in \ANN~can provide context separately, so we only count on the AFNB which has more FLOPs. Although the FLOPs of \CCNet~and \ISANet~are lower than \ANN, their actual running time is higher, which is attributed to the fact that dimension permutation and reshaping consume time but are not counted in FLOPs. Notably, a larger class number of the dataset can result in slower running for \OCRNet.
Compared to \NonlocalNetwork, FsaNet-Dot reduces computation by 90.04\% in memory, 98.18\% in run time, and 98.07\% in FLOPs; FsaNet-Lin reduces the computation by 87.29\% in memory, 97.56\% in run time, and 96.13\% in FLOPs. Considering that the input and output already account for 9.57\% of the non-local memory, the memory cost in self-attention processing drops even greater. 
\CheckRmv{
\begin{table}[tbp]
    \scriptsize
    \tabFormat
    \setlength{\tabcolsep}{2.2mm}
    \caption{Comparison of GPU, FLOPs and Run time.}
    \begin{tabular}{lccc}
    \toprule
    Method &   GPU memory(M) & FLOPs(G) &Run time(ms)\\
    \midrule
    Non-local\cite{wang2018non}& 402.82 & 25.30 & 35.1\\
    \CCNet  &264.5 & 16.56 & 12.80\\
    \ISANet &204.8 & 14.62 & 8.03\\        
    \ANN    &89.5  & 17.86 & 4.54\\
    \EMANet &94.0  & 11.72 & 3.84\\
    \OCRNet &82.7  & 11.50 & 2.90\\
    FsaNet-Lin &51.19  & 0.98  & 0.86\\
    FsaNet-Dot &40.14  & 0.49  & 0.64\\
    \bottomrule
    \end{tabular}
    \label{Tab:FLOPs}
\end{table}
}

\section{Conclusion and Future Work}\label{sec:futurework}
Inspired by the idea of decoupling high and low frequencies, we performed ablation experiments on high-frequency components of self-attention input and discovered that up to 98\% of the high-frequency components do not significantly affect the performance. Based on this, we transform the input to the frequency domain and exploit the sparsity of the frequency domain to propose a frequency self-attention mechanism with the lowest computational cost. With CNN-based backbone, our~\ourM consistently achieves an outstanding performance on the well-known datasets, namely Cityscapes\cite{cordts2016cityscapes}, ADE20K\cite{zhou2019semantic}, VOCaug\cite{pascal-voc-2012}, and COCO\cite{lin2014microsoft}. Our frequency self-attention also generates promising results in transformer structures, highlighting its great potential in the design of network structures.
\section*{Acknowledgments}
This work was supported by the Research and Application of Control System Technology for Group Operation of Heavy-haul Trains Program, grant number: China Shenhua Energy Co., Ltd (2022) 127. The authors are grateful to the High Performance Computing Center of Central South University for partial support of this work. Fengyu Zhang acknowledges the financial support from China Scholarship Council, grant number: 202006370295.

{\appendix[Definition and properties for \texorpdfstring{$P$}. and \texorpdfstring{$G$}.]
\section*{Definition of \texorpdfstring{$P$}. and \texorpdfstring{$G$}.}\label{sec:PG}
We first define two reshape functions and two 2D-DCT transform functions. 
We denote by $\psi$ a reshape  function of a 3d-tensor into a 2D-tensor. $\psi$ acts channel wise, and on every channel, row-wise vectorizes a 2D map to a 1D map. Note that we use the same notation $\psi$ for any matrix size. $\psi^{-1}$ represents the inverse reshape function from a 2D-tensor into a 3D-tensor,  converting a 1D map to a 2D map on every channel. Further, we denote by $\phi$ the 2D-DCT transform of the selecting $k$ lowest horizontal and $k$ lowest vertical bases. It transforms a 3D-tensor in $\bbR^{C\times  H\times W}$ into $\bbR^{C\times  k\times k}$. $\phi^{-1}$ represents 2D-IDCT reconstruction based on $k$ horizontal and $k$ vertical low-frequency bases and zero padding. It transforms a 3D-tensor in $\bbR^{C\times  k\times k}$ to $\bbR^{C\times  H\times W}$. 
Since the above operation does not introduce any non-linearity, we assume that there exists a matrix $P\in \bbR^{HW\times k^{2}}$ such that $\psi(\phi(\psi^{-1}(X^{'})))=X^{'}P$. 
Similarly, we also assume that there is a matrix $G\in \bbR^{k^{2}\times HW}$ such that $\psi(\phi^{-1}(\psi^{-1}(f^{'})))=f^{'}G$. 
Take $X^{'}$ as an identity matrix $E_{1}^{'} \in \bbR^{HW\times HW}$, and take $f^{'}$ as another identity matrix $E_{2}^{'} \in \bbR^{hw\times hw}$: 
\begin{eqnarray}
    \psi(\phi(\psi^{-1}(E_{1}^{'})))=E^{'}P,\quad 
    \psi(\phi^{-1}(\psi^{-1}(E_{2}^{'})))=E_{2}^{'}G.
\end{eqnarray}

Hence, we have confirmed the existence of $P$ and $G$: 
\begin{eqnarray}
\begin{aligned}
P=\psi(\phi(\psi^{-1}(E_{1}^{'}))),\quad G=\psi(\phi^{-1}(\psi^{-1}(E_{2}^{'}))).
\end{aligned}
\end{eqnarray}

\section*{Properties of \texorpdfstring{$P$}. and \texorpdfstring{$G$}.}
To prove \textbf{First property} $\mathbf{G=P^{T}}$, let $D_n$ denote a complete $n$-dimensional DCT matrix, and each column represent a DCT basis corresponding to a single frequency. We denote $D_n=\left[d_{i,j}:=\cos\left(\frac\pi n\left(i+\frac 12\right)j\right)\right]$ for $i=0,1,\ldots,n-1$ and $j=0,1,\ldots,n-1$\footnote{In this paper, we index an $n-$array by $0,1,\ldots,n-1$. This holds true for multi-dimensional arrays.}.
Suppose $D_{n,k}$ consists of the set of basis (columns) in $D_n$  corresponding to the $k\leq n$ lowest frequencies. In other words, $D_{n,k}=[d_{i,j}]$ for $i=1,2,\ldots,n$ and $j=1,2\ldots,k$. For a 1d signal $\mathbf{ x^{T}}$, $\mathbf{ x^{T}} D_{n,k}$ represents the low frequency coefficient vectors. As such, $\mathbf{ x^{T}}D_{n,k}D_{n,k}^{T}$ is the reconstructed signal that only contains the lowest $k$ frequencies, i.e. a low-pass filtered signal. Since the columns of $D_n$ are orthonormal, $D^T_{n,k}D_{n,k}$ is always the identity matrix. Next, we note that the 2D-DCT transform of an $H\times W$ signal $X$ (with a single channel $C=1$) can be obtained by performing $D_H^TXD_W$. In particular, the $k\times k$ coefficients of the $k$ lowest horizontal and vertical frequencies is given by $X_f=\phi(X)=D_{H,k}^TXD_{W,k}$. For multiple channels, this procedure is repeated individually for each channel. Now, we compute each element in $P$ and $G$ individually.

\textbf{Calculating $P$:} We recall $P=\psi(\phi(\psi^{-1}(E_{1}^{'})))$ and hence reshape $E_{1}^{'}$ to $\psi^{-1}(E_{1}^{'}) \in R^{HW\times H \times W}$. For $m=0,1,\ldots, HW-1$, $\psi^{-1}(E_{1}^{'})[m,:,:]$ represents the $m^{\mathrm{th}}$ matrix, in which only includes one non-zero element. To express the index of the non-zero element, the integer quotient ($//$) and remainder ($\%$) operators of integer division are adopted. All the elements in $\psi^{-1}(E_{1}^{'})[m,:,:]$ are zeros but $\psi^{-1}(E_{1}^{'})[m,m//W,m\%W]$ is one. Next, we perform 2D-DCT transform $\phi$ on the $m^{\mathrm{th}}$ matrix by left multiplying $D_{H,k}^{T}$ and right multiplying $D_{W,k}$, which provides $\phi(\psi^{-1}(E_{1}^{'})[m,:,:])=D_{H,k}^{T}\psi^{-1}(E_{1}^{'})[m,:,:]D_{W,k}=D_{H,k}[m//W,:]^{T}D_{W,k}[m\%W,:]$.
We may also write it as:
\begin{eqnarray}
    \begin{tiny}
        \phi(\psi^{-1}(E_{1}^{'}))[m,:,:]
        =\begin{bmatrix} 
        D_{H,k}[m//W,0]D_{W,k}[m\%W,:],\\
        D_{H,k}[m//W,1]D_{W,k}[m\%W,:],\\
        \cdots,\\
        D_{H,k}[m//W,k-1]D_{W,k}[m\%W,:]\\
        \end{bmatrix}
    \end{tiny}
\end{eqnarray}

Expanding $\phi(\psi^{-1}(E_{1}^{'})[m,:,:]$ row by row, we obtain the $m^{\mathrm{th}}$ row of $\psi(\phi((\psi^{-1}(E_{1}^{'}))$, which is also the $m^{\mathrm{th}}$ row of $P$. So the element in the $m^{\mathrm{th}}$ row and $n^{\mathrm{th}}$ column of $P$ for $m\in\{0,1,\ldots, HW-1\}$ and $n\in\{0,1,\ldots,k^2-1\}$ is:
\begin{eqnarray}
    \begin{tiny}
    P[m,n]=D_{H,k}[m//W,n//k]D_{W,k}[m\%W,n\%k].
    \end{tiny}
\end{eqnarray}

\textbf{Calculating $G$:} Remember that $G=\psi(\phi^{-1}(\psi^{-1}(E_{2}^{'})))$ and reshape $E_{2}^{'}$ to $\psi^{-1}(E_{2}^{'}) \in R^{k^{2}\times k \times k}$. For $m\in \{0,1,\ldots, k^2-1\}$, $\psi^{-1}(E_{2}^{'})[m,:,:]$ represents the $m^{\mathrm{th}}$ matrix, in which all elements are zeros but $\psi^{-1}(E_{2}^{'})[m,m//k,m\%k]$ is one. Next, we perform 2D-IDCT transform on the $m^{\mathrm{th}}$ matrix by left multiplying $D_{H,k}$ and right multiplying $D_{W,k}^{T}$, which provides $\phi^{-1}(\psi^{-1}(E_{2}^{'})[m,:,:])=D_{H,k}\psi^{-1}(E_{2}^{'})[m,:,:]D_{W,k}^{T}=D_{H,k}[:,m//k]D_{W,k}[:,m\%k]^{T}$. 
It can also be written as:
\begin{eqnarray}
    \begin{tiny}
        \phi^{-1}(\psi^{-1}(E_{2}^{'}))[m,:,:]
        =\begin{bmatrix} 
        D_{H,k}[0,m//k]D_{W,k}[:,m\% k]^{T},\\ 
        D_{H,k}[1,m//k]D_{W,k}[:,m\%k]^{T},\\
        \cdots,\\
        D_{H,k}[H-1,m//k]D_{W,k}[:,m\%k]^{T}\\
        \end{bmatrix}
    \end{tiny}
\end{eqnarray}

Expanding $\phi^{-1}(\psi^{-1}(E_{2}^{'}))[m,:,:]$ row by row, we obtain the $m^\mathrm{th}$ row of $\psi(\phi^{-1}(\psi^{-1}(E_{2}^{'})))$, which is also the $m^\mathrm{th}$ row of $G$. So the elements in the $m^\mathrm{th}$ row and the $n^\mathrm{th}$ column of $G$ for $m\in\{0,1,\ldots,k^2-1\}$ and $n\in\{0,1,\ldots,HW-1\}$ is:
\begin{eqnarray}
\begin{tiny}
    G[m,n]=D_{H,k}[n//W,m//k]D_{W,k}[n\%W,m\%k].
\end{tiny}
\end{eqnarray}

Now, we immediately observe this property by comparing the computed values of $P[m,n]$ and $G[m,n]$ and directly verifying that $P[m,n]=G[n,m]$. 

We verify \textbf{Second property} $\mathbf{P^{T}P=E_{k^{2}}}$ holds under the condition that $H=W$. We observe $P^{T}P[m,n]=\sum_{r=0}^{H^{2}-1}P^{T}[m,r]P[r,n]$ and it can be explicitly written as: 
\begin{eqnarray}
\begin{small}
    \begin{aligned} 
    &P^{T}P[m,n]=\sum_{r=0}^{H^{2}-1} D_{H,k}[r//H,m//k]D_{W,k}[r\%H,m\%k]\\
    &D_{H,k}[r//H,n//k]D_{W,k}[r\%H,n\%k]. 
    \end{aligned}
\end{small}
\end{eqnarray} 

When $r$ increases from $0$ to $H^{2}-1$, $r//H$ does not change during every period of $H$ values. Hence, 
\begin{eqnarray}
\begin{small} 
    \begin{aligned}
    &P^{T}P[m,n]\\
    =&D_{H,k}[0,m//k]D_{H,k}[0,n//k]\\
    &\sum_{r=0}^{H-1}D_{W,k}[r\%H,m\%k]D_{W,k}[r\%H,n\%k]+\hdots\\
    +&D_{H,k}[H-1,m//k]D_{H,k}[H-1,n//k]\\
    &\sum_{r=H^{2}-H}^{H^{2}-1}D_{W,k}[r\%H,m\%k]D_{W,k}[r\%H,n\%k]. 
    \end{aligned}
\end{small}
\end{eqnarray} 

On the other hand, $r\%H$ assumes similar values in every period of $H$ values. Accordingly, we can conclude that:
\begin{eqnarray}
\begin{small} 
    \begin{aligned}
&\sum_{r=0}^{H-1}D_{W,k}[r\%H,m\%k]D_{W,k}[r\%H,n\%k]=\hdots \\
=&\sum_{r=H^{2}-H}^{H^{2}-1}D_{W,k}[r\%H,m\%k]D_{W,k}[r\%H,n\%k]. 
    \end{aligned}
\end{small}
\end{eqnarray} 

Combining the above two formulas we can get: 
\begin{eqnarray}
\begin{small} 
    \begin{aligned}
&P^{T}P[m,n]\\
=&\sum_{r=0}^{H-1}D_{H,k}[r,m//k]D_{H,k}[r,n//k]\\
&\sum_{r=0}^{H-1}D_{W,k}[r,m\%k]D_{W,k}[r,n\%k]\\
=&D_{H,k}[:,m//k]^{T}D_{H,k}[:,n//k]D_{W,k}[:,m\%k]^{T}D_{W,k}[:,n\%k].
    \end{aligned}
\end{small}
\end{eqnarray} 

When $m=n$, $P^{T}P[m,n]=1$. When $m\neq n$, $P^{T}P[m,n]=0$. This concludes the result. We also provide pseudo code for generating $P$ and $G$.
\begin{algorithm}[H]
    \small
    \caption{Generate P, G}
    \begin{algorithmic}[1] 
        \REQUIRE {$E^{'}$ (Identity matrix with size $HW\times HW$)}
        \ENSURE {$P$}
        \STATE {$E=E^{'}.reshape(HW,H,W)$}
        \STATE {$Coef=DCT2d(E)$}
        \STATE {$LowCoef=Coef[:,:k,:k]$}
        \STATE {$P=LowCoef.reshape(HW,k^2)$}
        \STATE {$G=P.transpose(0,1)$}
    \end{algorithmic} 
\end{algorithm}
}\label{Appendix}

{\small
\bibliography{FsaNet}
\bibliographystyle{IEEEtran}
}

\vfill
\end{document}